\documentclass{article}
\usepackage{graphicx} 

\usepackage{blindtext}
\usepackage{amsmath}
\usepackage{amsfonts}
\usepackage{graphicx}
\usepackage{caption}
\usepackage{subcaption}
\usepackage{mathtools}
\usepackage{algorithm}
\usepackage{sidecap}
\usepackage{tikz}

\usepackage{booktabs}
\usepackage{pifont}
\usepackage{authblk}
\usepackage{url} 
\usepackage{algpseudocode}
\usepackage{geometry}
 \usepackage{natbib}
 \usepackage{xcolor}

\definecolor{dark-gray}{gray}{0.3}
\definecolor{dkgray}{rgb}{.4,.4,.4}
\definecolor{dkblue}{rgb}{0,0,.5}
\definecolor{medblue}{rgb}{0,0,.75}
\definecolor{rust}{rgb}{0.5,0.1,0.1}

\usepackage[colorlinks=true]{hyperref}
\hypersetup{linkcolor=dkblue}    
\hypersetup{citecolor=rust}      
\hypersetup{urlcolor=rust}

\title{Auto-differentiable data assimilation: \\ Co-learning of states, dynamics, and filtering algorithms}

\author[1]{Melissa Adrian\thanks{Corresponding author: \url{madrian@purdue.edu}}}
\author[2]{Daniel Sanz-Alonso}
\author[2,3,4]{Rebecca Willett}

\affil[1]{School of Mechanical Engineering, Purdue University, West Lafayette, IN 47907}
\affil[2]{Department of Statistics, The University of Chicago, Chicago, IL 60637}
\affil[3]{Department of Computer Science, The University of Chicago, Chicago, IL 60637}
\affil[4]{NSF-Simons National Institute for Theory and Mathematics in Biology, Chicago, IL 60611}

\begin{document}

\maketitle

\interfootnotelinepenalty=10000

\begin{abstract}
Data assimilation algorithms estimate the state of a dynamical system from partial observations, where the successful performance of these algorithms hinges on costly parameter tuning and on employing an accurate model for the dynamics. This paper introduces a  framework for jointly learning the state, dynamics, and parameters of filtering algorithms in data assimilation through a process we refer to as auto-differentiable filtering. The framework leverages a theoretically motivated loss function that enables learning from partial, noisy observations via gradient-based optimization using auto-differentiation.  
We further demonstrate how several well-known data assimilation methods can be learned or tuned within this framework. To underscore the versatility of auto-differentiable filtering, we perform experiments on dynamical systems spanning multiple scientific domains, such as the Clohessy-Wiltshire equations from aerospace engineering, the Lorenz-96 system from atmospheric science, the generalized Lotka-Volterra equations from systems biology, and the two-layer quasi-geostrophic model from geophysical fluid dynamics. Finally, we provide guidelines for practitioners to customize our framework according to their observation model, accuracy requirements, and computational budget.
\end{abstract}

\section{Introduction}
In many applied scientific settings, practitioners have access to observational data that provides a partial and noisy snapshot of a dynamical system. This data may be utilized for a variety of purposes, including inferring interpretable parameters of the system, improving forecast skill, and reconstructing the underlying state of the system in the past. However, the noisy and incomplete nature of these observations poses significant challenges for both inference and prediction.
\textit{Data assimilation} provides a class of methods to combine data sources of various modalities, denoise, and ``fill in the gaps'' in a principled way.  

Data assimilation was proposed in \citet{Richardson1922} for atmospheric state estimation with the goal of improving weather prediction. Early data assimilation methods include the Kalman filter \citep{Kalman1960} and optimal interpolation \citep{Gandin1966}, which handle assimilating observations from linear systems, both in terms of dynamics and observation models. As the need increased for data assimilation algorithms that could handle nonlinearities, the variational approaches 3DVar \citep{Lorenc1986} and 4DVar \citep{Courtier1994} were developed. The landscape for data assimilation significantly changed when ensemble-based methods \citep{Palmer1992} were introduced, allowing for improved notions of uncertainty quantification for nonlinear systems. A prototypical ensemble-based data assimilation method is the ensemble Kalman filter \citep[EnKF]{Evensen1994, Evensen2003}, which inspired hybrid variational-ensemble approaches such as the ensemble 3DVar \citep[Ens3DVar]{Hamill2000} and the ensemble 4DVar \citep[Ens4DVar]{Liu2008}. Among these choices, there is no one catch-all algorithm; practitioners weigh each approach's advantages and disadvantages and choose the algorithm that balances specific considerations for their applied needs.

One important challenge in data assimilation is that the dynamics model governing the evolution of the state is often expensive to simulate and subject to model error. A recent and promising approach to alleviate these issues is to leverage machine-learned surrogate models for the dynamics \citep{chen2023stochastic,Bach2025book,adrian2025data,sanz2025long, Chakraborty2026}. 
The wide success of machine learning in building surrogates, particularly in emulating physical systems, can be attributed to the open availability of high-quality training data. In many domains, expensive processing to curate this high-quality training data needs to occur before surrogate learning even begins. 
Particularly for high-dimensional settings, fine-grained information may not be readily available, and data assimilation algorithms may be required to distill fine-scale features from coarse measurements. Thus, there is a bidirectional relationship between machine learning and data assimilation: data assimilation algorithms benefit from machine-learned dynamics models, and machine learning surrogate models benefit from training data processed using data assimilation algorithms.

Data assimilation algorithms, though essential in many applications, require substantial tuning
before they can be deployed in practice. For example, in 3DVar a background covariance matrix describing the time-independent covariances of forecast errors needs to be curated prior to using this algorithm. In atmospheric data assimilation, the background error covariance has been traditionally estimated via the National Meteorological Center (NMC) method \citep{Parrish1992}, which aims to estimate this error covariance matrix in the absence of complete ground truth data. Moreover,  parameters such as covariance inflation and the covariance mixing parameter in ensemble schemes additionally require prior specification, requiring multiple runs of the algorithm for tuning.

In this paper, we present a general framework that allows for learning updates to the dynamics model and data assimilation parameters based on noisy, potentially sparse observations of an underlying dynamical system, thereby streamlining the process to directly utilize observations. The framework can be used to learn nonlinear dynamics given linear\footnote{We note that the proposed framework can be used as written for linear observation settings; however, modifications can be made to the filtering computation to accommodate nonlinear observations. The operation for the adjoint of observation operator $H$ may be avoided in ensemble-based filtering algorithms via ordering of operations, allowing for a nonlinear $H(\cdot)$ to be consistently applied. For single particle filtering methods, a linearization of $H(\cdot)$ may be applied such that $\mathbf{H} = \left.\frac{\partial H}{\partial x}\right|_{x = x_t}$ and used wherever $H^\top$ appears.} observations of the state. As we will detail below, our framework generalizes existing learned data assimilation methods, including AD-EnKF \citep{Chen2022} and AD-3DVar-$K$ \citep{Levine2022}. Furthermore, our framework leads to new methods that are efficacious for time-varying observation models or which require fewer computational resources. We formalize the problem setting in the next section.

\subsection{Problem setting}

Assume we collect observations $y_{1:T}$ derived from the following hidden Markov model 
\begin{align}
    x^{*}_t &= \mathcal{F}_{\theta_1^*}(x_{t-1}^{*}) + \tilde Q_{\theta_2^*}\xi_t, \qquad \xi_t \sim N(0,I_{d_x}),\label{eq:hmm_forecast} \\ 
    y_t &= H_tx_t^{*} + \eta_t, \quad \quad \quad  \quad  \quad \ \ \eta_t \sim N(0,R_t),\label{eq:hmm_obs} \\
    x^{*}_0  & \sim p_0(x_0) \label{eq:hmm_init}
\end{align}
\noindent for time points $t=1,\dots,T$, where $x_t^*\in \mathbb{R}^{d_x}$ is the true state vector at time $t$, $\mathcal{F}_{\theta_1^*}$ is the true dynamics map parameterized by true parameters $\theta_1^*$, $Q_{\theta_2^*}:=\tilde Q_{\theta_2^*}\tilde Q_{\theta_2^*}^\top\in \mathbb{R}^{d_x \times d_x}$ is the true forecast error covariance parameterized by $\theta_2^*$, $y_t\in \mathbb{R}^{d_{y_t}}$ are noisy, potentially partial observations of $x_t^*$ at time $t$, $H_t\in \mathbb{R}^{d_{y_t}\times d_x}$ is the time-varying, known linear observation matrix, $R_t \in \mathbb{R}^{d_{y_t} \times d_{y_t}}$ is the time-varying, known observation error covariance, and the initialization $x^{*}_0$ is assumed to be unknown and drawn from a given prior distribution $p_0.$ In this setting, we assume independence of $x^{*}_0$, $\xi_{1:T}$, and $\eta_{1:T}$.\footnote{To explain our notation, $\xi_{1:T}$, as an example, denotes the time series of forecast model errors for time points $t=1,\dots,T$.}

In practice, the true dynamics mapping $\mathcal{F}_{\theta_1^*}$ is unknown, and instead, an imperfect model given by $\mathcal{F}_{\theta_1}$, may be used, where $\theta_1$ are the parameters of this imperfect dynamics mapping. We note that  $\mathcal{F}_{\theta_1}$ may be structurally different from $\mathcal{F}_{\theta_1^*}$ (e.g., $\mathcal{F}_{\theta_1}$ could be a neural network with learnable parameters $\theta_1$, where these parameters do not have a one-to-one correspondence with the true governing parameters $\theta_1^*$). This imperfect specification of the dynamics induces forecast errors $\xi_t$ that are assumed to be Gaussian distributed as specified in \eqref{eq:hmm_forecast}.

One goal in this setup is to compute, for each $t =1 , \ldots, T,$ a state estimate $x_{t}$, referred to as an \emph{analysis}, given observations $y_{1:t}$ up to the current time $t$. This goal is achieved by the use of filtering algorithms in data assimilation, which often require specifying filtering parameters $\phi$ prior to deployment that control, for instance, the ``trust" given to the forecasts compared to the observations. A key idea underlying our framework is that information obtained from computing these analyses via filtering can be used to update the imperfect forecast model's parameters $\theta:= \{\theta_1,\theta_2\}$  and the filtering parameters $\phi$. In other words, we aim to learn forecast parameters $\theta$ and filtering parameters $\phi$ based solely on noisy, partial observations $y_{1:T}$, without the use of true states $x^*_{1:T}$. Section \ref{sec:method_learning} provides a general overview of our framework for joint learning of states, forecast parameters, and filtering parameters based on observations $y_{1:T}$, which can be applied to a broad class of scientific problems involving dynamical systems.

\subsection{Contributions}

Many prior works have explored utilizing partial, noisy observations to improve some aspect of the data assimilation pipeline, which we further detail in Section \ref{sec:related_work}. Notable examples include the auto-differentiable ensemble Kalman filter \citep[AD-EnKF]{Chen2022} and the auto-differentiable 3DVar where the Kalman gain matrix is directly learned \citep[AD-3DVar-$K$]{Levine2022}, among others. Each method was developed separately with different settings in mind, some focusing more on learning the dynamics, some focusing more on learning filtering parameters, and others focusing on learning both aspects. \textbf{In this paper, we show that these goals can all be achieved under our general framework.}

Furthermore, armed with this approach, we can easily derive new methods that have important performance advantages over past work. We provide derivations of particular auto-differentiable filtering methods in Sections \ref{sec:filtering} and \ref{sec:ad-filtering}. Specifically, we derive the auto-differentiable 3DVar algorithm where the static forecast error covariance $C$ is learned (AD-3DVar-$C$), as opposed to directly learning the Kalman gain $K$ in AD-3DVar-$K$, which is outlined in Section \ref{sec:related_work}. A benefit of AD-3DVar-$C$ is that it can handle time-varying observation models salient to many application domains. Additionally, we derive the auto-differentiable ensemble 3DVar (AD-Ens3DVar), which serves as a computational middle ground between the lighter-weight AD-3DVar-$C$ or -$K$ and AD-EnKF for jointly learning dynamics and filtering parameters. 
Our contributions in this work are summarized as follows:

\begin{enumerate}
    \item Formulate a general framework for co-learning of state, dynamics, and filtering algorithms and outline notable special cases of this framework (Section \ref{sec:method_learning}). 
    \item Propose two new algorithms: the auto-differentiable 3DVar algorithm (AD-3DVar-$C$) and the auto-differentiable hybrid ensemble 3DVar algorithm (AD-Ens3DVar) (Section \ref{sec:method_learning}).
    \item Modify the existing auto-differentiable ensemble Kalman filter (AD-EnKF), developed in \cite{Chen2022} as an algorithm for co-learning states and dynamics, to additionally co-learn filtering parameters, namely the covariance inflation (Section \ref{sec:method_learning}).
    \item Compare each particular case of our framework  (i.e., AD-EnKF, AD-Ens3DVar, AD-3DVar-$C$) on illustrative examples spanning multiple domains (astronomy, atmospheric sciences,  population dynamics, and fluid dynamics) that showcase advantages and disadvantages (Section \ref{sec:experiments}).
    \item Discuss settings where each example method derived from our framework may be useful based on performance in our examples in terms of accuracy, scalability, and computational efficiency (Section \ref{sec:discussion}). 
\end{enumerate}

\section{General learning framework}\label{sec:method_learning}
\subsection{Filtering}\label{sec:filtering}
In this work, we are particularly interested in learning from observations in a \textit{filtering} setting, where the analysis at each time $t$, denoted $x_t$, is computed using observations up to the current time point, $\{y_{1:t}\}$. 
We consider a general class of filtering algorithms of the form
\begin{equation}
    x_t^n = \left(I-\mathcal{K}_\phi(x_{t-1}^{1:N})H_t\right)\mathcal{M}_\theta(x_{t-1}^n) + \mathcal{K}_\phi(x_{t-1}^{1:N})\mathcal{Y}(y_t;R_t), \qquad \text{for }t=1,\dots,T,\label{eq:filter}
\end{equation}
\noindent where $\mathcal{\mathcal{M}}_\theta(\cdot)$ is the forecast model parameterized by $\theta$, $\mathcal{Y}(\, \cdot\, ;R_t)$ is a function that incorporates observations and the observation error covariance $R_t$, $\mathcal{K}_\phi(\cdot)$ is a Kalman gain parameterized by $\phi$ that controls the ``trust" given to the observations versus the forecast, $n$ is the ensemble member index ($n=1,\dots,N$), and $x_{t}^{1:N}$ denotes an ensemble comprising $N$ state estimates  at time $t$. 
For filtering algorithms that do not rely on an ensemble, $N=1$, and we drop the superscripts from the notation. Throughout this paper, we refer to the map from the forecast $\mathcal{M}_\theta(x_{t-1}^n)$ and observations $y_t$ to the analysis $x_t^n$ for any time $t$ as an \textit{analysis step}. We also refer to a \textit{filter} step as the map from the previous analysis $x_{t-1}^n$ and observations $y_t$ to the analysis $x_t^n$ at some time $t$, where the computation for this mapping is given in \eqref{eq:filter}.

Specific filtering algorithms can be expressed as equation \eqref{eq:filter} with particular choices of $\mathcal{K}_\phi$, $\mathcal{M}_\theta$, and $\mathcal{Y}$. The particular algorithms we consider in this paper are 
3DVar \citep{Lorenc1986} where the Kalman gain is parameterized by a fixed background error covariance matrix $C_\phi$ (denoted as ``3DVar-$C$''), the ensemble Kalman filter \citep[EnKF]{Evensen1994} where the time-evolving Kalman gain is defined in equations \eqref{eq:C_enkf_eq} and \eqref{eq:C_enkf}, and a hybrid approach, ensemble 3DVar \citep[Ens3DVar]{Hamill2000}, where we define this hybrid covariance in \eqref{eq:C_ens3dvar}.
Table \ref{tab:filtering_algs} shows how these three filtering algorithms can be expressed as particular instances of the abstract setup in equation \eqref{eq:filter}.

\begin{table}[ht]
    \centering
    \begin{tabular}{ccccc}\toprule

         Alg. &  $\mathcal{M}_\theta(x_{t-1}^n)$&  $\mathcal{K}_\phi(x_{t-1}^{1:N})$&  $\mathcal{Y}(\cdot; R_t)$&  Ens. Size \\ \midrule

        3DVar-$C$ &  $\mathcal{F}_{\theta_1}(x_{t-1})$&  $C_\phi H_t^\top (H_tC_\phi H_t^\top + R_t)^{-1}$&  $y_t$ &  $1$\\ 
  & & $C_\phi \in \mathbb{R}^{d_x\times d_x}$& \\ \midrule
  
         EnKF&  $\mathcal{F}_{\theta_1}(x_{t-1}^n)+\tilde Q_{\theta_2}\xi_t^n$ &  $\hat C_{t}^{\theta,\phi} H_t^\top(H_t\hat C_{t}^{\theta,\phi} H_t^\top + R_t )^{-1}$&  $y_t + \gamma_t^n$&  $N_{\text{EnKF}}$ \\
 & $\xi_t^n \sim \mathcal{N}(0,I_{d_x})$& ($\hat C_t^{\theta,\phi}$ defined in \eqref{eq:C_enkf})& $\gamma_t^n \sim \mathcal{N}(0,R_t)$ \\ \midrule 
         Ens3DVar&  $\mathcal{F}_{\theta_1}(x_{t-1}^n)+\tilde Q_{\theta_2}\xi_t^n$&  $\tilde C_t^{\theta,\phi} H_t^\top (H_t\tilde C_t^{\theta,\phi} H_t^\top + R_t)^{-1}$&  $y_t + \gamma_t^n$
&  $N_{\text{Ens3DVar}}$ \\ 
 & $\xi_t^n \sim \mathcal{N}(0,I_{d_x})$& ($\tilde C_t^{\theta,\phi}$ defined in \eqref{eq:C_ens3dvar}) & $\gamma_t^n \sim \mathcal{N}(0,R_t)$ \\  
 \bottomrule 

    \end{tabular}
    \caption{Table of examples of named special cases of the general filtering algorithm defined in \eqref{eq:filter}. We note that this list is not exhaustive.}
    \label{tab:filtering_algs}
\end{table}

\paragraph{Forecasting system $\mathcal{M}_\theta$.} The forecast system $\mathcal{M}_\theta$ consists of the forecasting model $\mathcal{F}_{\theta_1}$ and perturbations to the forecast model based on $\mathcal{N}(0,Q_{\theta_2})$ noise, where $\theta = \{\theta_1,\theta_2\}$.  In \eqref{eq:filter}, we define the forecast system as 
\begin{equation}
    \mathcal{M}_{\theta}(x_{t-1}^n) = \mathcal{F}_{\theta_1}(x_{t-1}^n) + s\tilde Q_{\theta_2}\xi_t^n,
\end{equation}
where $s =1$ if the user chooses to add forecast model perturbations and learn the covariance matrix $Q_{\theta_2}$, and $s=0$ if the user chooses to not add stochastic perturbations to the forecasting system. For each specific filtering algorithm we consider, we list this choice in Table \ref{tab:filtering_algs}. The decision to set $s=1$ or $s=0$ can be empirically motivated based on the choice of filtering method.

\paragraph{Forecast covariance matrix $C_t^{\theta,\phi}$ in the Kalman gain $\mathcal{K}_\phi(\cdot)$.} Each of the algorithms in Table \ref{tab:filtering_algs} constructs the Kalman gain differently, and each requires specifying a forecast error covariance matrix $C_t^{\theta,\phi}$. 

The 3DVar-$C$ algorithm relies on a fixed-time covariance matrix $C_\phi$ parameterized by static parameters $\phi$. To ensure that $C_\phi$ is a proper covariance matrix, we learn $C_\phi$ via a square matrix $B_\phi\in \mathbb{R}^{d_x \times d_x}$ such that $C_\phi = B_\phi B_\phi^\top$; this approach ensures that $C_\phi$ is a symmetric, positive semi-definite matrix. We provide a more detailed discussion with supplementary experiments regarding this choice in Appendix \ref{app:C_phi}.

For EnKF, an ensemble-based filtering algorithm, a time-\textit{dependent} empirical forecast covariance matrix is computed as 
\begin{equation}
    \hat C_t^\theta = \frac{1}{N-1}\sum_{n=1}^N(\mathcal{M}_\theta(x_{t-1}^n) - \hat m_t^\theta)(\mathcal{M}_\theta(x_{t-1}^n) - \hat m_t^\theta)^\top, \label{eq:C_enkf_eq}
\end{equation}
which approximates the forecast covariance with a finite ensemble, with the empirical forecast mean given by
\begin{equation}
    \hat m_t^\theta = \frac{1}{N}\sum_{n=1}^N \mathcal{M}_\theta (x_{t-1}^n).
\end{equation}

When $N \ll d_x$, EnKF estimates of the state substantially improve when a covariance tapering matrix $\rho \in \mathbb{R}^{d_x\times d_x}$ is used, particularly in applications where the states have a spatial structure, such as the tapering matrix described in \citet{Gaspari1999}. Moreover, the ensemble-based estimate of the forecast covariance matrix may exhibit under-dispersion, which can be combated with covariance inflation \citep{Anderson2012}. Combining the covariance inflation factor $\phi$ and the covariance tapering matrix $\rho$, the forecast covariance matrix used in EnKF filtering is  
\begin{equation}
    \hat C^{\theta,\phi}_t = (1+\phi)(\rho\circ \hat C_t^\theta),\label{eq:C_enkf}
\end{equation}
where $\circ$ denotes the element-wise product.

In Ens3DVar, the forecast covariance matrix is specified to be a convex combination of a static 3DVar covariance matrix $C_{\phi_2}$ and a time-dependent EnKF matrix $\hat C_t^{\theta,\phi}$ computed in \eqref{eq:C_enkf} with fixed covariance inflation and covariance tapering,\footnote{The covariance inflation for the EnKF component of the Ens3DVar covariance can additionally be learned, but we do not pursue this task in our experiments.}
where the weighting is controlled by a scalar parameter $\phi_1\in \mathbb{R}$,
\begin{equation}
    \tilde C^{\theta,\phi}_t = (1-\phi_1)C_{\phi_2} + \phi_1 \hat C_t^{\theta},\label{eq:C_ens3dvar}
\end{equation}
where $\phi = \{\phi_1,\phi_2\}$.   

We use a generalized notation $C_t^{\theta,\phi}$ for the forecast covariance matrix, where $C_t^{\theta,\phi}:=\hat C_t^{\theta,\phi}$ for EnKF, $C_t^{\theta,\phi} := \tilde C_t^{\theta,\phi}$ for Ens3DVar, and $C_t^{\theta,\phi} := C_\phi$ for 3DVar. This general notation for the forecast covariance matrix $C_t^{\theta,\phi}$ is used throughout our description of auto-differentiable filtering algorithms in Section \ref{sec:ad-filtering}. 

\paragraph{Observation perturbation $\mathcal{Y}$.} For ensemble based methods, observation perturbation when computing innovations can help to mitigate the underestimation of the analysis error covariance, thereby discouraging ensemble collapse \citep{Evensen1994, Houtekamer1998} and ensuring convergence to the Kalman filter with a large ensemble size in linear-Gaussian settings  \citep{furrer2007estimation,al2024non}. Therefore, the ensemble-based methods EnKF and Ens3DVar add a stochastic perturbation to the observations based on $\mathcal{N}(0,R_t)$ distributed noise. This step is omitted in the 3DVar algorithm, as shown in Table \ref{tab:filtering_algs}, since this algorithm does not rely on an ensemble for covariance estimation. 

\paragraph{Initial conditions $x_0^{1:N}$.} The specification of the ensemble of initial conditions $x^{1:N}_0$ is problem specific, and the quality of this initial condition depends on how much is known about the state of the system at initialization. An initial condition $x_0$ can be specified as a ``best guess" at $t=0$, which could include interpolated observations $y_0$. Alternatively, in the absence of detailed information at initialization, an arbitrary initialization can be specified, in which case the spin-up time for the filter to produce reasonable analyses increases. An ensemble can then be created by perturbing this initial condition with random noise distributed with initial covariance $C_0$, which can be arbitrarily specified or based on apriori knowledge of the system.

\subsection{Auto-differentiable filtering}\label{sec:ad-filtering}

With the filtering algorithm defined in \eqref{eq:filter}, we can learn the dynamics parameters $\theta$ and filtering parameters $\phi$ via auto-differentiation, then estimate states by using a filtering algorithm with these learned parameters.
Algorithm \ref{alg:ad-filter} specifies this joint learning and filtering task. This algorithm outlines a three-step procedure where (a) gradient information with respect to the learnable parameters $\theta$ and $\phi$ is tracked through the filtering algorithm, (b) a theoretically motivated loss function computes the discrepancy in the observations and the analysis means, and (c) gradient information computed based on our choice of loss function informs gradient descent on the learnable parameters. Steps (a)-(c) are repeated until convergence of the parameters $\theta$ and $\phi$. At each iteration $k$ of the optimization, state estimates $x_{1:T}^{1:N,(k)}$ are constructed using the filtering  algorithm in \eqref{eq:filter}. The final state estimates $x_{1:T}^{1:N}$ are constructed by filtering with the converged $\hat \theta$ and $\hat \phi$.  

\begin{algorithm}[ht]
\caption{Auto-differentiable filtering}\label{alg:ad-filter}
\begin{algorithmic}
\State \textbf{Input:} Initial time samples $x_0^{1:N}$ with ensemble size $N$, observations $y_{1:T}$, initialized forecast model $\mathcal{M}_{\theta^{(0)}}(\cdot)$, initialized Kalman gain $\mathcal{K}_{\phi^{(0)}}(\cdot)$, observation perturbation model $\mathcal{Y}(\cdot; R_{1:T})$, observation matrices $H_{1:T}$
\While{$\theta,\phi$ are not converging:}
\State $x_{1:T}^{1:N,(k)}$ = \text{Filter}$\big(x_0^{1:N},y_{1:T}; \mathcal{M}_{\theta^{(k)}}(\cdot), \mathcal{K}_{\phi^{(k)}}(\cdot), \mathcal{Y}(\cdot; R_{1:T}), H_{1:T}\big)$ \Comment{Eq. \eqref{eq:filter}}
\State $\mathcal{L}(\theta^{(k)},\phi^{(k)}) = -\sum_{t=1}^T \log \mathcal{N}(y_t; H_t\hat m_t^{\theta^{(k)}}, H_t C_t^{\theta^{(k)},\phi^{(k)}}H_t^\top + R_t)$
\State $\theta^{(k+1)} = \theta^{(k)} - \alpha_{1,k}\nabla_\theta\mathcal{L}(\theta^{(k)}, \phi^{(k)})$
\State $\phi^{(k+1)} = \phi^{(k)} - \alpha_{2,k}\nabla_\phi\mathcal{L}(\theta^{(k)}, \phi^{(k)})$
\State $k \leftarrow k+1$
\EndWhile
\State \textbf{Output:} Analysis estimates $x_{1:T}^{1:N}$, learned forecast parameters $\hat \theta$, learned analysis parameters $\hat \phi$
\end{algorithmic}
\end{algorithm}

Algorithm $\ref{alg:ad-filter}$ includes prespecified optimization hyperparameters $\alpha_{1,k}$ and $\alpha_{2,k}$, which correspond to the learning rates for $\theta$ and $\phi$, respectively, at iteration $k$. For each iteration of the algorithm, the same set of initial conditions $x_0^{1:N}$ is used.

\paragraph{Loss function.} We define our loss function $\mathcal{L}(\theta,\phi)$ using similar motivation as in \citet{Chen2022} and \citet{Bach2025book} (Theorem 8.12). Specifically, we aim to maximize the log-likelihood function of the observations $p(y_{1:T}|\theta,\phi)$ with respect to the parameters $\theta$ and $\phi$, which can be written as 
\begin{equation}
    \log p(y_{1:T}|\theta,\phi) = \sum_{t=1}^T \log p(y_t | y_{1:t-1}, \theta, \phi) \label{eq:forecast_LL}
\end{equation}
based on the law of total probability, setting $y_{1:0} = 0$ by convention.  Using the observation process model, $y_t = H_tx_t + \eta_t$, we can create a Gaussian ansatz of the distribution $p(y_t|y_{1:t-1}, \theta, \phi)$ by estimating the corresponding mean and variance as follows:

\begin{equation}
    \mathbb{E}(y_t | y_{1:t-1}, \theta, \phi) = \mathbb{E}(H_tx_t + \eta_t|y_{1:t-1},\theta,\phi) \approx H_t \hat m_t^\theta,
\end{equation}
and 
\begin{equation}
    \text{Cov}(y_t|y_{1:t-1},\theta,\phi) = \text{Cov}(H_tx_t + \eta_t|y_{1:t-1},\theta,\phi) \approx H_tC_t^{\theta,\phi}H_t^T + R_t:=S_t, \label{eq:S_t}
\end{equation}
where $\hat m_t^\theta$ and $C_t^{\theta,\phi}$ are the estimated forecast mean and forecast error covariance at time $t$, respectively. These estimated components allow us to construct the following Gaussian approximation
\begin{equation}
    \log p(y_{1:T}|\theta,\phi) \approx \sum_{t=1}^T \log \mathcal{N}(y_t;H_t \hat m_t^\theta, H_tC_t^{\theta,\phi}H_t^\top + R_t)=:-\mathcal{L}(\theta,\phi), \label{eq:LL_approx}
\end{equation}
which can be more explicitly written as 
\begin{equation}
    \mathcal{L}(\theta,\phi) := -\sum_{t=1}^T\left[ -\frac{1}{2}\log\det(S_t) - \frac{1}{2}\|y_t - H_t\hat m_t^\theta\|^2_{S_t}\right], \label{eq:loss}
\end{equation}
where $\mathcal{L}(\theta,\phi) = -\log p(y_{1:T}|\theta,\phi)$ up to an additive constant independent of $\theta$ and $\phi$. This function $\mathcal{L}(\theta,\phi)$ is our loss function.\footnote{We note that, depending on the application, one may consider adding regularization penalties to this loss to impose constraints on the learned $\theta$ and $\phi$. For instance, one may be interested in learning $\theta_1$ in a SINDy-style approach \citep{Brunton2016}, which would involve adding the penalization term $\lambda\|\theta_1\|_1$ with regularization parameter $\lambda$ to elicit sparsity. This approach is beyond the scope of the current work.}

Equation \eqref{eq:loss} provides a loss that allows us to (approximately) maximize the likelihood of our observations with respect to the forecast parameters $\theta$ and analysis parameters $\phi$ across the entire time series. We note that this derivation 
relies on the linearity of $H_t$. Moreover, the accuracy of the approximations in \eqref{eq:LL_approx} is contingent on the forecast distribution being well approximated with a Gaussian $ \mathcal{N}(\hat{m}_t^\theta, C_t^{\theta,\phi})$ distribution.

\paragraph{Optimization.} To (approximately) maximize $p(y_{1:T}|\theta,\phi)$ with respect to parameters $\theta$ and $\phi$, we consider the optimization problem
\begin{equation}
    \{\hat \theta, \hat \phi \} = \text{argmin}_{\theta,\phi} \mathcal{L}(\theta,\phi).
\end{equation}
The gradient descent steps in Algorithm \ref{alg:ad-filter} seek to solve this problem via gradient-based optimization. 

As is developed in \citet{Chen2022} for AD-EnKF, we similarly implement a training procedure for Algorithm \ref{alg:ad-filter} inspired by truncated backpropagation through time (TBPTT) \citep{Sutskever2014}. With this approach, updates to $\theta$ and $\phi$ are applied after iterating through a subsequence of $L$ assimilation steps, where $L \ll T$. This procedure results in $\lceil T/L \rceil$ gradient updates per epoch,\footnote{In the context of our offline auto-differentiable filtering framework, an epoch refers to filtering through and updating learnable parameters for all time points in the training data.} which can avoid the problem of vanishing or exploding gradients that can occur when auto-regressively applying learned models to time series data over long time horizons. TBPTT also aids in memory management since the computational graph only needs to be stored for a time series of length $L$ instead of the entire length $T$ time series, where $L\ll T$. However, truncating gradients introduces bias by systematically omitting dependencies longer than the truncation window, which impacts the trajectory of the overall optimization. This modified algorithm is shown in Algorithm \ref{alg:ad-filteringTBPTT} and includes the additional parameter $L$, which controls how often the parameters $\theta$ and $\phi$ are updated per epoch while training on the entire length $T$ time series.

\begin{algorithm}
\caption{Auto-differentiable filtering with TBPTT}\label{alg:ad-filteringTBPTT}
\begin{algorithmic}
\State \textbf{Input:} Initial time samples $x_0^{1:N}$ with ensemble size $N$, observations $y_{1:T}$, initialized forecast model $\mathcal{M}_{\theta^{(0)}}(\cdot)$, initialized Kalman gain $\mathcal{K}_{\phi^{(0)}}(\cdot)$, observation perturbation model $\mathcal{Y}(\cdot; R_{1:T})$, observation matrices $H_{1:T}$, number of time points to assimilate before backpropagation $L$
\While{$\theta,\phi$ are not converging:}
\For{$j =0,\dots,T/L-1$}
\State $t_0=jL$; \ $t_1=\text{min}\{(j+1)L,T\}$
\State $x_{t_0:t_1}^{1:N,(k)}$ = \text{Filter}$\big(x_{t_0}^{1:N},y_{t_0+1:t_1}; \mathcal{M}_{\theta^{(k)}}(\cdot), \mathcal{K}_{\phi^{(k)}}(\cdot), \mathcal{Y}(\cdot; R_{t_0+1:t_1}), H_{t_0+1:t_1}\big)$ \Comment{Eq. \eqref{eq:filter}}
\State $\mathcal{L}(\theta^{(k)},\phi^{(k)}) = -\sum_{t=t_0+1}^{t_1} \log \mathcal{N}(y_t; H_t\hat m_t^{\theta^{(k)}}, H_t C_t^{\theta^{(k)},\phi^{(k)}}H_t^\top + R_t)$ 
\State $\theta^{(k+1)} = \theta^{(k)} - \alpha_{1,k}\nabla_\theta\mathcal{L}(\theta^{(k)}, \phi^{(k)})$
\State $\phi^{(k+1)} = \phi^{(k)} - \alpha_{2,k}\nabla_\phi\mathcal{L}(\theta^{(k)}, \phi^{(k)})$
\State $k \leftarrow k+1$
\EndFor
\EndWhile
\State \textbf{Output:} Analysis estimates $x_{1:T}^{1:N}$, learned forecast parameters $\hat \theta$, learned analysis parameters $\hat \phi$
\end{algorithmic}
\end{algorithm}

We emphasize that while filtering is an \textit{online} algorithm, meaning that the filtering algorithms can be deployed as soon as new observations arrive, we formulate the optimization problem for learning $\theta$ and $\phi$ in an \textit{offline} fashion, meaning the observations need to be collected in a batch, after which Algorithm \ref{alg:ad-filter} is deployed.

We note that joint identifiability of $\theta$ and $\phi$ is not guaranteed. Errors in the forecast dynamics may be partly compensated by covariance and inflation choices; however, covariance factors are also non-unique. For example, covariance factors $B$ and $\tilde B = BO$ give the same covariance $C$ for an orthogonal $O$ since $C = BB^\top = \tilde B\tilde B^\top$ for $B\neq \tilde B$. Therefore, adequate filtering performance does not establish unique parameter recovery.

\paragraph{A note on necessary conditions.} We now emphasize a few necessary conditions for a method to fall within the class of auto-differentiable filtering algorithms that we define here. First and foremost, the algorithm used to compute state estimates $x_{1:T}$ must depend on current and past observations, so $x_t$ must only be informed by observations $y_{1:t}$. Second, we emphasize that the specifications of the components in \eqref{eq:filter} are flexible enough to handle time-varying $H_t$ in order to encompass a wide range of applied settings. Lastly, the filtering algorithm must provide a means to estimate the covariance in \eqref{eq:S_t} to compute the theoretically-motivated loss function in \eqref{eq:loss}. If a particular filtering algorithm does not satisfy any one of these criteria, it is not considered to be a part of this framework.

\subsection{Computational considerations} \label{sec:computation_memory}

In this section, we provide computational considerations for this framework, and particularly emphasize the computational and memory differences for learning via ensemble-based filtering algorithms (e.g., EnKF and Ens3DVar) compared to ensemble-free filtering algorithms  (e.g., 3DVar). We discuss the computational complexity in terms of particular filtering algorithms themselves since the filtering algorithm is the main driver of computational cost; the cost of auto-differentiation is proportional to the computational complexity of the filtering process up to a constant factor \citep{Griewank2008, Baydin2018}.

\paragraph{Computational costs.}

The computational costs of the forecast and analysis in one filtering step is given in Table \ref{tab:comp_cost} for 3DVar-$C$, EnKF, and Ens3DVar. In 3DVar-$C$ with our specification of $C_\phi = B_\phi B_\phi^\top$, where $B_\phi \in \mathbb{R}^{d_x\times d_x}$ and $H_t$ is linear, the computational cost of a 3DVar-$C$ analysis step is $\mathcal{O}(d_y^3 + d_y^2d_x + d_y d_x^2)$.
If $H_t\equiv H$, meaning the observed states remain constant across time, the Kalman gain $C_\phi H^T(HC_\phi H^T + R)^{-1}$ only needs to be computed once per fixed $\phi$, after which the computational cost for each 3DVar-$C$ analysis step becomes $\mathcal{O}(d_yd_x)$. Denoting by $c_\mathcal{F}$ the computational cost of a forecast from $\mathcal{F}_{\theta_1},$ since 3DVar-$C$ relies on a single particle, the cost of the forward pass would be $\mathcal{O}(c_\mathcal{F})$.

As noted in \citet{Mandel2006}, the computational complexity of a straight-forward implementation of an analysis step of EnKF is $\mathcal{O}(d_y^3 +d_y^2N +d_yN^2 + d_xN^2)$, where the dominating term depends on the relative values of $d_x,$ $N$, and $d_y$, but in many applications, $d_x\gg d_y \gg N$. 
For the computational cost of alternative implementations of the EnKF analysis step, we refer to \cite{tippett2003ensemble}. The computational cost for forecasting the ensemble of $N$ total members is $\mathcal{O}(Nc_\mathcal{F})$.

For Ens3DVar, the computational cost depends on the dominating terms of 3DVar-$C$ and EnKF, which gives $\mathcal{O}(d_y^3 + d_y^2d_x + d_x^2d_y)$ assuming that $N\ll d_y$. For the forecast step, Ens3DVar incurs a computational cost of $\mathcal{O}(Nc_\mathcal{F})$ for propagating the ensemble of size $N$ with the forecast model.

In many applications, the cost of a filtering step is usually dominated by the computational cost of evaluating the forward model, which is amplified by a factor of $N$ for ensemble-based methods when performed serially. However, this cost can be reduced via parallelization. 

\begin{table}[ht]
    \centering
    \begin{tabular}{ccc}\toprule
          &   Forecast cost& Analysis cost \\ \midrule
         3DVar-$C$ & $c_\mathcal{F}$& $\mathcal{O}(d_y^3 + d_x^2d_y + d_y^2d_x)$\\\midrule
          EnKF &$Nc_\mathcal{F}$&$\mathcal{O}(d_y^3 +d_y^2N +d_yN^2 + d_xN^2)$\\\midrule
          Ens3DVar &$Nc_\mathcal{F}$& $\mathcal{O}(d_y^3 + d_x^2d_y + d_y^2d_x)$\\
          \bottomrule 
    \end{tabular}
    \caption{Table of computational costs for the forecast and analysis in one filtering step for straight-forward implementations of the algorithms 3DVar-$C$, EnKF, and Ens3DVar, where we assume that $N \ll d_y$ for Ens3DVar.}
    \label{tab:comp_cost}
\end{table}

\paragraph{Memory costs.} One aspect of auto-differentiable filtering that substantially impacts the memory footprint, as additionally noted in \citet{Chen2022}, is the length of the time series $T$, or $L$ if training with TBPTT. The computational graph needs to be stored as the filter sequentially processes data, and memory can only be cleared after backpropagating through the time series (for $T$ steps, or $L$ steps if using TBPTT) and updating the learnable parameters. When using TBPTT, choices of large $L$ can lead to substantial memory consumption; on the other hand, too small $L$ can bias training by ignoring potentially important longer-range dependencies \citep{Aicher2020}. This choice requires empirical tuning.

For ensemble-based methods, another choice that can substantially impact the memory footprint is the number of ensemble members $N$.  The memory footprint of these ensemble-based methods increases by a factor of $N$ compared to single-particle filters, but this increased memory footprint as $N$ increases generally comes with the benefit of better learning, as shown for AD-EnKF in \citet{Chen2022}. For high-dimensional systems, storing large ensembles, tracking the trajectories through a long time sequence, and backpropagating to update the learnable parameters can easily exceed memory limits if memory is not precisely considered.

\section{Related work}\label{sec:related_work}

There is a vast body of literature developed in recent years regarding machine learning in data assimilation. \citet{Bach2025book} provides a thorough review of the current state of machine learning for inverse problems and data assimilation. We highlight particular works related to our learning goal, including papers that aim to learn forecast model parameters, learn analysis parameters, or jointly learn forecast and analysis parameters from noisy, potentially partial observations. 

\paragraph{Learning forecast parameters from observations.} Numerous previous works have proposed various approaches to updating a forecast model based on observations via data assimilation: to name a few, \citet{Brajard2020} and \citet{Chen2022} (AD-EnKF) use an ensemble Kalman filter and auto-differentiation to learn a surrogate model; \citet{Gottwald2021} learns a forecast model based on sparse information using a random feature map and EnKF data assimilation; \citet{Bocquet2021} studies joint estimation of dynamics and states via EnKF, comparing online and offline learning with sparse observations; and \citet{Farchi2021} corrects model error via 4DVar data assimilation. Additionally, \citet{Wang2025} presents EnKF and particle filter variants that allow for forecast parameter learning via formulating the problem as a discrete time Markov decision process from a reinforcement learning perspective. \citet{Chen2023} proposes to learn a reduced order model that performs data assimilation in a latent space, where a decoder then maps latent states back to the original state space, thereby learning a low-dimensional, computationally efficient filter.  
Though less frequently used, auto-differentiable particle filters have also been developed for learning forecast components, such as in \citet{Scibior2021}; we refer to \citet{Chen2025} for a review of these methods.

\paragraph{Learning analysis parameters from observations.} Recent works have explored learning portions of the analysis step in data assimilation methods while keeping the dynamics mapping fixed. \citet{Melinc2024} learns an autoencoder to perform a 3DVar update in a reduced space, thereby allowing for faster computation. \citet{Bach2025} learns a Kalman gain for variational filtering problems. \citet{Lu2025} uses machine learning to learn improved localized error covariance matrices within EnKF. \citet{Revach2022} learns a Kalman-type update from partial, noisy observations that does not rely on  potentially limiting assumptions of Gaussian error statistics that are assumed in traditional data assimilation. \citet{Bach2026} learns ensemble filtering algorithms via a mean-field state-space formulation of filtering. Methods that retain the traditional Gaussianity assumptions on the error statistics and aim to learn components of the analysis could be integrated into the framework we propose in this work.

\paragraph{Joint learning of forecast and analysis parameters from observations.} \citet{Chinellato2025} propose an algorithm based on the Kalman filter, where Kalman gain-type matrices are learned for each filter step along with the dynamics mapping. Additionally, that work explores regularization on the dynamics mapping via SINDy \citep{Brunton2016}. \citet{Lunderman2021} jointly learns analysis and forecast parameters via derivative-free global Bayesian optimization for offline learning from observations. Another algorithm that achieves the goal of jointly learning forecast and analysis parameters is \citet{Levine2022}. We provide a detailed discussion of this method below.

\paragraph{A comparable algorithm: AD-3DVar-$K$.} We now describe in detail a method we will refer to as AD-3DVar-$K$ that is proposed in \citet{Levine2022}. This work aims to learn forecast model errors via auto-differentiation through a 3DVar-type algorithm. Since this algorithm achieves the same overarching goals as our general learning framework (learning states, forecast and filtering parameters), we include this method in our experimental results as a point of comparison. However, this method departs from the framework we outline in Section \ref{sec:method_learning} in nontrivial ways, which we detail here. Most notably, this method is unable to handle observations that come from a time-varying $H_t$, which limits its use in many practical settings.

AD-3DVar-$K$ specifies the 3DVar filtering algorithm (denoted as 3DVar-$K$ in this work) as
\begin{equation}
    x_t = (I - K_\phi H)\mathcal{F}_{\theta_1}(x_{t-1})+K_\phi y_t, \label{eq:3dvark}
\end{equation}
where $K_\phi \in \mathbb{R}^{d_x \times d_y}$ is a Kalman gain matrix directly parameterized by $\phi$ (i.e., this matrix is not computed using information in $H$, $R$ or some choice of $C$). We also note the subscript $t$ on $d_y$ and $H$ has been deliberately dropped because this method only applies to settings where 
the observation matrix $H$ is fixed
across the entire assimilation window.

AD-3DVar-$K$ outlines an algorithm of the same form as Algorithm $\ref{alg:ad-filter}$, except a key distinction is in the choice of loss function $\mathcal{L}(\theta,\phi)$. Since the Kalman gain matrix $K_\phi$ is learned directly, the forecast covariance in the observation space $(HC^{\theta,\phi}H^\top + R)$ is not directly available, and therefore, the loss derived in \eqref{eq:loss} cannot be directly computed. Instead, the loss is defined as 
\begin{equation}
    \mathcal{L}_{\text{3DVar-$K$}}(\theta,\phi) := \frac{1}{T}\sum_{t=1}^T\|y_t - H\mathcal{F}_{\theta_1}(x_{t-1})\|^2, \label{eq:loss_ad3dvark}
\end{equation}
which implicitly assumes that the residuals across all components carry equal variance, therefore that $HCH^\top + R$ is a multiple of the identity matrix. Gradient information is similarly tracked with respect to $\theta$ and $\phi$ throughout the time series, where $\theta = \theta_1$, and gradient descent is performed to minimize this loss. Though $\phi$ does not explicitly appear in \eqref{eq:loss_ad3dvark}, $\phi$ implicitly appears in $x_{t-1}$ since it is constructed using \eqref{eq:3dvark}. The optimization problem can be written as 
\begin{equation}
    \{\hat\theta,\hat\phi\} = \text{argmin}_{\theta,\phi}\mathcal{L}_{\text{3DVar-$K$}}(\theta,\phi),
\end{equation}
where gradient-descent optimization steps of a similar form: $\theta^{(k+1)} = \theta^{(k)} -\alpha_{1,k}\nabla_\theta \mathcal{L}(\theta^{(k)}, \phi^{(k)})$ and $\phi^{(k+1)} = \phi^{(k)} -\alpha_{1,k}\nabla_\phi \mathcal{L}(\theta^{(k)}, \phi^{(k)})$ for iterations $k=1,\dots$ are taken until convergence. Once convergence is achieved, state estimates $x_{1:T}$ are computed by filtering with the learned $\hat \theta$ and $\hat \phi$ parameters.

We now briefly discuss the computational demands of this approach
compared to the particular methods discussed in Section \ref{sec:computation_memory} that fall into our framework.  Among the particular methods mentioned, AD-3DVar-$K$ is the most computationally- and memory-efficient. Since AD-3DVar-$K$ is ensemble-free and directly learns a matrix $K_\phi\in \mathbb{R}^{d_x\times d_y}$, this approach avoids the computationally intensive $d_y\times d_y$ matrix inversion in AD-3DVar-$C$ at each time point $t$, and additionally only stores the Kalman gain matrix, which contains $(d_x-d_y)d_x$ fewer elements than a dense $C_\phi$ matrix for $d_y\ll d_x$. 
The overall computational complexity of each analysis step is $\mathcal{O}(d_yd_x)$ and $\mathcal{O}(c_\mathcal{F})$ for each forecast.

\section{Experiments}\label{sec:experiments}
Our experiments aim to explore our framework across multiple scientific settings with different learning goals for the forecast component, namely learning residual neural networks and parameter estimation to improve forecast performance. We evaluate the forecast performance of $\hat\theta_1$ and the filtering performance using the learned $\hat\theta$ and $\hat\phi$ across four methods: AD-3DVar-$C$, AD-3DVar-$K$, AD-EnKF modified to additionally learn the covariance inflation factor,\footnote{Learning the covariance inflation factor in \eqref{eq:C_enkf} corresponds to the modification we make to the algorithm presented in \citet{Chen2022}; therefore, in our experiments, we label this approach as ``AD-EnKF (Modified)." The tapering matrix $\rho$ in \eqref{eq:C_enkf} could additionally be learned, but doing so can be computationally expensive. We further discuss this point in Appendix \ref{app:learning_rho}.} and AD-Ens3DVar. In these evaluations, we hold the parameters fixed at the learned values, $\hat \theta$ and $\hat \phi$, where evaluations are performed on held-out observations. Section \ref{sec:CW}, motivated by a simplified problem in astrodynamics, explores learning from observations of a partially observed linear dynamical system. Experimenting on a system with linear dynamics and linear observations allows us to compare how well each method approximates the optimal solution given by the Kalman filter when using the true forecast model parameters. Section \ref{sec:L96}, motivated by applications in the atmospheric sciences, illustrates learning a residual neural network in a nonlinear dynamics setting with a \textit{static} observation matrix $H.$ Section \ref{sec:GLV}, motivated by problems in population dynamics, illustrates that methods that fall under our framework are able to learn from partial observations where the observation matrix and its dimension may vary across time. Lastly, Section \ref{sec:QG}, motivated by settings in geophysical fluid dynamics, explores the scalability of the methodology outlined in this paper by operating on a high-dimensional state vector.

For all experiments, we visualize the performance of the learned forecast model parameters $\hat \theta_1$ across a variety of values of the problem (e.g., error level $\sigma_0^2$ in the initialized model, state observability ratio $d_y/d_x$, training data size $T$, etc.).\footnote{An additional comparison where the ensemble sizes $N$ varies for the methods AD-EnKF (Modified) and AD-Ens3DVar is provided in Appendix \ref{app:enkf_vs_ens3dvar}.} We additionally evaluate the performance of the learned components $\hat\theta$ and $\hat\phi$ in a filtering task for assimilating unseen observation data for each algorithm. Evaluation metrics used in these experiments are defined in Appendix \ref{app:eval_metrics}, and additional training details for each example, including a discussion of hyperparameter tuning for the optimization learning rates, are listed in Appendix \ref{app:training_details}.  In each experiment, the forecast model $\mathcal{F}_{\theta_1}$ in \eqref{eq:hmm_forecast} corresponds to the $\Delta t$-flow of each considered differential equation, where $\Delta t$ represents the time between observations. Additionally, we experiment on deterministic dynamical systems, where the true forecast error parameters are $Q_{\theta_2^*} = \mathbf{0}_{d_x \times d_x}$ when forecasting with the true mapping $\mathcal{F}_{\theta_1^*}$.

\subsection{Astrodynamics: Clohessy–Wiltshire equations}\label{sec:CW}

The Clohessy-Wiltshire equations \citep{Clohessy1960}, built upon work in \citet{Hill1878}, are a simplified model for orbital motion, where a ``target" object is in a circular orbit and a ``chaser" object is in either an elliptical or circular orbit. These equations are the basis for planning orbital rendezvous of a spacecraft with another object. The equations are 
\begin{align}
    \frac{d^2x_1}{dt^2} &= 3 \theta^2x_1 + 2\theta \frac{dx_2}{dt}, \label{eq:CW_d2xdt2} \\
    \frac{d^2x_2}{dt^2} &= -2\theta \frac{dx_1}{dt}, \\
    \frac{d^2x_3}{dt^2} &= - \theta^2 x_3, \label{eq:CW_d2zdt2}
\end{align}
\noindent where $\theta_1 \in \mathbb{R}_+$ is the orbital rate of the target, $[x_1,x_2,x_3]$ is the position vector, $\left[\frac{dx_1}{dt},\frac{dx_2}{dt},\frac{dx_3}{dt}\right]$ is the velocity vector, and $\left[\frac{d^2x_1}{dt^2},\frac{d^2x_2}{dt^2},\frac{d^2x_3}{dt^2}\right]$ is the acceleration vector of the chaser object relative to the target. The goal 
is to construct estimates of a state vector $x = \left[x_1,x_2,x_3,\frac{dx_1}{dt},\frac{dx_2}{dt}, \frac{dx_3}{dt}\right]$ across time.\footnote{Here and in the following examples, $x_i$ denotes the $i$-th component of the state vector $x$ rather than the assimilation time.}

\subsubsection{Learning task}

This experiment is formulated as a simple parameter estimation problem where the goal is to recover a true scalar $\theta_1^*:=0.0013$ based on only noisily observing the position vector $[x_1,x_2,x_3]$ across time, therefore $H=I_{3\times6}$ is a time-independent identity matrix with rows deleted that correspond to the velocity state components. Initial guesses $\theta_1^{(0)}$ are sampled from $\mathcal{N}(\theta_1^*,\sigma_0^2)$, where $\sigma_0^2\in\mathbb{R}$ controls the variance of the noise for our initial guess for the parameter. This $\sigma^2_0$ parameter is varied in some experiments. 

\subsubsection{Results}

\begin{figure}
    \centering
    \begin{subfigure}{0.35\textwidth}
        \centering
        \includegraphics[width=\textwidth]{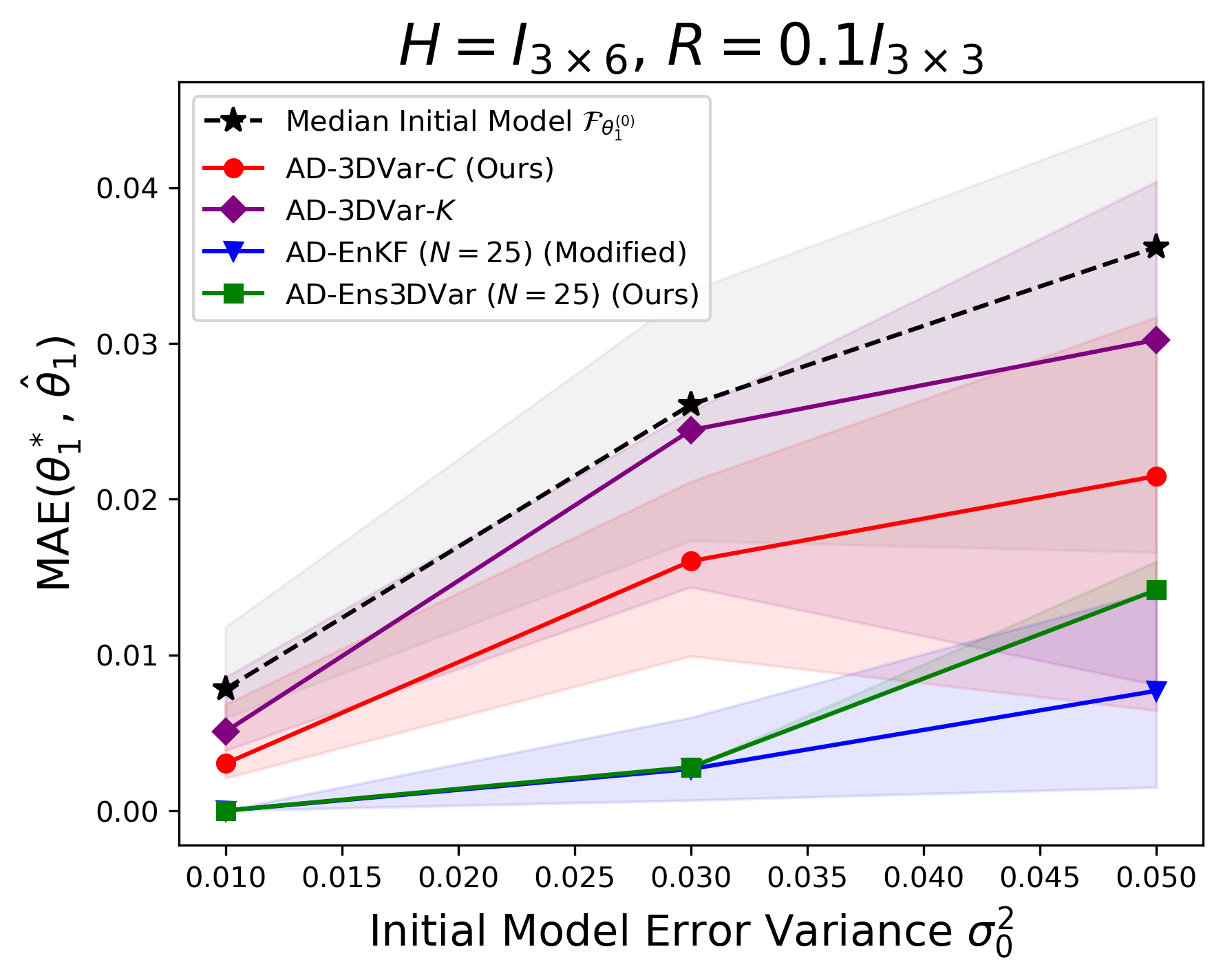}
        \caption{Parameter estimation error across choices of error level $\sigma_0^2$ in the initialized models, where $H=I_{3\times 6}$ and $R=0.1I_{3\times3}$.}\label{subfig:forecast_dx}
    \end{subfigure}
    ~
    \begin{subfigure}{0.35\textwidth}
        \centering
        \includegraphics[width=\textwidth]{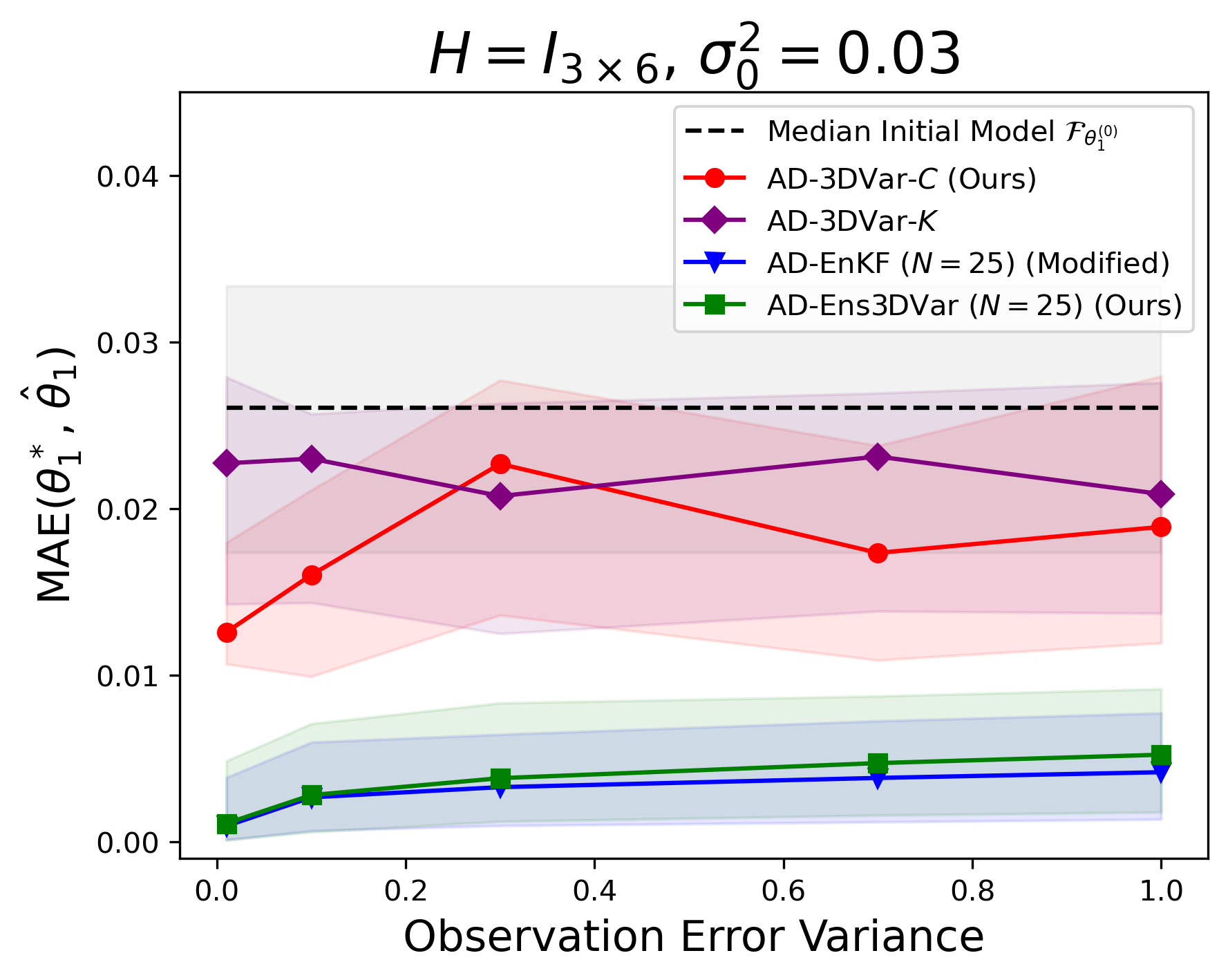}
        \caption{Parameter estimation error across observation error variance, where $H=I_{3\times 6}$ and $\sigma_0^2=0.03$.} \label{subfig:forecast_noise}
    \end{subfigure}

    \caption{Clohessy-Wiltshire parameter estimation (Section \ref{sec:CW}). Parameter estimation error of the learned forecast parameter $\hat \theta_1$ compared to the ground truth parameter $\theta_1^*=0.0013$ across choices of $\sigma^2_0$ and variance on the diagonals of $R$. The shaded regions correspond to the 0.25 and 0.75 quantiles of errors across 10 independent simulations. We additionally plot the median error for the initialized parameters $\{\theta_{1,j}^{(0)}\}_{j=1}^J$ for $j=1,\dots,J=10$ in the dashed black line representing the errors in the initialized forecast parameter $\theta_1^{(0)}$, and the gray region corresponds to the 0.25 and 0.75 quantiles of the parameter MAEs across these 10 initialized models.}
    \label{fig:CW_param_estim}
\end{figure}

\begin{figure}
\centering
\includegraphics[width=\textwidth]{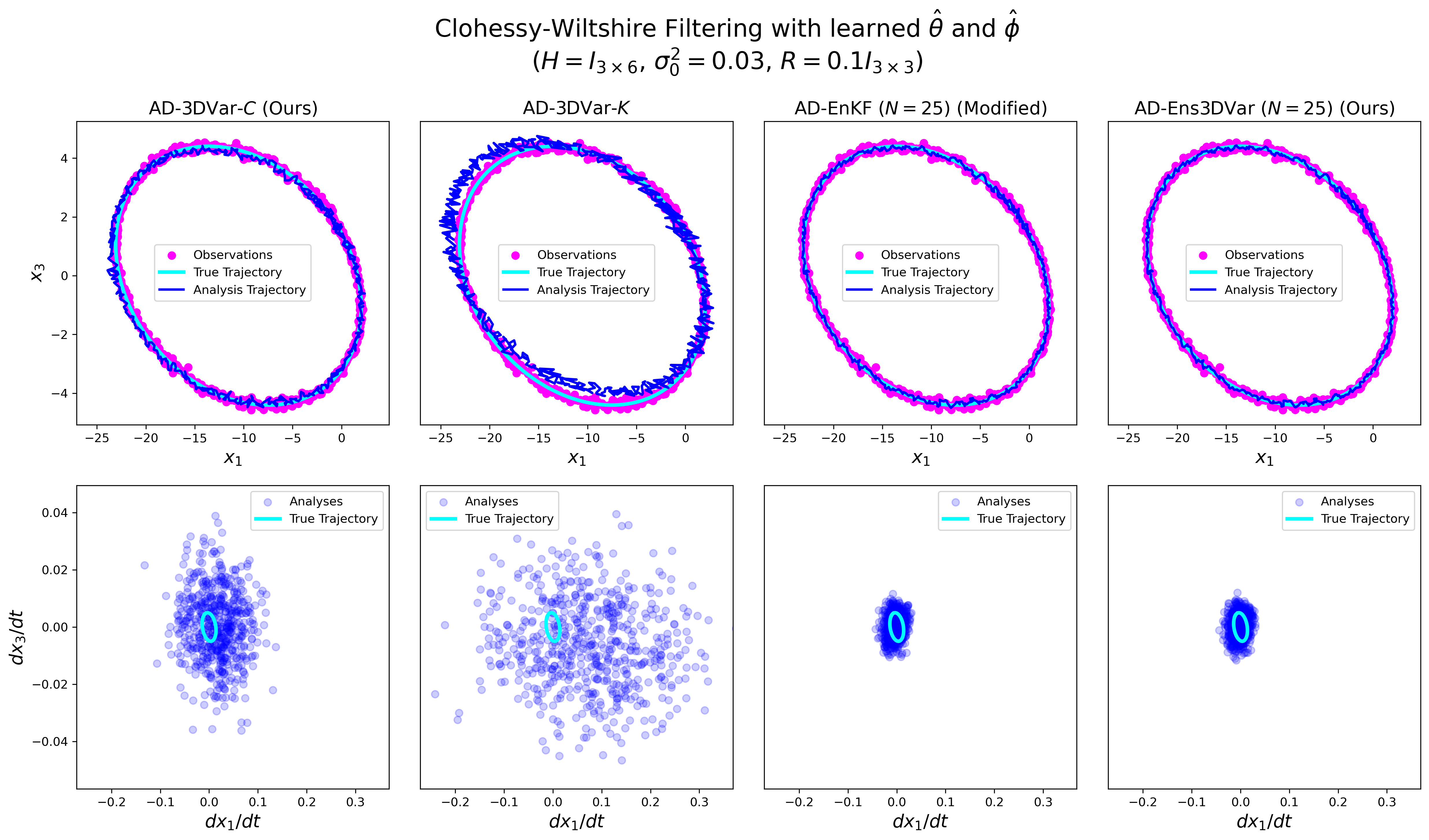}
\caption{Clohessy-Wiltshire filtering (Section \ref{sec:CW}). Visualization of observations, true trajectories, and estimated positions for $x_1$ and $x_3$, and velocities $\frac{dx_1}{dt}$ and $\frac{dx_3}{dt}$ across each of the four filtering methods using the learned forecast parameters $\hat \theta$ and learned filtering parameters $\hat \phi$, evaluated on test data. In this example, the observation noise is fixed at $R = 0.1I_{d_y\times d_y}$, and the variance of the initialization of $\theta_1^{(0)}$ is fixed at $\sigma_0^2 = 0.03$.}
\label{fig:CW_filtering}
\end{figure}

Data generation and training details for this example can be found in Appendix \ref{app:CW_training_LL}. Optimization hyperparameters were tuned for each simulation and parameter value across the $J=10$ independent simulated initial models. The time step between observations for this example is $\Delta t=10$.

\paragraph{Performance in learning forecast parameters $\hat \theta_1$.} Figure \ref{fig:CW_param_estim} shows the forecast performance using the estimated $\hat\theta_1$ across various parameter settings, namely the noise level $\sigma^2_0$ in the initialized $\theta_1^{(0)}$ and the observation noise variance. The ensemble-based methods AD-EnKF and AD-Ens3DVar achieve low MAE errors in recovering the true parameter $\theta_1^*$ across all choices of $R$ and $\sigma_0^2$. On the other hand, the AD-3DVar algorithms achieve less significant improvements to the initial imperfect $\theta_1^{(0)}$, as evidenced by the fact that the quantile spread of these two methods overlap with the quantile spread of the collection of initialized models.

\paragraph{Filtering performance using the learned $\hat \theta$ and $\hat \phi$.}  Figure \ref{fig:CW_filtering} visualizes the filtering performance using the learned forecast parameters $\hat \theta$ and filtering parameters $\hat \phi$ across each of the four methods in the setting where $H=I_{3 \times 6}$, $\sigma^2_0=0.03$, and $R=0.1I_{3\times 3}$ for one randomly initialized model. The ensemble-based methods AD-EnKF and AD-Ens3DVar provide the best filtering performance, and show significantly more precise estimation of the unobserved velocity state components $\frac{dx_1}{dt}$ and $\frac{dx_3}{dt}$ compared to both AD-3DVar-$C$ and AD-3DVar-$K$. AD-3DVar-$C$ provides an accurate estimate of the observed components but struggles more to recover the velocity components. AD-3DVar-$K$, on the other hand, most notably struggles to estimate the states for both the position and velocity components. An additional visualization similar to Figure \ref{fig:CW_filtering} for positions $x_2$ and $x_3$ and velocities $\frac{dx_2}{dt}$ and $\frac{dx_3}{dt}$ across methods is shown in Figure \ref{fig:CW_filtering_yz} in Appendix \ref{app:CW_filtering_yz}, and similar conclusions are derived from this figure. An evaluation of the learned covariance inflation factor in AD-EnKF (Modified) and the learned covariance mixing parameter in AD-Ens3DVar for the Clohessy-Wiltshire equations can be found in Figure \ref{fig:learned_alpha_gamma} in Appendix \ref{app:training_details}.

\paragraph{Estimation of the true forecast log-likelihood.} Since our dynamics mapping and our observation function $H_t \equiv H$ are both linear, the  filtering distribution  
for this problem is given by 
the Kalman filter \citep{Kalman1960}. We utilize this fact to evaluate how well EnKF, Ens3DVar, and 3DVar with their corresponding learned parameters $\hat \theta$ and $\hat \phi$ estimate the  true forecast log-likelihood given by the Kalman filter (Eq. \eqref{eq:forecast_LL}) evaluated at observations. This comparison shows how close each method is to optimal in a simple linear setting. We provide implementation details of the Kalman filter in this setting in Appendix \ref{app:CW_Kalman_filter}. Figure \ref{fig:CW_LL} shows the approximation to the true log-likelihood based on 3DVar, EnKF, and Ens3DVar given the respective learned $\hat\theta$ and $\hat \phi$ from AD-3DVar-$C$, AD-EnKF, and AD-Ens3DVar compared to the log-likelihood given by the Kalman filter. We note that AD-3DVar-$K$ does not appear in this comparison since learning $K$ does not allow us to directly construct this log-likelihood. Consistent with the results presented in Figures \ref{fig:CW_param_estim} and \ref{fig:CW_filtering}, the ensemble-based approaches AD-EnKF and AD-Ens3DVar provide a better approximation to the true 
forecast log-likelihood, and AD-3DVar-$C$ provides a much noisier approximation to this quantity. Both components of this log-likelihood ($-\frac{1}{2}\log \det S_t$ and $-\frac{1}{2}\|y_t-H_t\hat m_t^\theta\|^2_{S_t}$) are individually visualized in Figure \ref{fig:CW_LL_components} and discussed in Appendix \ref{app:CW_Kalman_filter}.

\begin{figure}
\centering
\includegraphics[width=0.8\textwidth]{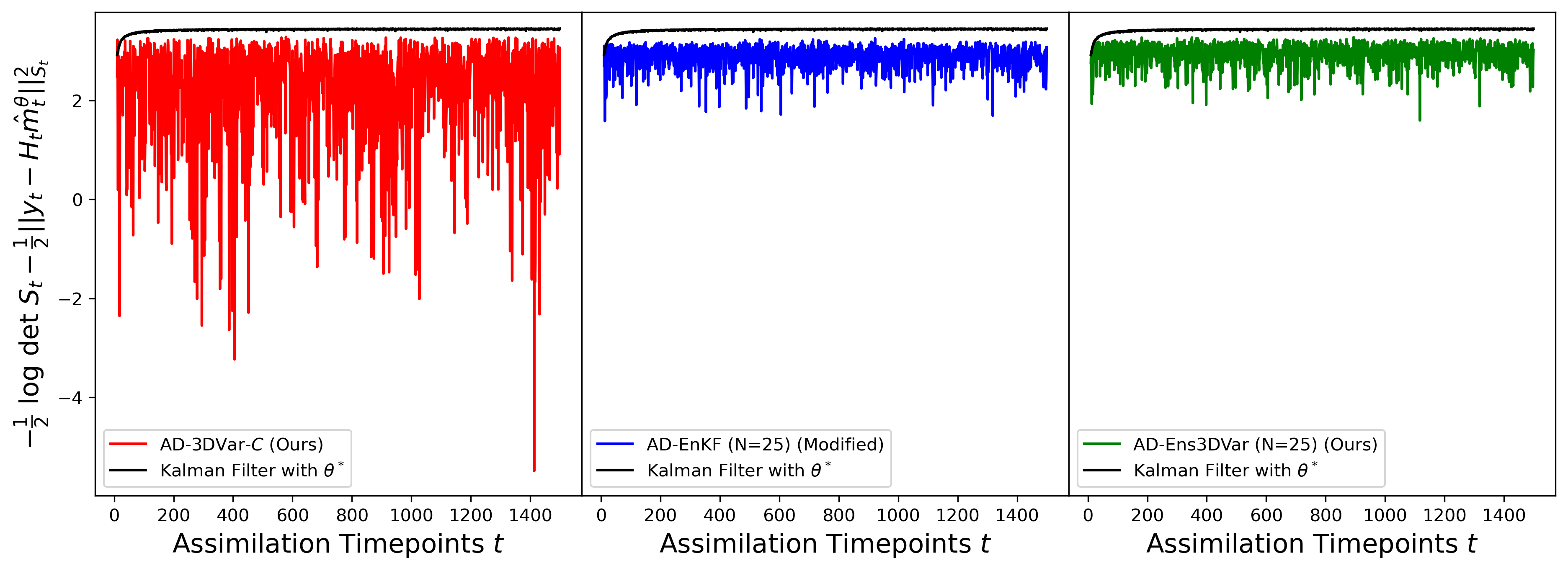}
\caption{Clohessy-Wiltshire data log-likelihood (Section \ref{sec:CW}). Plots of the forecast log-likelihood evaluated at observations from \eqref{eq:forecast_LL} for assimilating test observations into filters parameterized by the learned forecast parameters $\hat \theta$ and $\hat \phi$ learned from AD-3DVar-$C$, AD-EnKF ($N=25$), and AD-Ens3DVar ($N=25$). The algorithm used for filtering corresponds to the algorithm used in the learning task (i.e., EnKF is used as the filter with the learned parameters from AD-EnKF). The log-likelihood from the optimal solution to the filtering task, the Kalman filter (detailed in Algorithm \ref{alg:CW_KF}) using the true forecast parameters $\hat \theta$, is shown in each plot as a comparison.}
\label{fig:CW_LL}
\end{figure}

\subsection{Atmospheric dynamics: Lorenz-96}\label{sec:L96}

Next, we perform experiments on the Lorenz-96 system \citep{Lorenz1996}

\begin{equation}
    \frac{dx_i}{dt} = (x_{i+1} - x_{i-2})x_{i-1} - x_i + F, \label{eq:l96}
\end{equation}
\noindent where $x_i\in\mathbb{R}$, $i=1,\dots,d_x$ and $F$ is a forcing term that controls the chaotic behavior of the system, and we set $F=8$ to induce chaos. We additionally impose the standard periodic boundary conditions $x_{-1}:=x_{d_x-1}, x_{-2}:= x_{d_x-2}$, and $x_{d_x}:=x_0$ where $d_x \ge 4.$
The state vector $x_t\in\mathbb{R}^{d_x}$ corresponds to the solution to \eqref{eq:l96} at time $t+ \Delta t,$ where $\Delta t$ is the time between observations; we recall the convention we use in the examples, where $x_i$ denotes the $i$-th component of the state vector $x \in \mathbb{R}^{d_x}.$
This system is used as a testbed for data assimilation methods for atmospheric applications due to its observed chaotic behavior for certain parameterizations and its simplified approximation of nonlinear advection in weather patterns.

\subsubsection{Learning task}

For this example, we consider learning a neural network residual correction to an imperfect model $\mathcal{F}_0$. The forecast model is parameterized as 
\begin{equation}
    \mathcal{F}_{\theta_1} := \mathcal{F}_0 + \mathcal{G}_{\theta_1},\label{eq:f_theta}
\end{equation}
where $\mathcal{G}_{\theta_1}$ is a residual neural network parameterized by learnable weights $\theta_1$ that aim to correct $\mathcal{F}_0$'s predictions. This $\mathcal{F}_{\theta_1}$ model computes forecasts $\hat x_t = \mathcal{F}_0(x_{t-1}) + \mathcal{G}_{\theta_1}(x_{t-1})$ for any time $t$, where $x_t \in \mathbb{R}^{d_x}$.

The residual structure of $\mathcal{F}_{\theta_1}$ is motivated by its practical use in many settings where the aim is to blend physics-based and data-driven models. For example, $\mathcal{F}_0$ can be simplified partial differential equations, and $\mathcal{G}_{\theta_1}$ can be a learnable neural network, which is common to the task of learning closure models \citep{Sanderse2024}.

\paragraph{Imperfect initial model $\mathcal{F}_0$.} Given our goal to learn a model correction as specified in \eqref{eq:f_theta}, we first describe how we create imperfect initial models based on the Lorenz-96 system. To create an imperfect model $\mathcal{F}_0$, we take inspiration from experiments in \citep{Chen2022} and similarly define a basis consisting of 2-degree polynomials of the terms $[x_{i-2}, x_{i-1},x_{i}, x_{i+1}, x_{i+2}]$ to construct a differential equation of the form 
\begin{equation}
\begin{split}
    \mathcal{F}^{(i)}_\beta(x)=\frac{dx_i}{dt} = & [1, \  x_{i-2}, \ x_{i-1}, \ x_{i}, \ x_{i+1}, \ x_{i+2}, \\ 
     &x_{i-2}^2, \  x_{i-1}^2, \ x_i^2, \ x_{i+1}^2, \ x_{i+2}^2, \\
     & x_{i-2}x_{i-1}, \ x_{i-1}x_{i}, \ x_ix_{i+1}, \ x_{i+1}x_{i+2}, \
     x_{i-2}x_i, \ x_{i-1}x_{i+1}, \ x_ix_{i+2}]\beta,\label{eq:basis}
\end{split}
\end{equation}
for $i = 1,\dots ,d_x$, where $\beta\in\mathbb{R}^{18}$ are the coefficients of the basis polynomials. Equation \eqref{eq:l96} can be represented as \eqref{eq:basis} if $\beta = \beta^*$, where $\beta^*\in\mathbb{R}^{18}$ and $\beta_0^*=8, \ \beta_3^* = -1, \ \beta_{11}^*=-1, \ \beta_{16}^* = 1$, $\beta_m^*=0$ otherwise, and $m$ indexes the 18 total coefficients in the 2-degree polynomial library for the collection of terms $[x_{i-2},x_{i-1}, x_i, x_{i+1}, x_{i+2}]$.

We use the ground truth coefficients $\beta^*$ of the Lorenz-96 system to randomly generate noisily perturbed coefficients $\beta$ that are centered around the true parameter. More concretely, we specify $\mathcal{F}_0$ as 
\begin{equation}
    \mathcal{F}_0 = \{\mathcal{F}^{(i)}_\beta\text{ for }i=1,\dots,d_x\},  \ \text{ where } \  \beta_m \sim \begin{cases}\mathcal{N}(\beta_m^*, \sigma^2_0), \ \ \qquad  \text{ if  }m=0, \\ \mathcal{N}(\beta_m^*,0.1\sigma^2_0),  \ \quad \text{ if } m \in \{1,\dots, 5\}, \\ 
    \mathcal{N}(\beta_m^{*}, 0.01\sigma_0^2), \ \ \ \text{ if } m \in \{6,\dots, 17\},\end{cases} \label{eq:f_0}
\end{equation}
where $\sigma^2_0\in\mathbb{R}$ is the noise level of the perturbations. 

\paragraph{Residual model $\mathcal{G}_{\theta_1}$.} We define the structure of our residual model $\mathcal{G}_{\theta_1}$ from \eqref{eq:f_theta} to be a convolutional neural network with learnable parameters $\theta_1$. We specify $\mathcal{G}_{\theta_1}$ as in \citet{Chen2022}; see \citet{Chen2022} Figure 11 for details of the model correction network specification. The total number of parameters in the model $\mathcal{F}_{\theta_1}$ is 9,387, which does not depend on the state dimension $d_x$.

\subsubsection{Results}

Data generation and training details are described in Appendix \ref{app:details_L96}. We also note that in this experiment, the optimization hyperparameters used across all repeated simulations were selected from one simulation, so it is possible that closer to optimal hyperparameter sets could be chosen for each independent simulation. Across all Lorenz-96 experiments, the observation matrix $H_t$ is held fixed across all time points $t$, where observed components are deterministic. For example, every 3rd state is observed for $d_y/d_x = 0.3$. The time step used to solve the differential equation is $\Delta t = 0.05$, which corresponds to the same $\Delta t$ in which observations arrive. Our dataset consists of 1200 time points spaced with $\Delta t=0.05$. With a Lyapunov time of 0.6 time units, this time range corresponds to 60 time units.

\begin{figure}
    \centering
    \begin{subfigure}{0.32\textwidth}
        \centering
        \includegraphics[width=\textwidth]{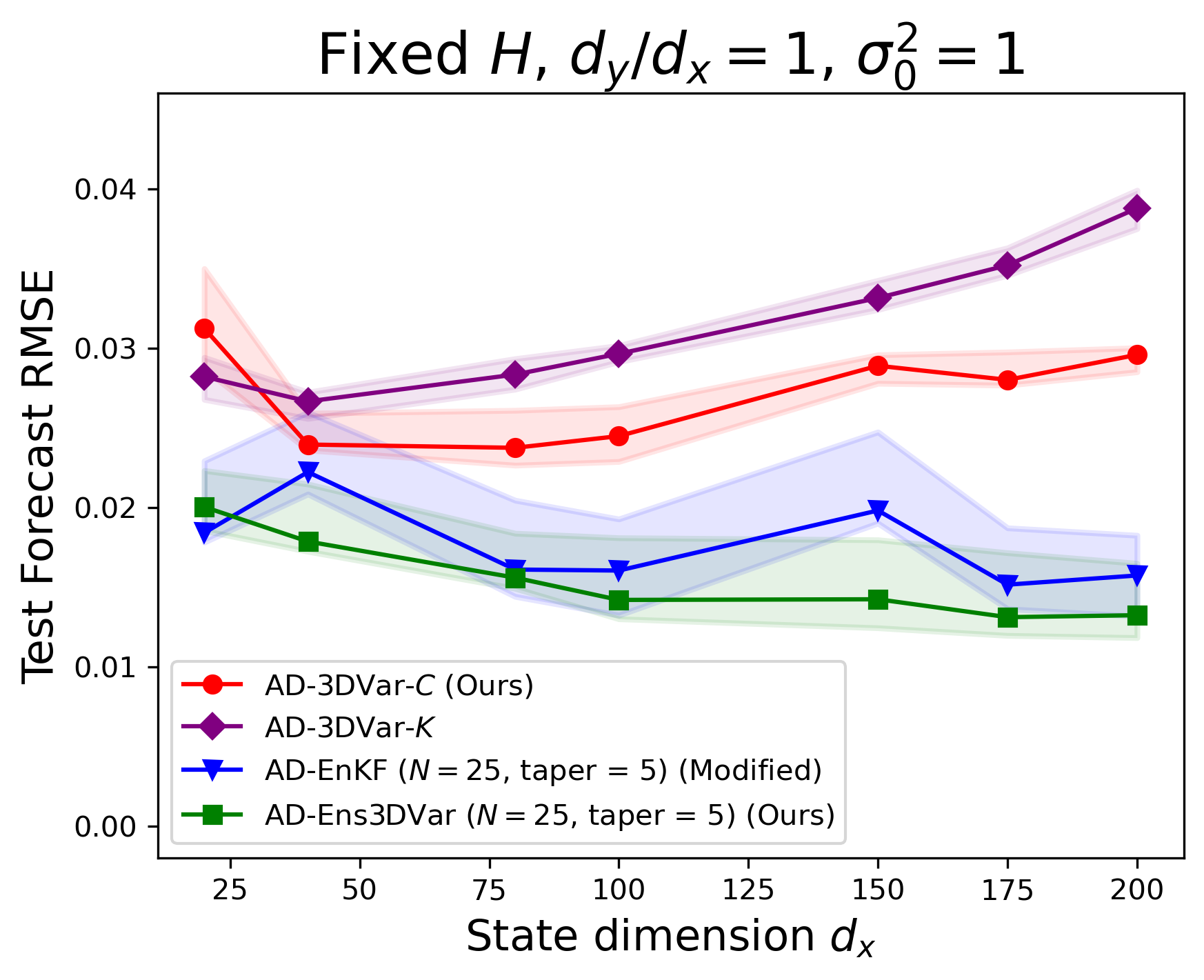}
        \caption{Test forecast performance of the model $\mathcal{F}_{\theta_1}$ across state dimensionalities $d_x$, where $H_t = I_{d_x}$ and $\sigma_0^2=1.$}\label{subfig:L96_forecast_dx}
    \end{subfigure}
    ~
    \begin{subfigure}{0.32\textwidth}
        \centering
        \includegraphics[width=\textwidth]{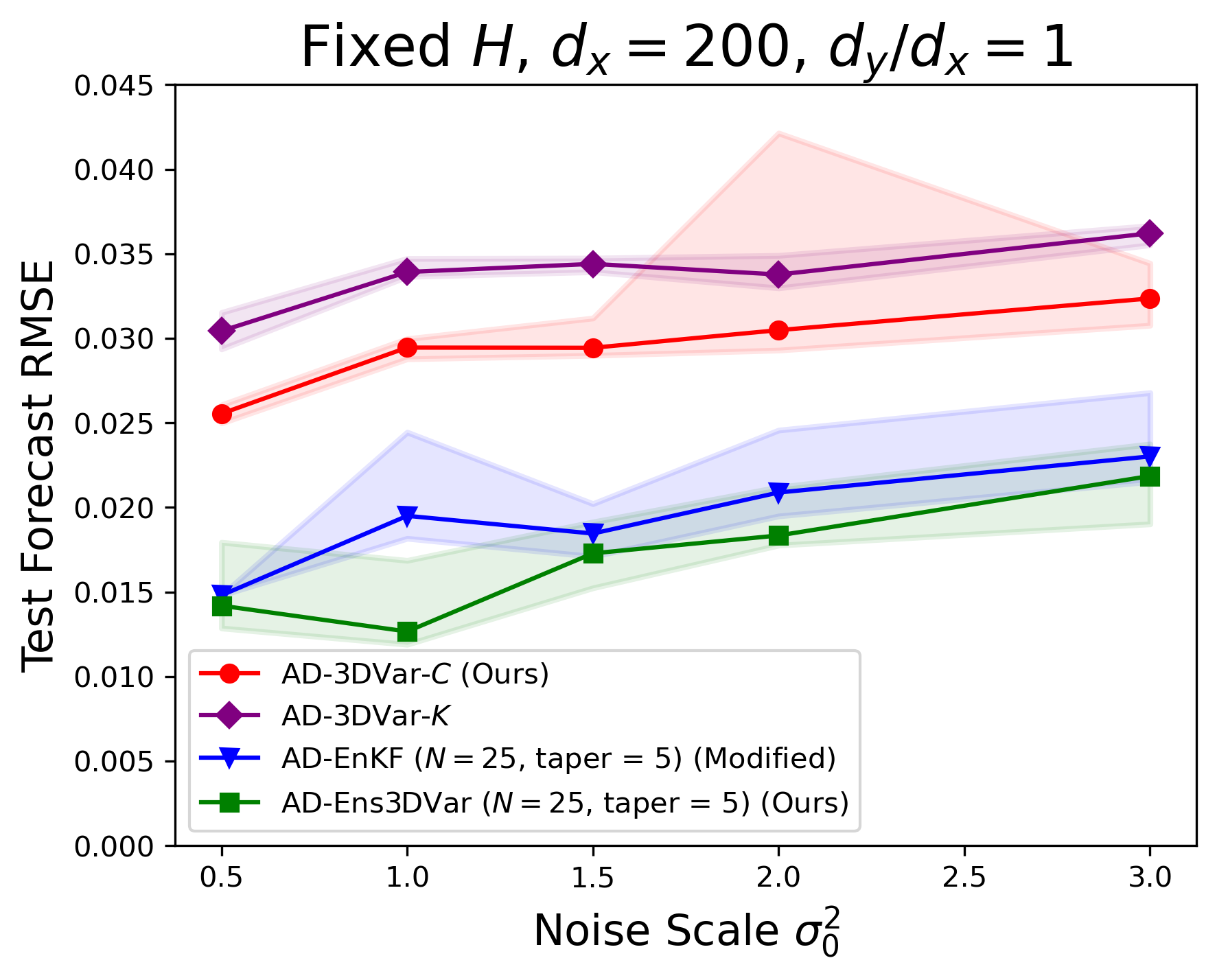}
        \caption{Test forecast performance of the model $\mathcal{F}_{\theta_1}$ across noise levels $\sigma_0^2$, where $d_x  =200$ and $H_t = I_{d_x}$.} \label{subfig:L96_forecast_noise}
    \end{subfigure}
    ~
        \begin{subfigure}{0.32\textwidth}
        \centering
        \includegraphics[width=\textwidth]{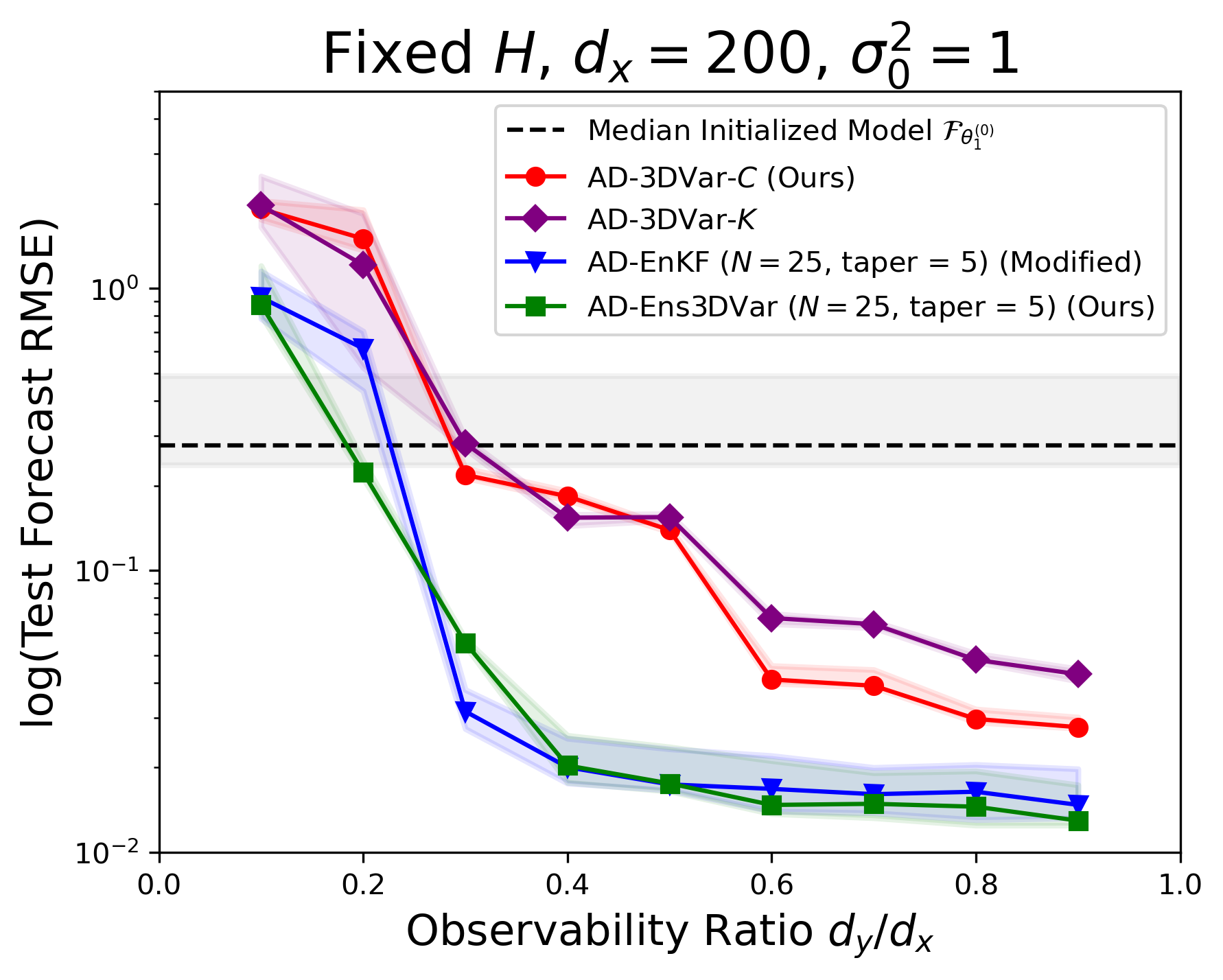}
        \caption{Test forecast performance of the model $\mathcal{F}_{\theta_1}$ across observability ratios $d_y/d_x$, where $d_x = 200$ and $\sigma^2_0 = 1$.}\label{subfig:forecast_dydx}
    \end{subfigure}
    
    \caption{Lorenz-96 learned forecast performance (Section \ref{sec:L96}). Test performance of the learned Lorenz-96 model in \eqref{eq:l96} learned from various methods: AD-3DVar-$C$,  AD-3DVar-$K$, AD-EnKF with a covariance tapering radius of 5 and $N=25$, and AD-Ens3DVar with a covariance tapering radius of 5 and $N=25$. The shaded regions correspond to the 0.2 and 0.8 quantiles of test forecast RMSEs across 10 independent simulations. In the plot varying $d_y/d_x$, we additionally plot the median forecast RMSE for the initialized models $\mathcal{F}_{\beta_j}$ for $j=1,\dots,J=10$ in the dashed black line (c), and the gray region corresponds to the 0.2 and 0.8 quantiles of test forecast RMSEs across these 10 initialized models.}
    \label{fig:L96_forecast}
\end{figure}

\paragraph{Performance in learning forecast parameters $\hat\theta_1$.} Figure \ref{fig:L96_forecast} shows the forecast performance of the learned $\mathcal{F}_{\hat \theta_1}$ across state dimensionalities $d_x$, variance of the noise $\sigma^2_0$ used to create initialized models $\mathcal{F}_0$, and observability ratios $d_y/d_x$. The ensemble-based learning methods AD-EnKF and AD-Ens3DVar consistently outperform AD-3DVar-$C$ and AD-3DVar-$K$ across all experiments, as expected. AD-EnKF and AD-Ens3DVar remain robust to the state dimension $d_x$ and noise scale $\sigma_0^2$ across various choices as seen in Figures \ref{subfig:L96_forecast_dx} and \ref{subfig:L96_forecast_noise}, whereas both the forecast performances of the models learned from AD-3DVar-$C$ and AD-3DVar-$K$ appear to increase as $d_x$ and $\sigma_0^2$ increase. Additionally, the ensemble-based learning algorithms are able to effectively learn from sparser observational datasets than the 3DVar-based algorithms, with the ensemble-based approaches degrading when $d_y/d_x < 0.3$ while the 3DVar-based approaches start noticeably degrading in forecast performance for $d_y/d_x <0.6$. For extremely low observability settings ($d_y/d_x$= 0.1 or $d_y/d_x$= 0.2), all methods fail to update $\mathcal{F}_{0}$ in a meaningful way, as evidenced by the forecast errors after training exceeding the forecast error of the initialized models, suggesting that the initialized models provide a better forecast model without any training. Overall, these results show that the ensemble-based AD-Ens3DVar and AD-EnKF methods are more robust than AD-3DVar-$C$ or AD-3DVar-$K$ to noise and can more effectively learn in low-observability regimes. However, AD-3DVar-$C$ and AD-3DVar-$K$ can still provide computationally cheap improvements to the forecast model in many problem settings.

\begin{figure}
\centering
\includegraphics[width=\textwidth]{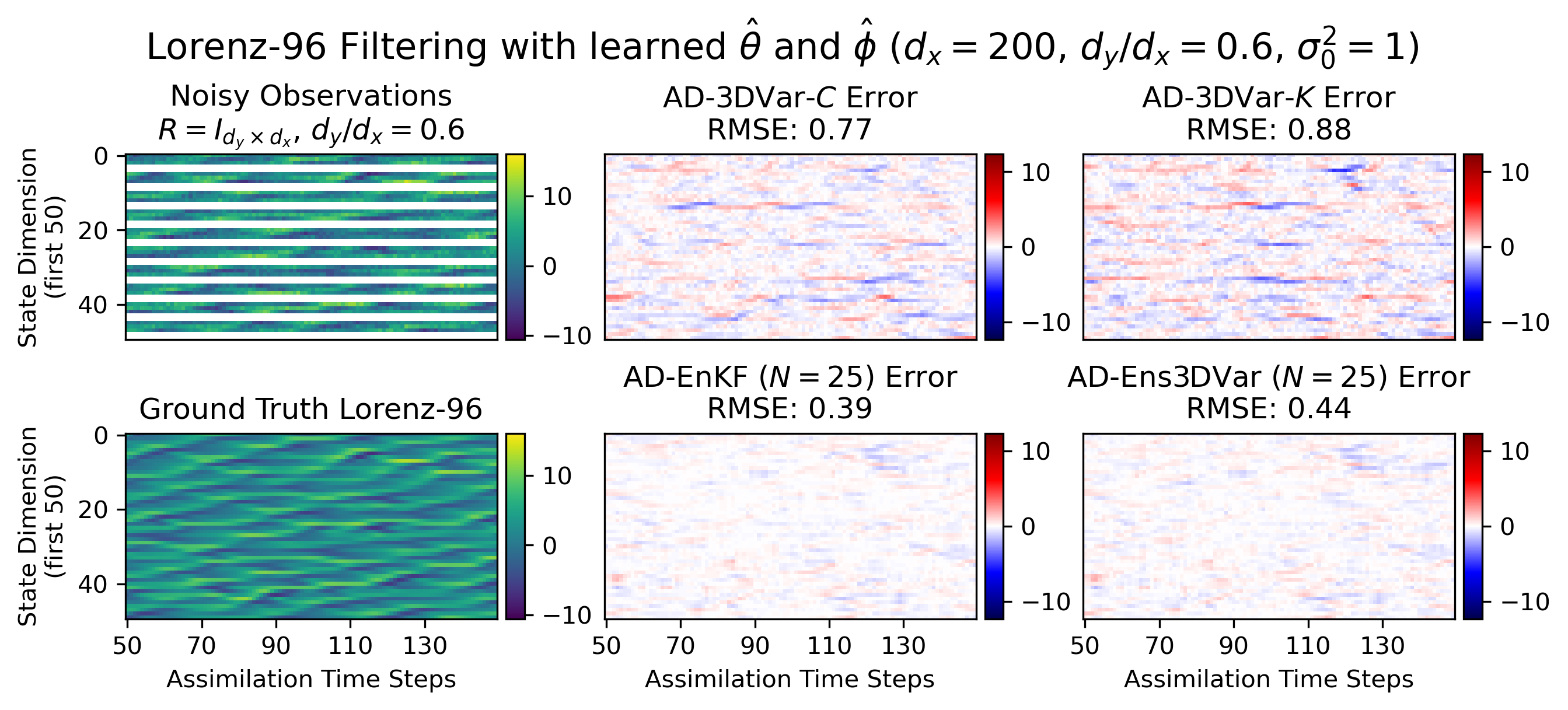}
\caption{Lorenz-96 filtering (Section \ref{sec:L96}). Visualization of filtering on test observation data using the learned $\hat\theta$ and $\hat \phi$ values across the methods AD-3DVar-$C$, AD-3DVar-$K$, AD-EnKF with $N=25$ ensemble members and a tapering radius of 5, and AD-Ens3DVar with $N=25$ ensemble members and a tapering radius of 5 compared to the noisy observations and ground truth Lorenz-96 simulations. In this plot, the settings of the filtering problem are $R=I_{d_y\times d_x}$, $d_y/d_x=0.6$, $\sigma^2_0=1$, and $d_x=200$. For ease of visualization, only the first 50 out of 200 state components are shown in the plot. Filtering is done with the base-filtering algorithm used during training (i.e., the results presented for AD-EnKF use the learned $\hat\theta$ and $\hat\phi$ in an EnKF algorithm).}
\label{fig:L96_filtering}
\end{figure}

\paragraph{Filtering performance using the learned $\hat \theta$ and $\hat \phi$.} Figure \ref{fig:L96_filtering} visualizes filtering with the learned $\hat\theta$ and $\hat \phi$ values across methods on unseen test observations with $R=I_{d_y\times d_x}$, $d_y/d_x = 0.6$, static $H$, $\sigma^2_0=1$, and $d_x=200$ for one simulated initial imperfect model $\mathcal{F}_0$. The ensemble-based methods provide the best filtering performance, with the error approximately halving compared to the 3DVar-based learning algorithms AD-3DVar-$C$ and AD-3DVar-$K$. Notably, the state estimates from AD-Ens3DVar and AD-EnKF appear to have errors of consistent magnitude across observed and unobserved components of the state. On the other hand, the analyses using the learned parameters from AD-3DVar-$C$ and AD-3DVar-$K$ show larger errors corresponding to unobserved components, which suggests that the learned $C_\phi$ and $K_\phi$ struggle to learn the covariances corresponding to unobserved state components, leading to increased filtering errors. This phenomenon was also observed in Figure \ref{fig:CW_filtering} for estimating the velocity components, and is a consequence of $H$ being held static across the entire assimilation. Additionally, an evaluation of the learned covariance inflation factor in AD-EnKF and the learned covariance mixing parameter in AD-Ens3DVar can be found in Figure \ref{fig:learned_alpha_gamma} in  Appendix \ref{app:training_details}.

\subsection{Population dynamics: Generalized Lotka-Volterra}\label{sec:GLV}

The generalized Lotka-Volterra equations \citep{Lotka1925} are 
\begin{equation}
    \frac{dx_i}{dt} = x_i f_i(x), \label{eq:glv}
\end{equation}
\noindent where $f(x) = r + Ax$, $x\in \mathbb{R}^{d_x}$ is a vector of $d_x$ species abundances, $r\in \mathbb{R}^{d_x}$ is the species birth/death rate vector, $A\in \mathbb{R}^{d_x\times d_x}$ is the species interaction matrix, and $i=1,\dots,d_x$ indexes the $d_x$ total species. These equations provide a simplified model for high-dimensional species interactions in an ecosystem. 

\subsubsection{Learning task}

In this setting, we are interested in parameter estimation. Therefore, we set $\theta_1 := \{A,r\}$, and our goal is to learn the elements of the species interaction matrix $A$ and the birth/death rate vector $r$.

\begin{figure}
\centering
\includegraphics[width=0.29\linewidth]{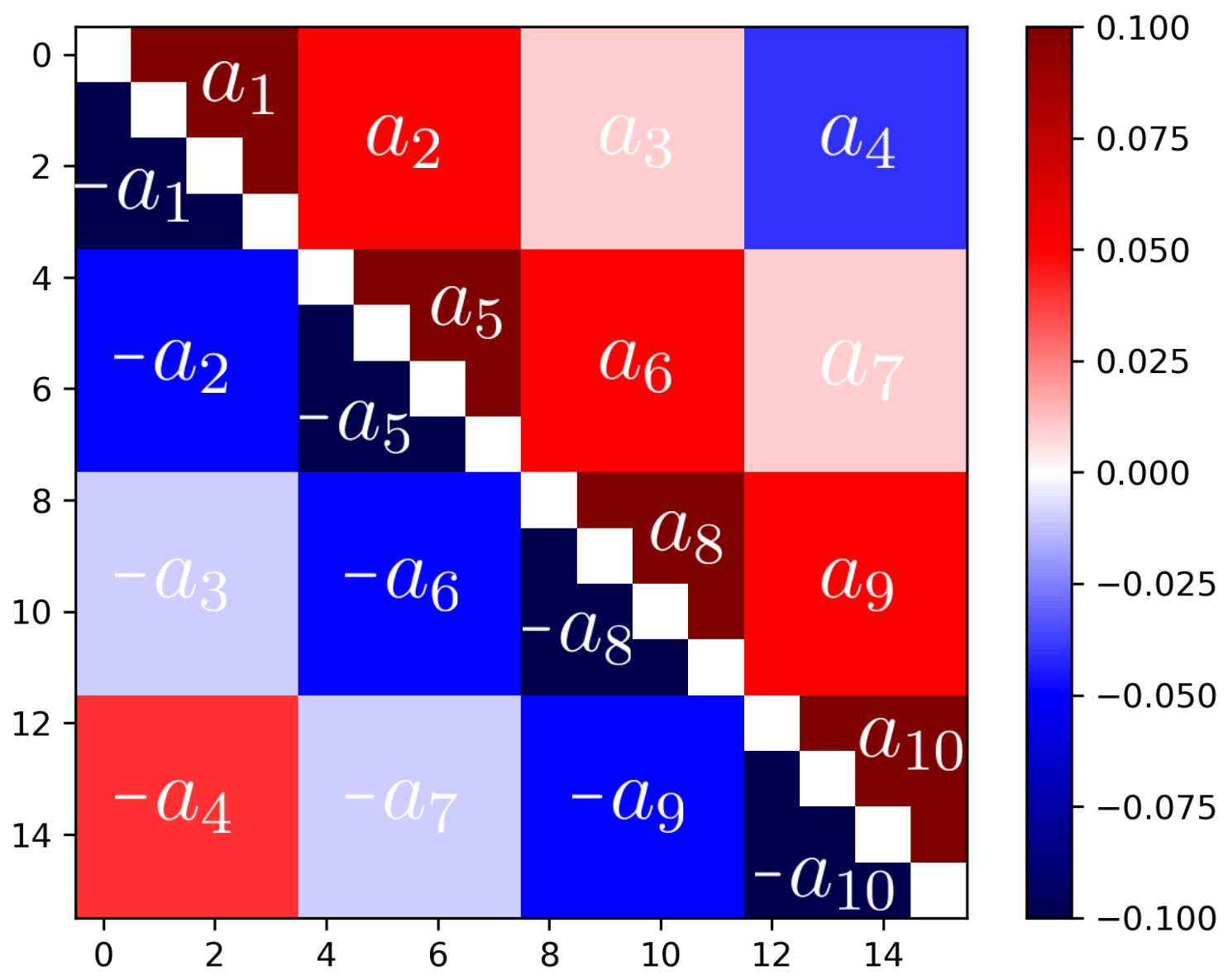}
\caption{Example interaction matrix $A$ for 16 interacting species in the generalized Lotka-Volterra equations. The values $\pm a_{1:10}$ denote the shared parameter values within each block.}
\label{fig:A_GLV}
\end{figure}

Hierarchical or block structures in species interactions can arise in real biological systems, and \citet{Poley2023} studies the stability of these hierarchical and block structures in generalized Lotka-Volterra equations. With this setting in mind, we learn $A$ as a specialized block structure, where an example is given in Figure \ref{fig:A_GLV}. Learning $A$ consists of learning parameters $\{a_1,\dots, a_{10}\}$ that govern the block-wise interactions among the species. We specify that the number of learnable parameters is invariant to the number of species in the system, meaning the number of species contained within each block increases as $d_x$ increases. This example is motivated by a setting where we have oracle knowledge of the group structure of species, but do not know the parameterization of the interactions among groups of species. To estimate the birth/death rate vector $r$, we learn a steady-state vector $x_s\in \mathbb{R}^{d_x}$ and use the property that $r = -Ax_s$ \citep{May1973}. Therefore, the total number of parameters estimated for $\mathcal{F}_{\theta_1}$ is $d_x + 10$.

To create imperfect initializations $\theta_1^{(0)}=\{A^{(0)}, r^{(0)}\}$, we utilize the true parameters $\theta_1^*=\{A^*,r^*\}$, where $A^*$ is parameterized by true $a^*=\{a_1^*,\dots,a_{10}^*\}$ and $r^*$ is constructed from the true $x_s^*$ and the true $A^*$ via $r^* = - A^*x_s^*$, and a noise level $\sigma_0^2$ to perturb these true values. For independent simulations $j=1,\dots,J=10$, we create imperfect interaction matrices via $a^{(0)}_{j}\sim \mathcal{N}(a^*,\sigma^2_0I_{10\times 10})$, and similarly construct $x_j^{(0)}\sim \mathcal{N}(x^*,\sigma^2_0I_{d_x\times d_x})$ and solve $r_j^{(0)} = -A_j^{(0)}x_j^{(0)}$.

\subsubsection{Results}

\begin{figure}
    \centering
    \includegraphics[width=0.98\linewidth]{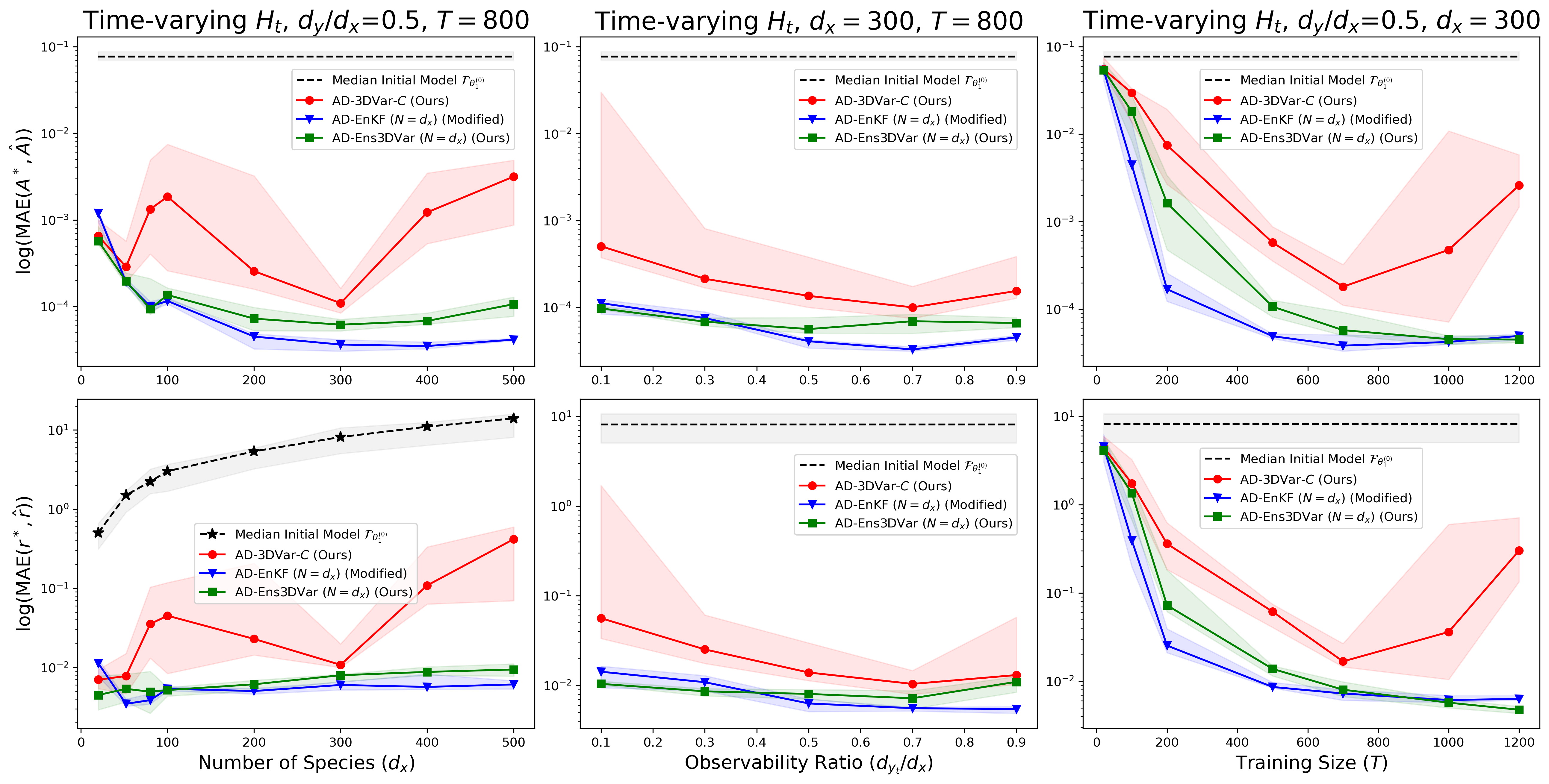}
    \caption{Generalized Lotka-Volterra parameter estimation (Section \ref{sec:GLV}). Parameter estimation mean absolute errors (MAEs) of the learned generalized Lotka-Volterra model in \eqref{eq:glv} compared to our specified ground truth set of model parameters. The parameters are learned using the methods AD-3DVar-$C$, AD-EnKF with $N=d_x$, and AD-Ens3DVar with $N=d_x$. The shaded regions correspond to the 0.2 and 0.8 quantiles of
    parameter MAEs
    across 10 independent simulations. In each plot, we plot the median MAEs for the initialized models $\mathcal{F}_{\theta_1^j}$ for $j=1,\dots,J=10$ in the dashed black line, and the gray region corresponds to the 0.2 and 0.8 quantiles of parameter MAEs across these 10 initialized models. The top row of plots corresponds to the error in the estimated interaction matrix $\hat A$ component of $\hat\theta_1$, and the second row of plots corresponds to the error in the estimated population change vector $\hat r$ component of $\hat\theta_1$.}
    \label{fig:GLV_forecast}
\end{figure}

\begin{figure}
\centering
\includegraphics[width=0.98\textwidth]{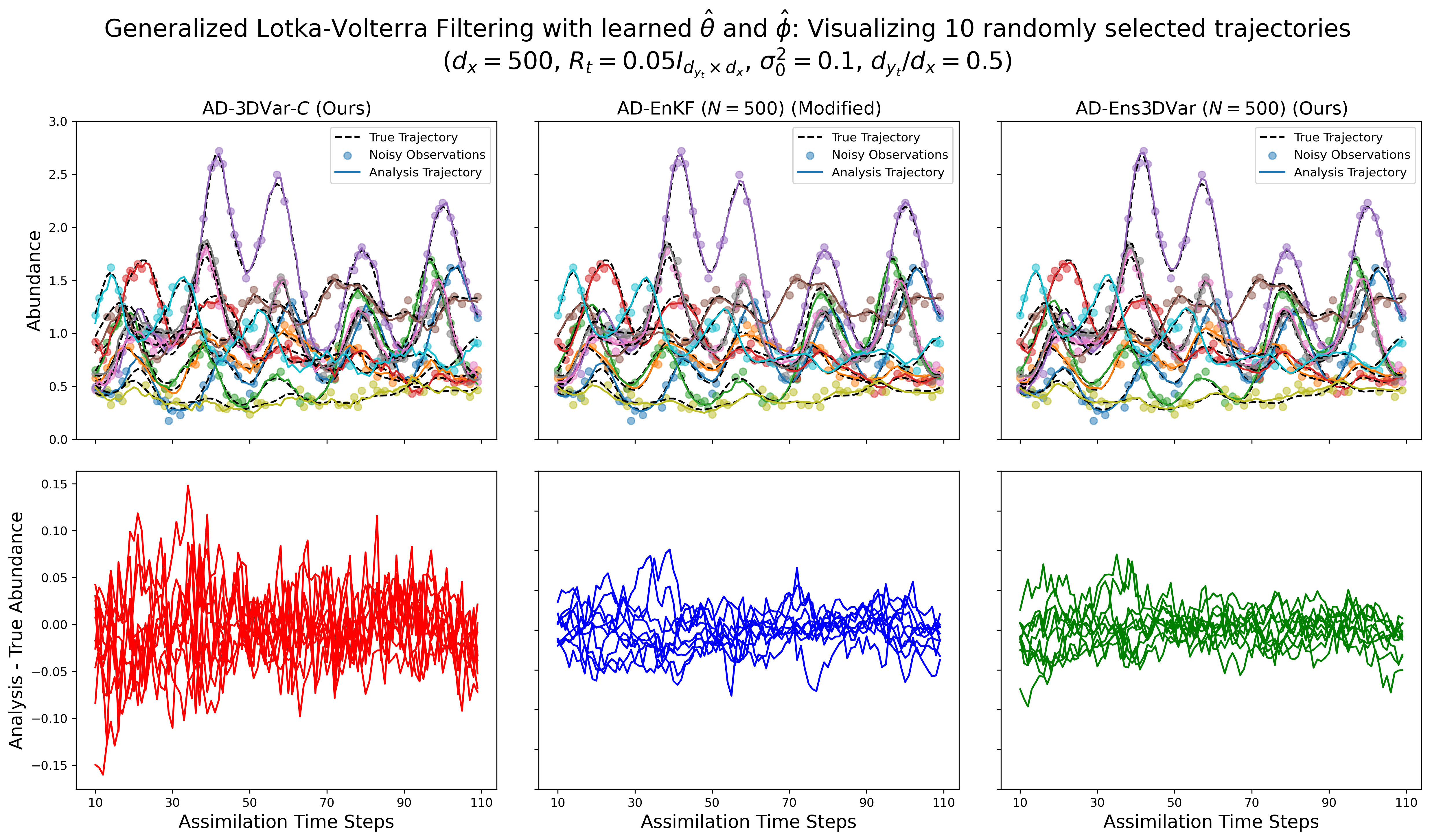}
\caption{Generalized Lotka-Volterra filtering (Section \ref{sec:GLV}). Visualization of filtering on test observational data using the learned $\hat\theta$ and $\hat\phi$ values across the methods AD-3DVar-$C$, AD-EnKF with $N=500$ ensemble members, and AD-Ens3DVar with $N=500$ ensemble members where $d_x = 500,$ $R_t = 0.05I_{d_{y_t}\times d_x}$, $\sigma_0^2 =0.1,$ and $d_{y_t}/d_x=0.5$. For ease of visualization, only 10 randomly selected species abundance trajectories out of the 500 total species are visualized. In each plot in the first row, the true trajectories are given by the dashed black lines, the sparse, noisy observation are plotted as a scatterplot, and the analyses are plotted as solid colored lines. The second row plots the errors in the analyses compared to the ground truth trajectories.}
\label{fig:GLV_filtering}
\end{figure}

We focus on a setting common to many problems in systems biology: one where the system is only partially observed at any given time, and the components (species) observed (and the number of observed components) may change across time. Therefore, $H_t$ varies with time and $d_{y_t}/d_x<1$. In our experiments, for each time point $t$, the observed species are randomly selected such that the ratio $d_{y_t}/d_x$ is fixed across time. We investigate the performance of  AD-3DVar-$C$, AD-EnKF, and AD-Ens3DVar, which are able to handle time-varying $H_t$. We omit AD-3DVar-$K$, since this method is \textit{not} able to handle observational data of this form. Training and data generation details are outlined in Appendix \ref{app:training_details}. Optimization hyperparameters were tuned for each simulation and parameter value across the $J=10$ independent simulated initial models. The time step used to solve the differential equation is $\Delta t = 0.05$, which corresponds to the same $\Delta t$ in which observations arrive.

\paragraph{Performance in learning forecast parameters $\hat\theta_1$.} Figure \ref{fig:GLV_forecast} visualizes the performance across the methods AD-3DVar-$C$, AD-EnKF, and AD-Ens3DVar in recovering the ground truth parameters $\theta_1^*:=\{A^*,r^*\}$ for a variety of settings, where the number of species $d_x$ in the system, the proportion of species that are observed $d_{y_t}/d_x$, and the training dataset size $T$ are all varied. Across all results in Figure \ref{fig:GLV_forecast}, AD-EnKF and AD-Ens3DVar perform similarly, with AD-EnKF performing better with some parameter choices. The performance of both ensemble-based methods remain relatively robust to the choices of parameters considered. AD-3DVar-$C$ exhibits a bit more volatile behavior across parameter values, particular for varied species $d_x$ and training size $T$, which could be in part due to suboptimal hyperparameter selections from observational validation data.

\paragraph{Filtering performance using the learned $\hat \theta$ and $\hat \phi$.} Figure \ref{fig:GLV_filtering} visualizes the analyses when filtering with the learned $\hat \theta$ and $\hat \phi$ parameters from the methods AD-3DVar-$C$, AD-EnKF, and AD-Ens3DVar for the parameter settings $d_x=500$, $R_t = 0.05I_{d_{y_t}\times d_x}$, $\sigma_0^2=0.1$, and $d_{y_t}/d_x=0.5$. This figure shows that all three methods are able to recover reasonable estimates of the species abundance trajectories given sparse, noisy observations where $H_t$ varies over time. As is consistent with our results in Figure \ref{fig:GLV_forecast}, the ensemble-based methods AD-EnKF and AD-Ens3DVar provide better estimates of the ground truth abundance trajectories, and AD-3DVar exhibits larger errors in the analyses. All methods, however, produce reasonable reconstructions of the species trajectories. An evaluation of the learned covariance inflation factor in AD-EnKF and the learned covariance mixing parameter in AD-Ens3DVar for the generalized Lotka-Volterra system can be found in Figure \ref{fig:learned_alpha_gamma} in Appendix \ref{app:training_details}.

\subsection{Fluid dynamics: two-layer QG model}\label{sec:QG}

The two-layer quasi-geostrophic model (QG) from geophysical fluid dynamics proposed in \citet{Phillips1954} describes large-scale geophysical fluid flow, such as fluid flow in oceanic and atmospheric processes. The equations for upper layer $i=1$ and lower layer $i=2$ are 
\begin{equation}
\label{eq:phillips-pv}
\partial_t q_i
+ U_i\,\partial_x q_i
+ J(\psi_i,q_i)
+ Q_{iy}\,\partial_x\psi_i
=
-\delta_{i2}\,\tau_f\nabla^2\psi_2
-\nu(\partial_x^8+\partial_y^8)q_i.
\end{equation}

where the potential vorticities $q_i$ and stream functions $\psi_i$ are related as 
  \begin{equation}
    \label{eq:phillips-inversion}
    q_1 = \nabla^2 \psi_1 - k_d^2 (\psi_1 - \psi_2),
    \qquad
    q_2 = \nabla^2 \psi_2 + k_d^2 (\psi_1 - \psi_2),
  \end{equation}
and where for generic inputs $a$ and $b$, $J(a,b) = \partial_x a\, \partial_y b - \partial_y a\, \partial_x b$ is the Jacobian, $U_1 = +U/2$ and $U_2 = -U/2$ are the mean zonal flows, and
  \begin{equation}
    \label{eq:phillips-Qy}
    Q_{1y} = \beta + k_d^2 U,
    \qquad
    Q_{2y} = \beta - k_d^2 U
  \end{equation}
are the mean potential vorticity gradients. The chosen values for the parameters are $\beta^*=0.196$, $\nu^*=1.024\times10^{-3}$, $\tau_f^*=0.2$, $k_d^*=1.0$, and $U^*=1.0$, which produce turbulent eddies throughout the domain. The continuous system is discretized on a uniform $96 \times 48$ grid, which represents the system \eqref{eq:phillips-pv} with state dimension $d_x = 2 N_yN_x = 9,216$.

\subsubsection{Learning task}

\paragraph{Parameter estimation.} Our first set of experiments focuses on parameter estimation of a subset of parameters in the QG model, namely $\beta$, $U$, and $\tau_f$. We set $\theta_1 := \{\beta, U, \tau_f\}$, and our goal is to estimate these parameters. The remaining parameters are set to their ground truth values. 
Details of data generation are provided in Appendix \ref{app:subsubsec:qg_datagen}.

To construct a set of independent initially imperfect models, we set $\theta_{1,j}^{(0)} = \theta_{1}^*(1+s\varepsilon_j),$ where $\varepsilon_j \sim \mathcal{N}(0.5, 0.05)$ and $s \in \{-1,1\}$. This procedure constructs a set of 10 initial perturbed parameter sets that differ from their ground truth values by roughly $50\%$.

\paragraph{Emulation.} We next perform experiments to learn an emulator of the two-layer QG dynamics under the same parameter settings. We construct a residual U-Net such that the network predicts persistence (i.e., the forecast is simply the input to the system) at initialization, and we independently generate 10 initial parameter sets that all predict persistence at initialization. The number of learnable parameters in this specified model is $|\theta_1| = 1,949,730$. Details on the network architecture and data generation are provided in Appendix \ref{app:subsubsec:qg_datagen}.

Since both of these learning tasks involve a much higher dimensional state than prior examples in this paper, modifications were made to the implementation in the learning of $C_\phi$ in AD-3DVar-$C$. These details are outlined in Appendix \ref{app:3dvarC_qg}.

\subsubsection{Results}

For results on the two-layer QG system, we focus on a setting where locations on the 2D discretized grid are randomly observed across time, so $H_t$ varies with time for $d_{y_t}/d_x < 1$. For each time point $t$, the locations are randomly selected such that the ratio $d_{y_t}/d_x$ is fixed across time. Since $H_t$ varies with time in these experiments, we assess learning only with the methods AD-3DVar-$C$, AD-EnKF (Modified), and AD-Ens3DVar, which fall within our framework. AD-3DVar-$K$ cannot learn in such a setting. Training and data generation details are outlined in Appendix \ref{app:training_details}. Optimization hyperparameters were tuned for one simulation and parameter value across the $J=10$ independent simulated initial models. The time step used to solve the differential equation is $\Delta t = 0.005$ with observations arriving at the cadence $\Delta t_\text{obs} =0.5$ for the parameter estimation experiments, and $\Delta t_\text{obs} = 0.1$ for the emulator experiments. Across all experiments, observation noise is set to $R = 0.05 I_{d_y}$.

\begin{figure}
    \centering
    \includegraphics[width=\linewidth]{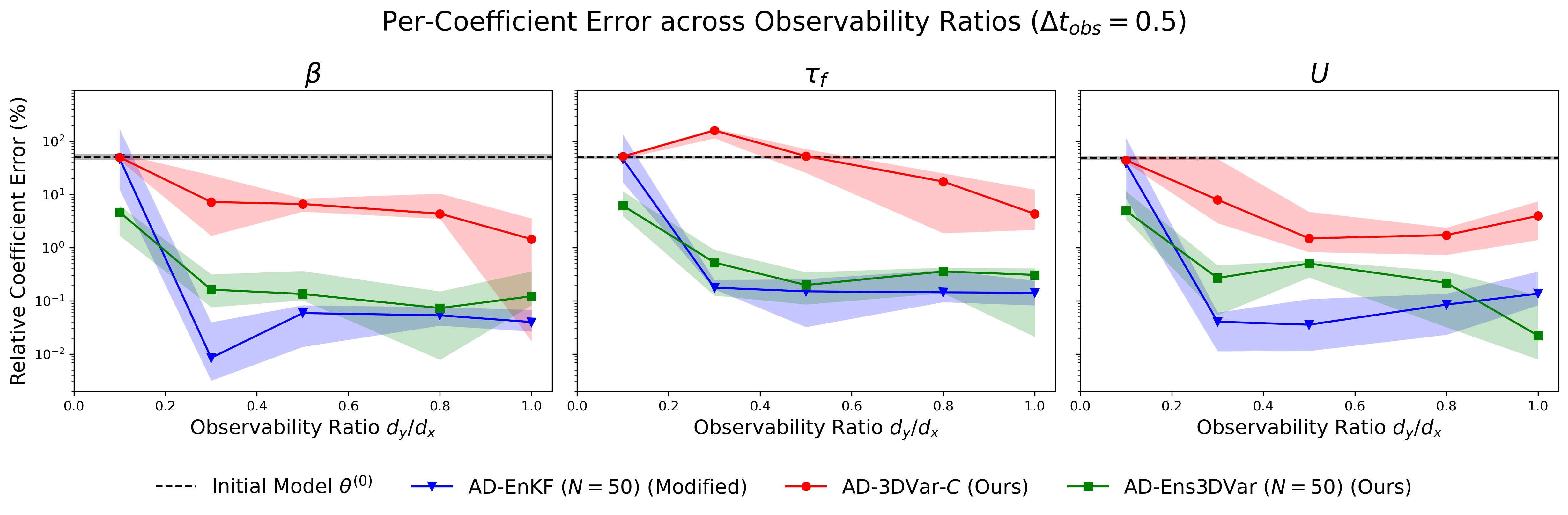}
    \caption{Two-layer QG parameter estimation (Section \ref{sec:QG}). Parameter estimation errors computed as relative coefficient errors of the learned two-layer QG model in  
    \eqref{eq:phillips-pv}
    learned from the methods AD-3DVar-$C$, AD-EnKF with $N=50$, and AD-Ens3DVar with $N=50$. The shaded regions correspond to the 0.1 and 0.9 quantiles of relative parameter estimation error across 10 independent simulations. In each subfigure, we plot the median relative coefficient error (\%) for the initialized models $\mathcal{F}_{\theta_1^j}$ for $j=1,\dots,J=10$ in the dashed black line, and the gray region corresponds to the 0.1 and 0.9 quantiles of relative coefficient errors across these 10 initialized models.}
    \label{fig:QG_forecast}
\end{figure}

\begin{figure}[htbp]
\centering
\includegraphics[width=0.8\textwidth]{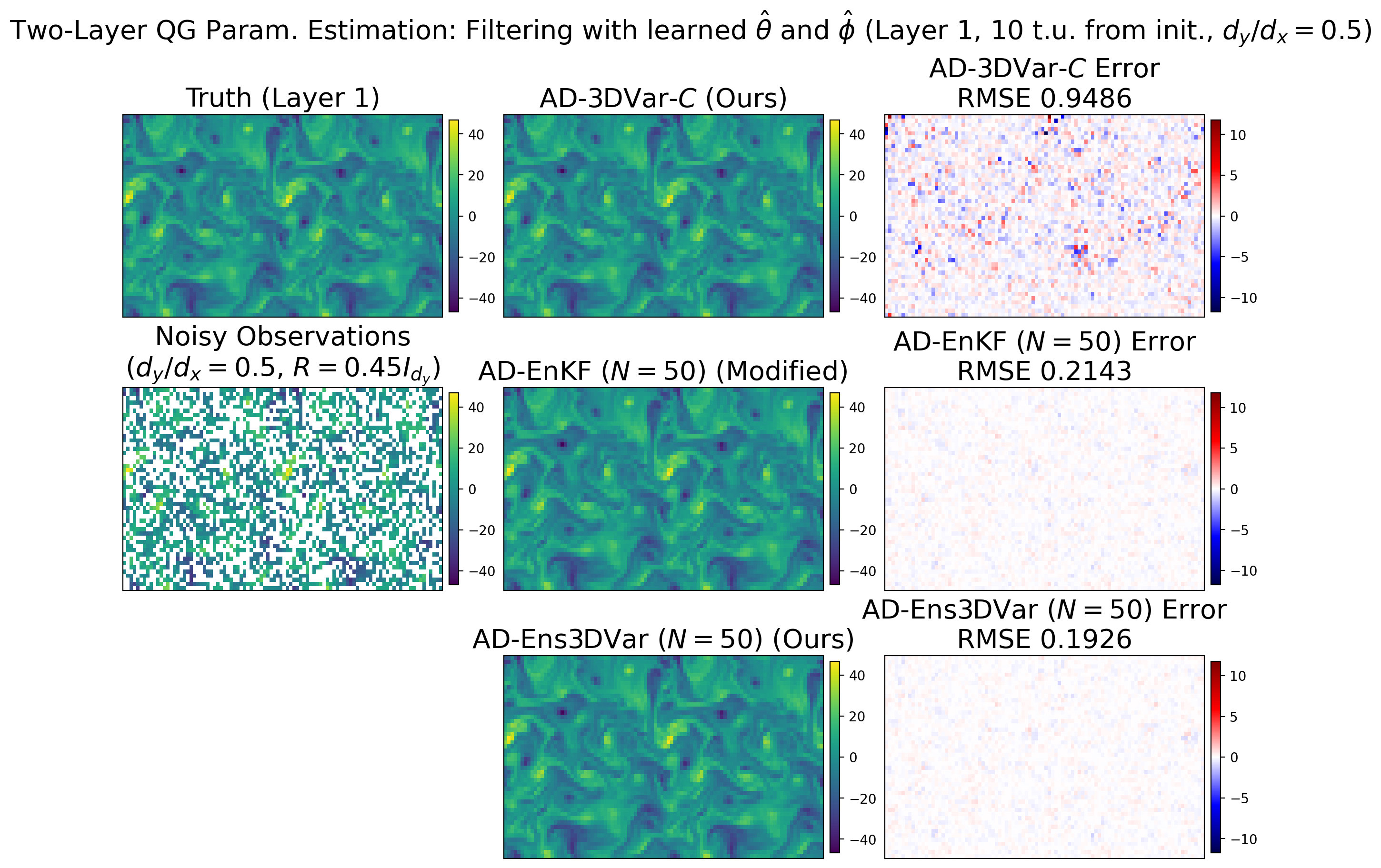}
\caption{Two-layer QG parameter estimation (Section \ref{sec:QG}). Visualization of filtering on test observational data for layer 1 using the learned $\hat\theta$ and $\hat\phi$ values across the methods AD-3DVar-$C$, AD-EnKF with $N=50$ ensemble members, and AD-Ens3DVar with $N=50$ ensemble members where $d_{y_t}/d_x=0.5$. The image shows a snapshot of the analyses across methods at the end of the test filter window, corresponding to 10 time units (t.u.) from initialization.}
\label{fig:QG_filtering_layer1}
\end{figure}

\begin{figure}
\centering
\includegraphics[width=0.35\textwidth]{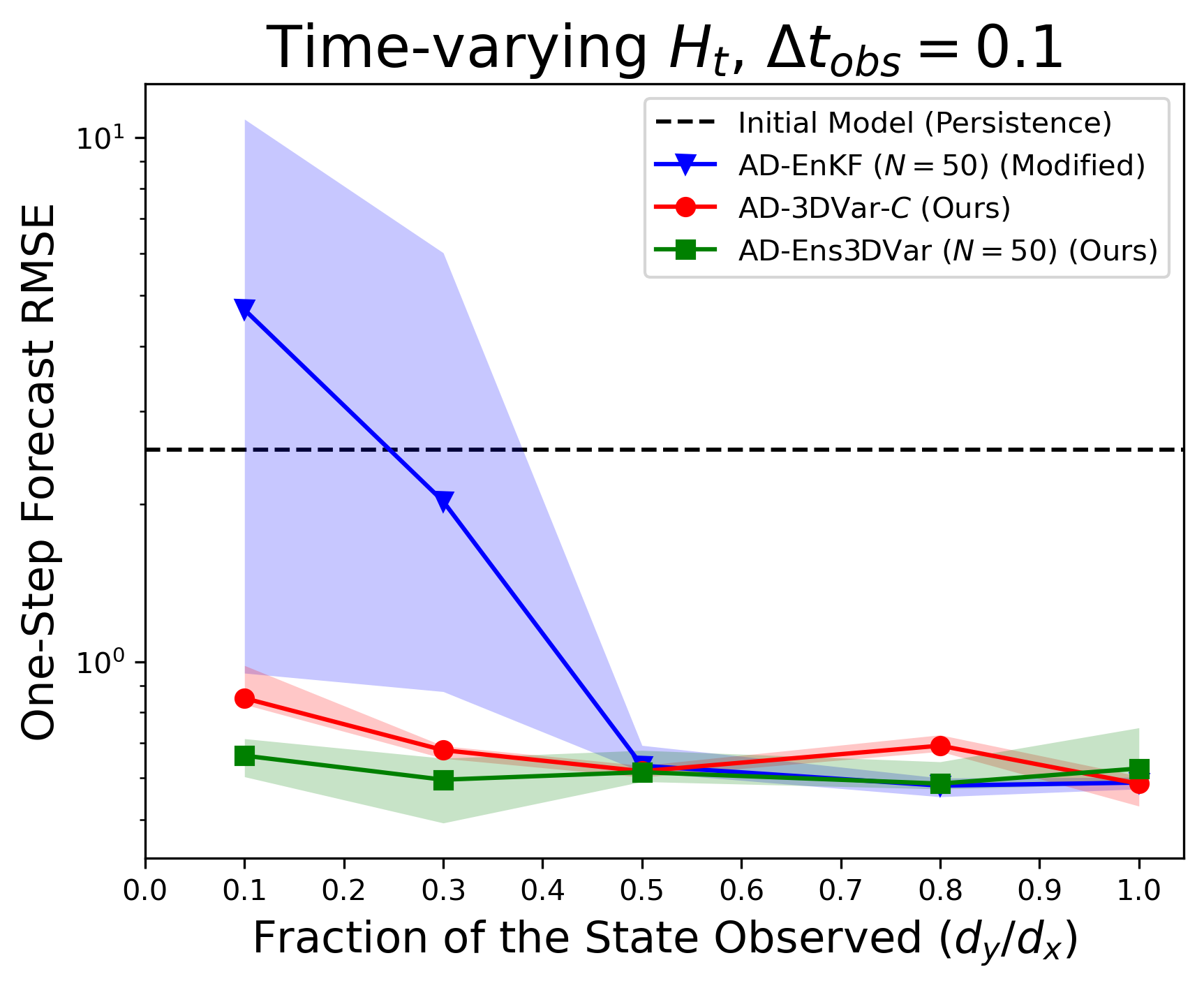}
\caption{Two-layer QG emulation (Section \ref{sec:QG}). Visualization of the RMSE for a one-step forecast using the learned emulator parameters $\hat \theta_1$ across the methods AD-EnKF ($N=50$), AD-3DVar, and AD-Ens3DVar ($N=50$). The shaded regions correspond to the 0.1 and 0.9 quantiles of forecast RMSE across 10 independent simulations. We plot the error corresponding to persistence, the initialization of all 10 learnable models, with the dashed black line.}
\label{fig:QG_forecast_emulator}
\end{figure}

\begin{figure}
\centering
\includegraphics[width=0.8\textwidth]{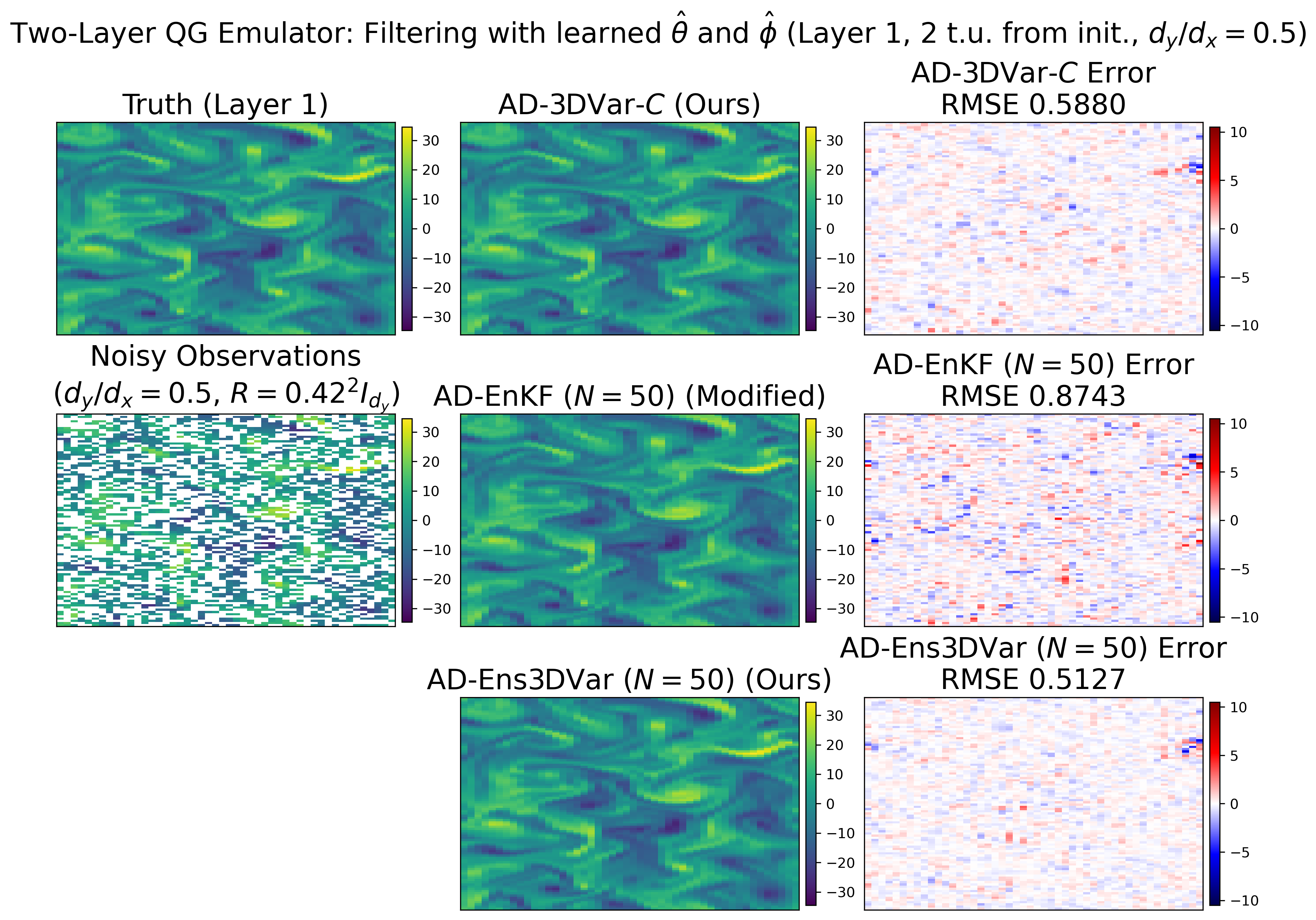}
\caption{Two-layer QG emulation (Section \ref{sec:QG}). Visualization of filtering on test observational data for layer 1 using the learned $\hat\theta$ and $\hat\phi$ values across the methods AD-3DVar-$C$, AD-EnKF with $N=50$ ensemble members, and AD-Ens3DVar with $N=50$ ensemble members where $d_{y_t}/d_x=0.5$. The image shows a snapshot of the analyses across methods at the end of the test filter window, corresponding to 2 time units (t.u.) from initialization.}
\label{fig:QG_filtering_emulator_layer1}
\end{figure}

\begin{figure}
\centering
\includegraphics[width=\textwidth]{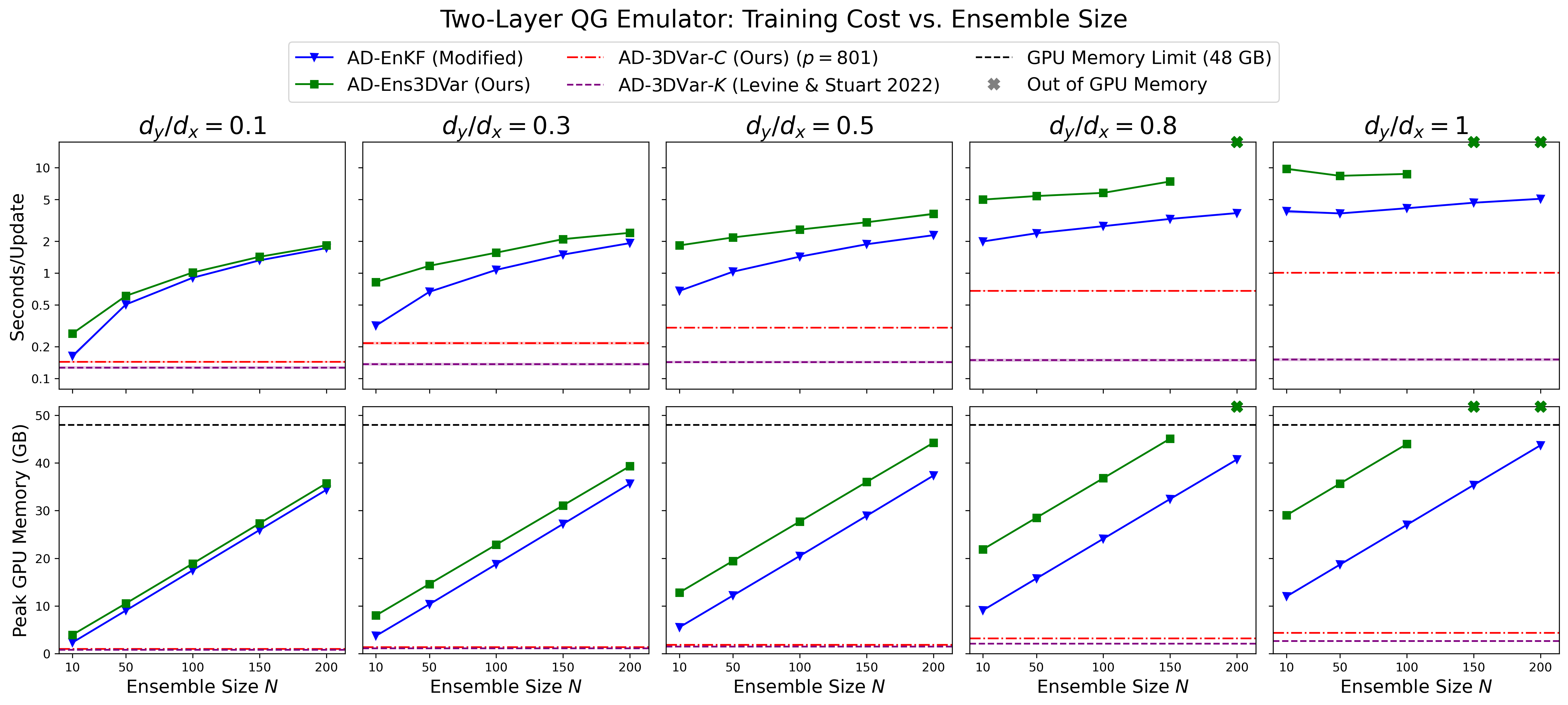}
\caption{Two-layer QG emulation (Section \ref{sec:QG}). Training cost in terms of seconds per update and peak GPU memory (GB) versus ensemble size $N$ across observability ratios $d_y/d_x$ for AD-EnKF and AD-Ens3DVar, where AD-3DVar-$C$ $(p=801)$ and AD-3DVar-$K$ are included for comparison. The black dashed line corresponds to the GPU memory limit based on a 48GB NVIDIA RTX A6000 (Ampere) GPU. For experiments that ran out of memory, an "X" is shown at that ensemble member index.}
\label{fig:QG_ens_time_mem}
\end{figure}

\begin{figure}
\centering
\includegraphics[width=\textwidth]{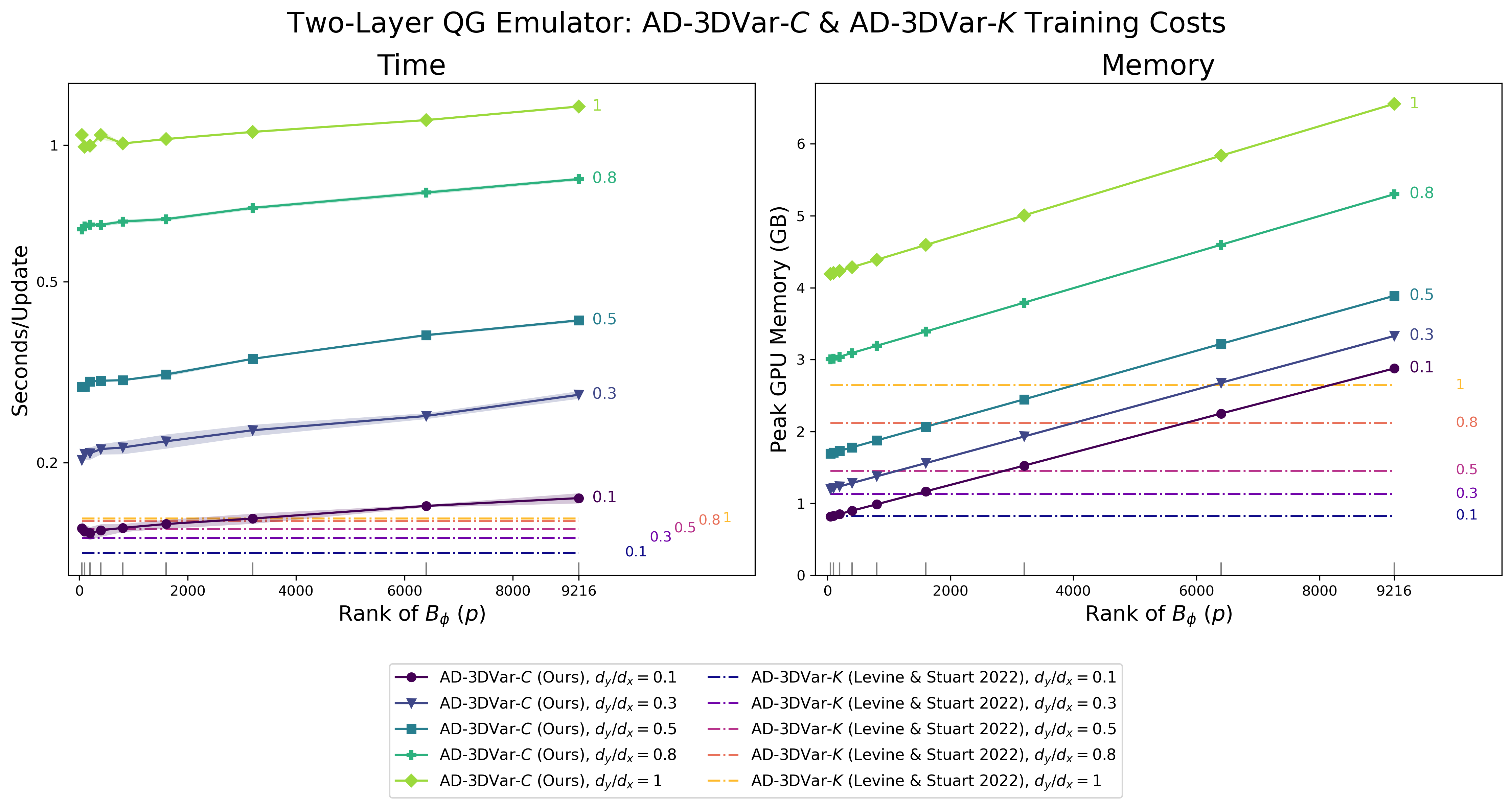}
\caption{Time and memory consumption for learning an emulator of the two-layer QG system for AD-3DVar-$C$ across $p$ and AD-3DVar-$K$. Seconds per update for a subsequence of length $L=10$ and peak GPU memory (GB) are compared across varying ranks ($p$) of the learned $B_\phi$ in the AD-3DVar-$C$ method across observability ratios $d_y/d_x$. For comparison, AD-3DVar-$K$ performance metrics are included, which are independent of $p$. Shaded regions correspond to the $\pm 1$ standard deviation across 10 independent subsequence updates. All experiments were performed on a 48GB NVIDIA RTX A6000 (Ampere) GPU. }
\label{fig:QG_rank_time_mem}
\end{figure}

\paragraph{Parameter estimation: Performance in learning forecast parameters $\hat\theta_1$.} Figure \ref{fig:QG_forecast} visualizes the performance across the methods AD-3DVar-$C$, AD-EnKF, and AD-Ens3DVar in recovering the ground truth parameters $\theta_1^*:=\{\beta^*,U^*, \tau_f^*\}$ for a variety of settings, where the proportion of 2D locations that are observed $d_{y_t}/d_x$ varies. Across the results in Figure 
\ref{fig:QG_forecast}
, AD-EnKF and AD-Ens3DVar perform similarly, with AD-EnKF performing better with some parameter choices. AD-3DVar-$C$ largely exhibits consistent learning, with one instance of degraded parameter performance in $\hat\tau_f$ at $d_y/d_x = 0.3$, likely due to a suboptimal choice of hyperparameters based on the validation log-likelihood. 

\paragraph{Parameter estimation: Filtering performance using the learned $\hat \theta$ and $\hat \phi$.} The filtering performance for one of the 10 independently generated models trained is visualized for layer 1 in Figure \ref{fig:QG_filtering_layer1} and Figure \ref{fig:QG_filtering_layer2} for layer 2 in the setting where $d_y/d_x=0.5$. Consistent with previously presented results, the two ensemble-based learning approaches AD-EnKF and AD-Ens3DVar provide superior filtering results compared to the single particle AD-3DVar estimates. All three methods produce highly accurate analyses using their respective estimates of $\hat \theta$ and $\hat \phi$.

\paragraph{Emulation: Performance in learning forecast parameters $\hat\theta_1$.} Figure \ref{fig:QG_forecast_emulator} visualizes the one-step-ahead forecast performance for the learned models AD-EnKF ($N=50$), AD-3DVar-$C$, and AD-Ens3DVar $(N=50)$ across observability ratios $d_y/d_x$. All methods perform comparably across high observability regimes, showing significant improvements to the learned model compared to initialization. At low observability regimes, however, AD-EnKF appears to degrade in performance compared to the other approaches. This result is likely due to the regularization that the specified $C_\phi$ provides at initialization for both AD-3DVar-$C$ and AD-Ens3DVar; this matrix was initialized with dense covariance structure, whereas the initial ensemble for AD-EnKF was generated with a diagonal covariance, making spatial dependencies harder to recover in low-observability regimes.

\paragraph{Emulation: Filtering performance using the learned $\hat \theta$ and $\hat \phi$.} The filtering performance on test observations is visualized in Figure \ref{fig:QG_filtering_emulator_layer1} using the learned forecast parameters $\hat \theta$ and learned filtering parameters $\hat \phi$ across methods. Each method demonstrates visually similar estimates of the state at the end of the test assimilation window (2 t.u. from initialization), where AD-3DVar-$C$ and AD-Ens3DVar show similar filtering performance and AD-EnKF shows slightly better performance.

\subsubsection{Emulation: Cost comparisons.}
To tangibly measure the differences in memory and performance across all evaluated methods, we performed an experiment with static observation locations across time so that we are able to compare with AD-3DVar-$K$, which can only operate in this setting. We fixed the hardware to a single 48 GB NVIDIA RTX A6000 (Ampere) GPU and measured the memory consumption and the seconds per update for each method. For the ensemble based methods AD-EnKF and AD-Ens3DVar, we computed these metrics across choices of ensemble size $N$, and we computed these metrics across the rank of $B_\phi$ ($p$) for AD-3DVar-$C$. AD-3DVar-$K$ is independent of both of these choices, so the computation time and memory usage remain constant across $N$ and $p$. An update is defined here as the time to process a full subsequence of length $L$, involving the forward pass, auto-differentiating to compute the gradients, and updating the parameters. In this experiment, $L=10.$

Figure \ref{fig:QG_ens_time_mem} shows the time per update and peak GPU memory across ensemble sizes $N$ and observability ratios $d_y/d_x$, and Figure \ref{fig:QG_rank_time_mem} shows the time per update and peak GPU memory across the choices of rank $p$ for $B_\phi$ and observability ratios $d_y/d_x$. Between these two figures, the two single-member assimilation methods AD-3DVar-$C$ and AD-3DVar-$K$ are the cheapest and lightest weight, with AD-3DVar-$K$ being most time and memory efficient. At full observability and full rank for $B_\phi$, the memory footprint for AD-3DVar-$C$ does not exceed 7 GB, making this method scalable to much higher resolutions than presently evaluated. The ensemble-based methods AD-EnKF and AD-Ens3DVar, on the other hand, are more memory and time expensive, with AD-Ens3DVar being the most expensive for a fixed choice of ensemble size $N$ due to the additional learned $C_\phi$ matrix. 
These methods approach or exceed the GPU memory limit for ensemble sizes in the low hundreds,
with AD-Ens3DVar exceeding memory for $N\geq 150$ in the full observations case. These results suggest that for larger scale systems, attention needs to be given to tuning the system enough to perform well in low ensemble-size regimes, which can be achieved through covariance tapering and inflation techniques.

\section{Discussion}\label{sec:discussion}

This work proposes a general framework for jointly learning states, forecasting parameters, and filtering parameters via noisy and potentially sparse observations of dynamical systems. Our approach achieves this goal by leveraging filtering algorithms in an offline learning optimization algorithm. We outline and compare particular special cases of our framework that may be used directly by practitioners; however, we note that similar learning algorithms from our general framework can be derived. As previously emphasized, there is no one formulation of our auto-differentiable filtering framework that is best suited for all problems, which leads practitioners to make informed choices for their particular application. We now provide general guidelines for practitioners to consider when making these choices for their particular application, informed by our results. In addition, we discuss possible future directions that stem from this work.

\subsection{General guidelines for practitioners} 

The following discussion of the general guidelines for practitioners when selecting or constructing an auto-differentiable filtering algorithm is summarized in Table \ref{tab:summary}.

We first discuss when it is advantageous to learn $\theta$ and $\phi$ with the ensemble-based learning approaches, AD-Ens3DVar or AD-EnKF. Across all experiments, the ensemble-based filtering approaches provided the best performance in forecasting with the learned forecast model and in filtering with the learned forecast and filter components. However, this performance comes at the price of higher computational cost and memory footprint. Therefore, if accuracy is of utmost importance to a practitioner, AD-EnKF or AD-Ens3DVar would be the preferred choices. Another setting where AD-EnKF or AD-Ens3DVar would be preferred is when the observation matrix $H_t \equiv H$ is fixed. As noted in Sections \ref{sec:CW} and \ref{sec:L96}, when some state components are completely unobserved for the entire time series trajectory, the lighter-weight methods AD-3DVar-$C$ and AD-3DVar-$K$ struggle to accurately estimate these components on test data, whereas this phenomenon did not occur with the ensemble-based learning methods. One significant consideration with ensemble-based methods, however, is their memory footprint, which may be prohibitive for off-the-shelf use in high-dimensional settings and may require simplifying modifications to be used.

For settings where computational speed is important, particularly in high-dimensional settings, AD-3DVar-$C$ or AD-3DVar-$K$ could be preferable since these approaches are computationally cheaper and memory-light. AD-3DVar-$C$ could be particularly useful in settings where $H_t$ is time-varying, since this method can learn reasonable approximations of all state components, as seen in Figure \ref{fig:GLV_filtering}. AD-3DVar-$C$ can compute off-the-shelf on higher-dimensional problems compared to ensemble-based methods without simplifying modifications. Since AD-3DVar-$K$ is even lighter-weight than AD-3DVar-$C$, if a practitioner is especially in need of a fast computational learning method, AD-3DVar-$K$ could be useful in settings where $H$ is fixed for the entire time series. However, this method may come with a slight degradation in the ability to improve a forecast model and filter with the learned Kalman gain.

\begin{table}[ht]
    \centering
    \begin{tabular}{cccccc}\toprule

         & & Time-&High-& &  \\
         Algorithm & $H_t \equiv H$& varied $H_t$ &dimensional $d_x$ & Memory &  Learning $\mathcal{F}_{\theta_1}$ \\ \midrule
         AD-EnKF (Modified)&  Reasonably  & \ding{51}& May require  & High; scales &  Accurate \\ 
         \citep{Chen2022}&recovers&&non-trivial &with $N$&\\
         & all states & & modifications& &  \\\midrule
         AD-Ens3DVar& Reasonably  & \ding{51}&May require & High; scales  & Accurate  \\  
         (Ours)&recovers&&modifications &with $N$&\\
         & all states& & or small $N$ && \\ \midrule 
         AD-3DVar-$C$ &  Struggles & \ding{51} & Computation& Lower & Moderate  \\ 
         (Ours)&to recover &&scales better&&accuracy\\
         & unobserved states& & for large $d_x$ & & \\ \midrule 
        AD-3DVar-$K$&Struggles& \ding{55}&Computation& Lowest & Moderate\\
        \citep{Levine2022}&to recover&&scales best&&accuracy\\
        & unobserved states & & for large $d_x$& & \\
        \bottomrule 

    \end{tabular}
    \caption{Summary table of particular examples of auto-filtering algorithms in terms of their ability to learn accurate filtering parameters when $H_t \equiv H$ when $d_{y} <d_x$, ability to filter with a time-varying $H_t$, relative scalability to high-dimensional systems, relative memory footprint, and relative accuracy in improving the forecast model $\mathcal{F}_{\theta_1}$.}
    \label{tab:summary}
\end{table}

\subsection{Future directions} 

For high-dimensional systems, AD-Ens3DVar and AD-EnKF may be computationally or memory prohibitive off-the-shelf; therefore, one may consider making simplifying modifications to break the problem down into smaller, parallelizable tasks. For example, a modified version of EnKF that allows for parallelization, particularly the local ensemble transform Kalman filter (LETKF) \citep{Ott2004, Hunt2007} algorithm may provide comparable accuracy while allowing for computation on large-scale problems using multiple GPUs for high-dimensional systems where large ensembles are needed. These modifications can be made for practitioners desiring a learning method that is both scalable to high-dimensional problems and accurate. Evaluating an AD-LETKF type of algorithm is left as a future direction.

As previously noted, the formulation of our optimization task is offline, and all of our experiments operate in this setting. However, one can easily imagine slight modifications that allow for on-the-fly corrections to the forecast and filtering parameters through an online learning approach. This modification would be particularly of interest in settings where near-real-time inference is required. Moreover, it is known that smoothing-based data assimilation methods provide more accurate state estimation, and therefore may provide more accurate learning of forecast and filtering parameters. Though more computationally costly, an interesting future direction would be to assess the computational feasibility and empirical gain from a similar learning framework that utilizes smoothing rather than filtering. The framework presented in this work lays the foundation for further development in these directions.

\section*{Acknowledgments}
This research was supported in part by grants from the NSF (DMS-2235451) and Simons Foundation (MPS-NITMB-00005320) to the NSF-Simons National Institute for Theory and Mathematics in Biology (NITMB). We also gratefully acknowledge the support of DOE 0J-60040-0023A and AFOSR FA9550-24-1-0327. The work of DSA was partly funded by the NSF CAREER award DMS-2237628. MA was partly supported by the NSF Graduate Research Fellowship Program under Grant No. DGE-1746045.

\section*{Declaration of generative AI and AI-assisted technologies in the\\ manuscript preparation process}

During the preparation of this work the authors used Claude in order to assist with code generation for experiments and manuscript editing. After using this tool/service, the authors reviewed and edited the content as needed and take full responsibility for the content of the published article.

\bibliographystyle{apalike} 
\bibliography{ref}

\begin{appendix}
\section{Specification of $C_\phi$ in AD-3DVar-$C$}\label{app:C_phi}

\subsection{Lorenz-96}

As discussed in the main text, we specify in AD-3DVar-$C$ that we learn $B_\phi$ such that $C_\phi = B_\phi B_\phi^\top$, where we learn elements of the dense matrix $B_\phi \in \mathbb{R}^{d_x\times d_x}$.
In practice, particularly for large-scale systems, it may be infeasible to learn a dense matrix $B_{\phi}\in\mathbb{R}^{d_x \times d_x}$ since this matrix may be too large to store, or even if it is small enough to store, learning this dense matrix may introduce additional computation that may be avoided by learning a more efficient choice of $B_{\phi}$. One alternative option is to learn a sparse upper triangular matrix $B_\phi^U$ such that $C_\phi = B_\phi^U (B_\phi^U)^\top$ with $|\phi|\leq \frac{1}{2}d_x(d_x+1)$, where sparsity constraints on the upper diagonal structure of $B_\phi^U$ reduce the number of learnable parameters and reduce memory by only storing nonzero elements. 

Another alternative that we more thoroughly explore is defining $C_\phi$ such that we learn a low-rank update to some initial $C_0$. 
Specifically, we consider parameterizations of the form
\begin{equation}
    C_{\phi} = C_0 + B_{\phi}B_{\phi}^\top,
\end{equation}
where $B_{\phi}\in \mathbb{R}^{d_x\times p}$ and $p\ll d_x$. This setting assumes that we have some reasonable initialization of $C_{\phi}$, potentially constructed based on domain knowledge and the performance of the initialized $\mathcal{F}_{\theta_1^{(0)}}$, which we want to efficiently improve with a low-rank update. 

\begin{figure}
    \centering
    \includegraphics[width=0.75\linewidth]{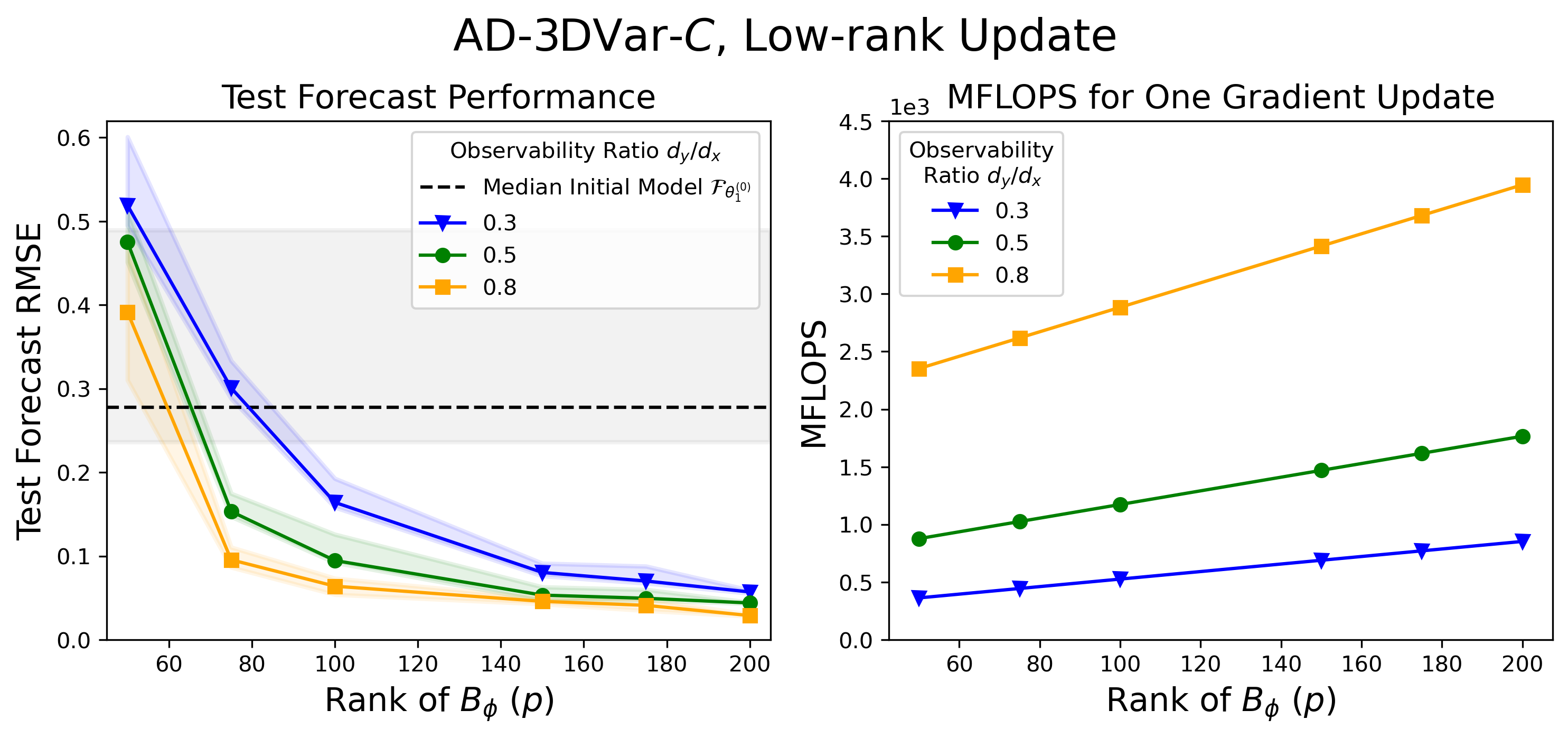}
    \caption{Experiment based on the Lorenz-96 system described in Section \ref{sec:L96}. \textbf{(Left)} Median test forecast RMSE for learning a low rank update to the background covariance such that $C_{\phi} = C_{0} + B_{\phi}B_{\phi}^\top$, where $B_{\phi}\in\mathbb{R}^{d_x \times p}$ and $p<d_x$. In this experiment, $d_x=200,$ $d_y/d_x\in[0.3,0.5,0.8]$, $\sigma_0^2=1$, and observations at each time $t$ are randomly sampled across state dimensions (i.e., $H_t$ varies with time). Each training task was replicated for $J=10$ independent simulations. We additionally plot the median forecast RMSE for the initialized models $\mathcal{F}_{\beta^{(j)}}$ for $j=1,\dots,M$ in the dashed black line, and the gray region corresponds to the 0.2 and 0.8 quantiles of test forecast RMSEs across these 10 initialized models. \textbf{(Right)} Plot of millions of floating point operations per second (MFLOPS) across values of $p$ for processing one subsequence (both the forward and backward pass) for the setting where $d_x=200$, $p\in[50,75,100,150,175,200], L=20,$ and $d_y/d_x\in[0.3,0.5,0.8]$.}
    \label{fig:lowrankB}
\end{figure}

 We numerically investigate the performance of AD-3DVar-$C$ with low-rank parameterization of $C_\phi$ using the Lorenz-96 system under different observability ratios. Details on the training task on Lorenz-96 data can be found in Section \ref{app:training_details}. We explore the impact of the choice of $p$ (the maximum rank of $B_\phi$) on our ability to learn an improved forecast model $\mathcal{F}_{\theta_1}$, and these results are visualized in Figure \ref{fig:lowrankB}. In this experiment, $p\in[50, 75, 100, 150, 175, 200]$, $d_x =200$, $d_y/d_x \in [0.3,0.5,0.8]$, $\sigma^2_0=1$, $H_t$ varies with time $t$ with randomly selected observed dimensions at each time point, and $C_0 = I_{d_x}$.  Figure \ref{fig:lowrankB} shows that learning a low rank update for high observability regimes (e.g., $d_y/d_x = 0.8$) remains robust across choices of $p$ up until approximately $p=d_x/2$. This robustness degrades as the observability decreases. 
This experiment shows evidence that learning a low-rank update can be feasible, but the performance of the learned model can vary depending on specific aspects of the problem, including how low-rank $B_\phi$ is specified to be.

\subsection{Two-layer QG}\label{app:3dvarC_qg}

In our higher-dimensional two-layer QG example, nontrivial modifications are needed to learn $C_\phi$ to minimize computational cost and storage burden during computation. We aim to avoid storing a dense representation of $C_\phi$ since this matrix has shape $d_x \times d_x$, which may be infeasible to store on a single GPU. We provide an alternative learning scheme for $C$ that accounts for this limitation in higher-dimensional systems. We can instead specify $C_\phi$ as 
\begin{equation}
    C_\phi = \rho \circ [(L_0 + B_\phi)(L_0 + B_\phi)^T],
\end{equation}
\noindent where $\rho$ is a covariance tapering matrix, $B_\phi \in \mathbb{R}^{d_x \times p}$ is a learnable matrix with parameters $\phi$, and $L_0\in\mathbb{R}^{d_x \times p}$ is a precomputed forecast matrix with $p$ columns. To create this matrix, a ground truth initial state $x_0^{*}$ is perturbed to create a noisy initial state $\tilde x_0 \sim \mathcal{N}(x_0^*, \sigma^2_{\text{init}}I_{d_x})$. This initial condition is rolled out via $\mathcal{F}_{\theta_1^{(0)}}$ for 200 time units as spin-up, and after this spin-up period, $p=801$ snapshots are retained every $0.5$ time units, corresponding to samples on the imperfectly initialized model's attractor. These samples are denoted as $\hat x^{(i)}_{\theta_1^{(0)}}$ for $i =1,\dots, p$. The matrix $L_0$ is then computed as 
\begin{equation}
    L_0 = \frac{\gamma}{\sqrt{p-1}}[\hat x^{(1)}_{\theta_1^{(0)}} - \hat m, \  \hat x^{(2)}_{\theta_1^{(0)}} - \hat m,  \ \dots \ , \hat x^{(p)}_{\theta_1^{(0)}} - \hat m],
\end{equation}
\noindent where $\hat m = \frac{1}{p}\sum_{i=1}^p \hat x^{(i)}_{\theta_1^{(0)}}$, and $\gamma$ is a scaling factor. We initialize $B_\phi : = \{0\}^{d_x\times p}$ so that the filtering covariance $C_\phi$ is initialized at a static forecast covariance aligned with the initial model's forecast errors. Since $(L_0 + B_\phi)(L_0 + B_\phi)^\top$ yields a rank-deficient matrix, tapering via $\rho$ is applied element-wise to the resulting matrix to increase the rank of $C_\phi$, which improves the well-posedness of the matrix inversion of the Kalman gain.

\section{Learning the covariance tapering matrix $\rho$ as part of $\phi$ in AD-EnKF (Modified)}\label{app:learning_rho}

We can modify \eqref{eq:C_enkf} to include learning of the covariance tapering matrix by rewriting the equation as 

\begin{equation}
    \hat C^{\theta,\phi}_t = (1+\phi_1)(\rho(\phi_2)\circ \hat C_t^\theta),
\end{equation}
\noindent where $\phi_1$ corresponds to the covariance inflation factor,  $\rho(\phi_2)$ is a tapering matrix with tapering radius $\phi_2$, and $\phi = \{\phi_1,\phi_2\}$. A common way to specify the tapering matrix, particularly in applications with spatial components, is via the Gaspari-Cohn tapering function \citep{Gaspari1999}. We adopt this standard parameterization of the tapering matrix in what follows.

\begin{figure}
\centering
\includegraphics[width=0.35\textwidth]{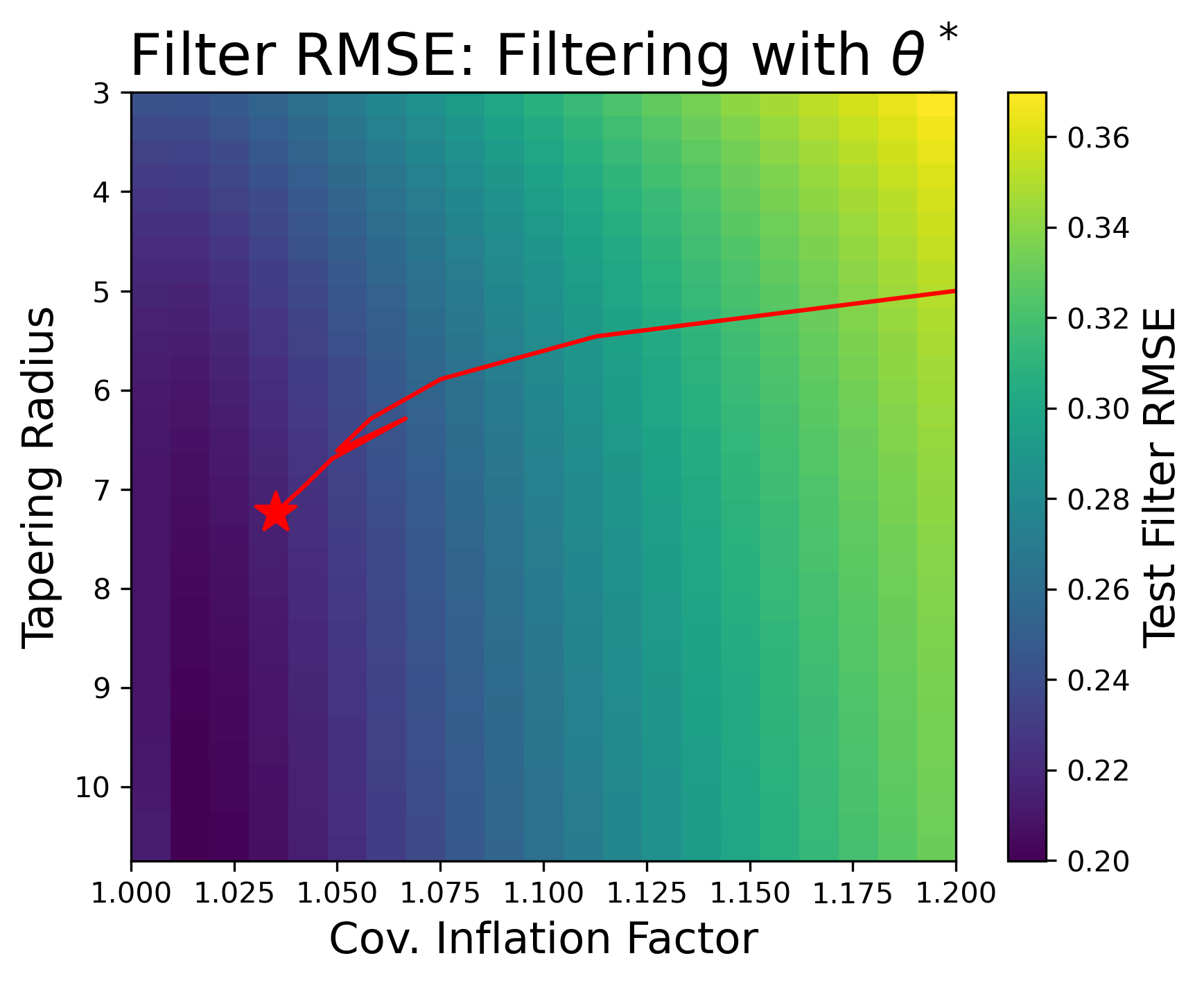}
\caption{Heatmap of the EnKF filter RMSE values when filtering with the true $\mathcal{F}_{\theta_1^*}$ on unseen test data across different choices of tapering radius and covariance inflation for $d_x=40$, $d_y/d_x=1$, and $\sigma_0^2 =1$. The red trajectory corresponds to the convergence of the analysis parameter vector $\phi\in \mathbb{R}^{2}$ for the covariance tapering and inflation for AD-EnKF (Modified), which was initialized at tapering = 5 and inflation = 1.2. The red star corresponds to location of the converged value for the (tapering, inflation) pair.}
\label{fig:L96_taper_infl}
\end{figure}

Figure \ref{fig:L96_taper_infl} shows the convergence of the $\phi\in\mathbb{R}^2$ parameters for AD-EnKF (Modified) that learn the covariance inflation and the tapering radius superimposed on the test filter RMSE of EnKF with the true $\theta^*$ parameters. We can see that the optimization trajectory correctly traverses to an area of low test filter RMSE, but does not necessarily attain the minimum test filter loss. This result is due to the fact that observation noise impacts the optimization landscape, which can lead to convergence to a suboptimal region. 

We also note that learning the covariance tapering radius \textit{can be extremely computationally expensive, even for small $d_x$} due to the fact that $\rho(\phi_2)$ needs to be recomputed every time $\phi_2$ is updated after backpropagation. When implementing TBPTT, parameters are updated $\lceil T/L\rceil$ times per epoch. Since this matrix can be large with many elements, frequently updating this matrix and backpropagating through can be computationally expensive. As an alternative to gradient-based learning, one could choose a best initial guess of $\phi^{(0)}_2$, fix it during AD-EnKF (Modified) training, then perform a grid search over possible choices of $\phi_2$ by evaluating EnKF with $\hat\phi$ and $\hat \theta$ based on the forecast log-likelihood on validation observations (i.e., Eq. \eqref{eq:loss} with held-out validation observations). This process is likely to be computationally lighter than a gradient-based learning approach.

\section{Varying the ensemble size $N$ for AD-EnKF (Modified) and AD-Ens3DVar}\label{app:enkf_vs_ens3dvar}

\begin{figure}
    \centering
    \begin{subfigure}{0.33\textwidth}
        \centering
        \includegraphics[width=\textwidth]{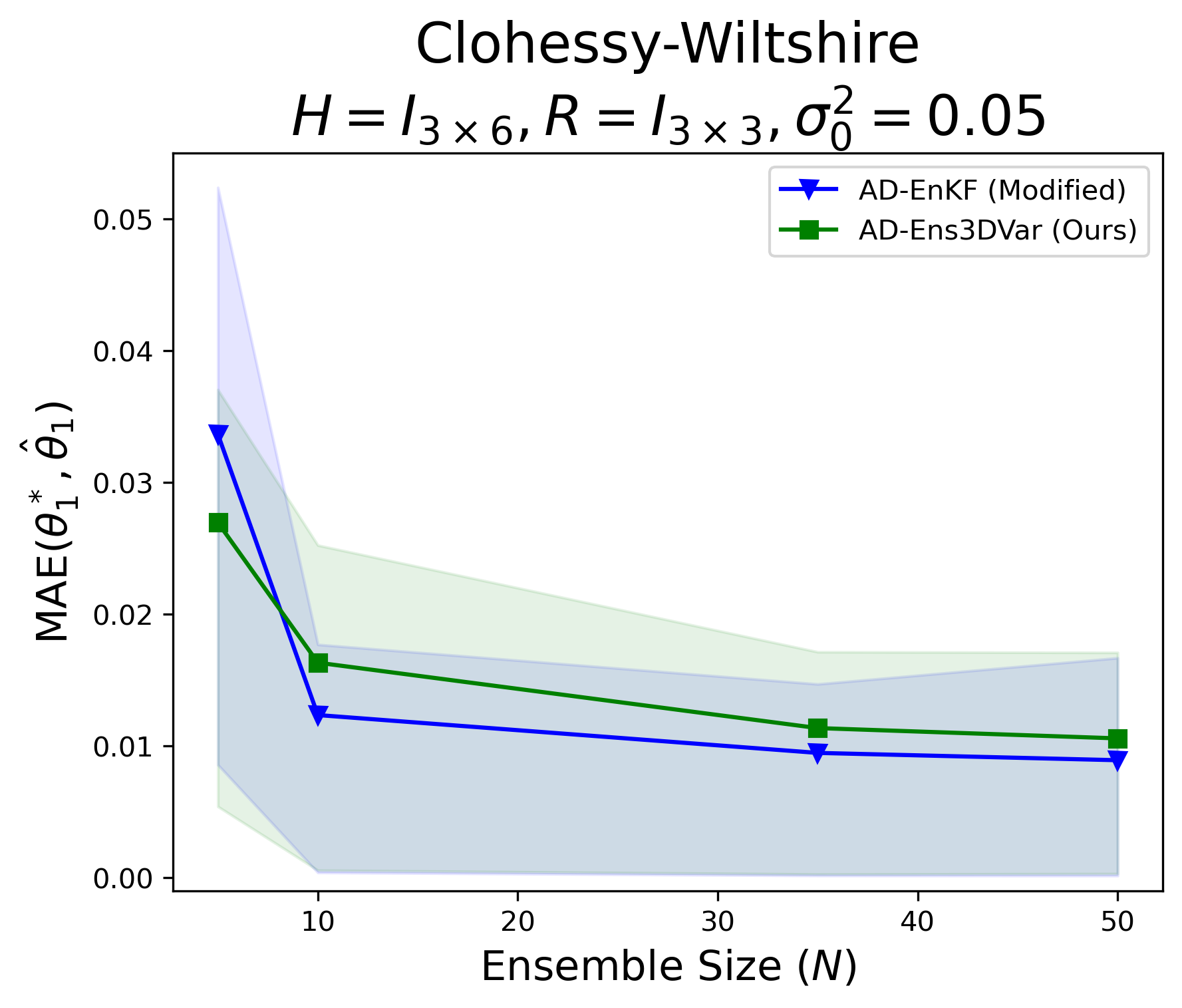}
        \caption{Clohessy-Wiltshire (Section \ref{sec:CW}): learning the forecast parameter $\theta_1$ where $H=I_{3\times 6}$, $R=I_{3\times 3}$, and $\sigma^2_0=0.05$.}\label{subfig:CW_varyN}
    \end{subfigure}
    ~
    \begin{subfigure}{0.33\textwidth}
        \centering
        \includegraphics[width=\textwidth]{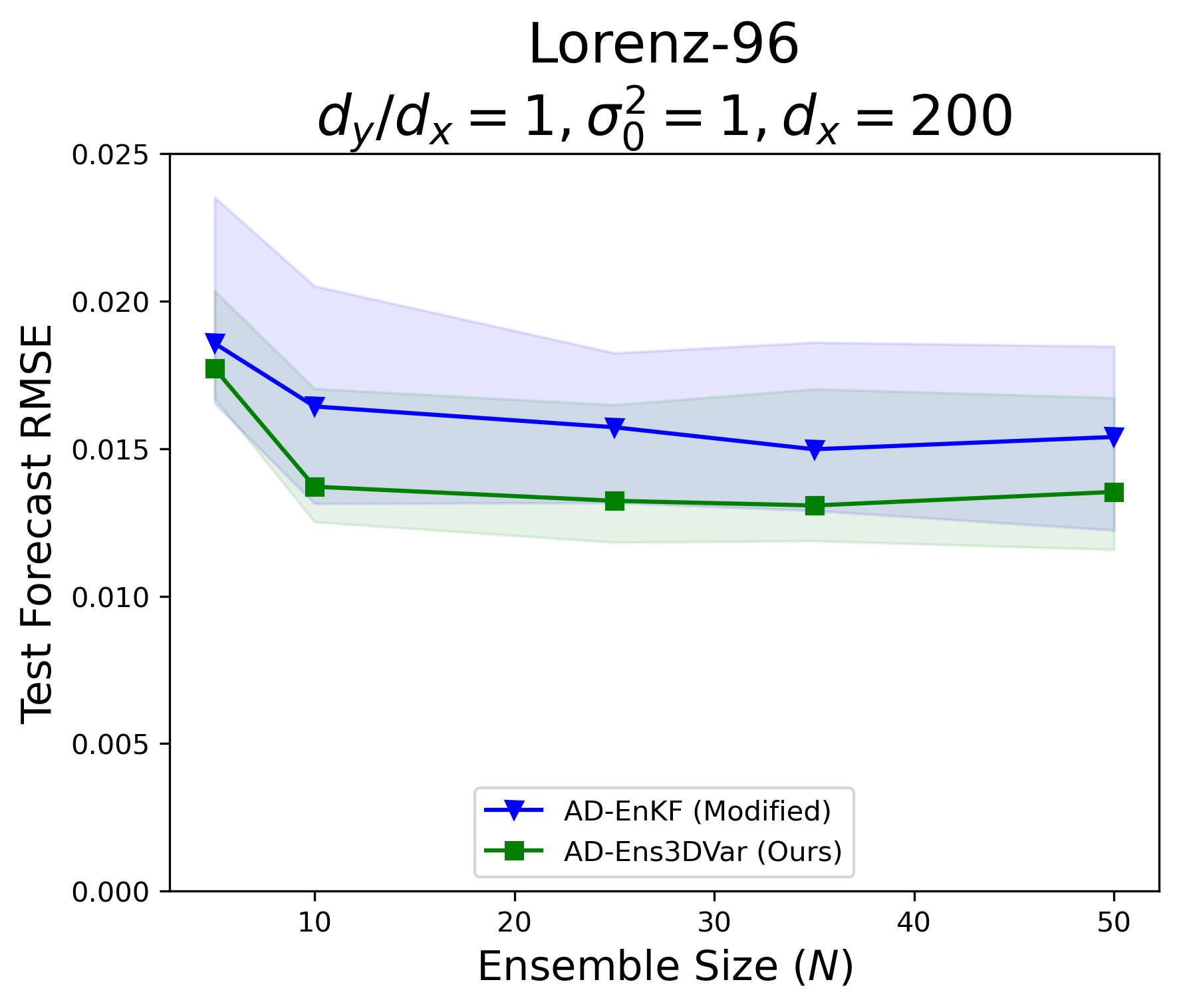}
        \caption{Lorenz-96 (Section \ref{sec:L96}): test forecast performance where $H=I_{d_x\times d_x}$, $R=I_{3\times 3}$, $\sigma^2_0=I_{d_x}$, and $d_x=200$. \\ } \label{subfig:L96_varyN}
    \end{subfigure}
    ~
    \begin{subfigure}{0.33\textwidth}
        \centering
        \includegraphics[width=\textwidth]{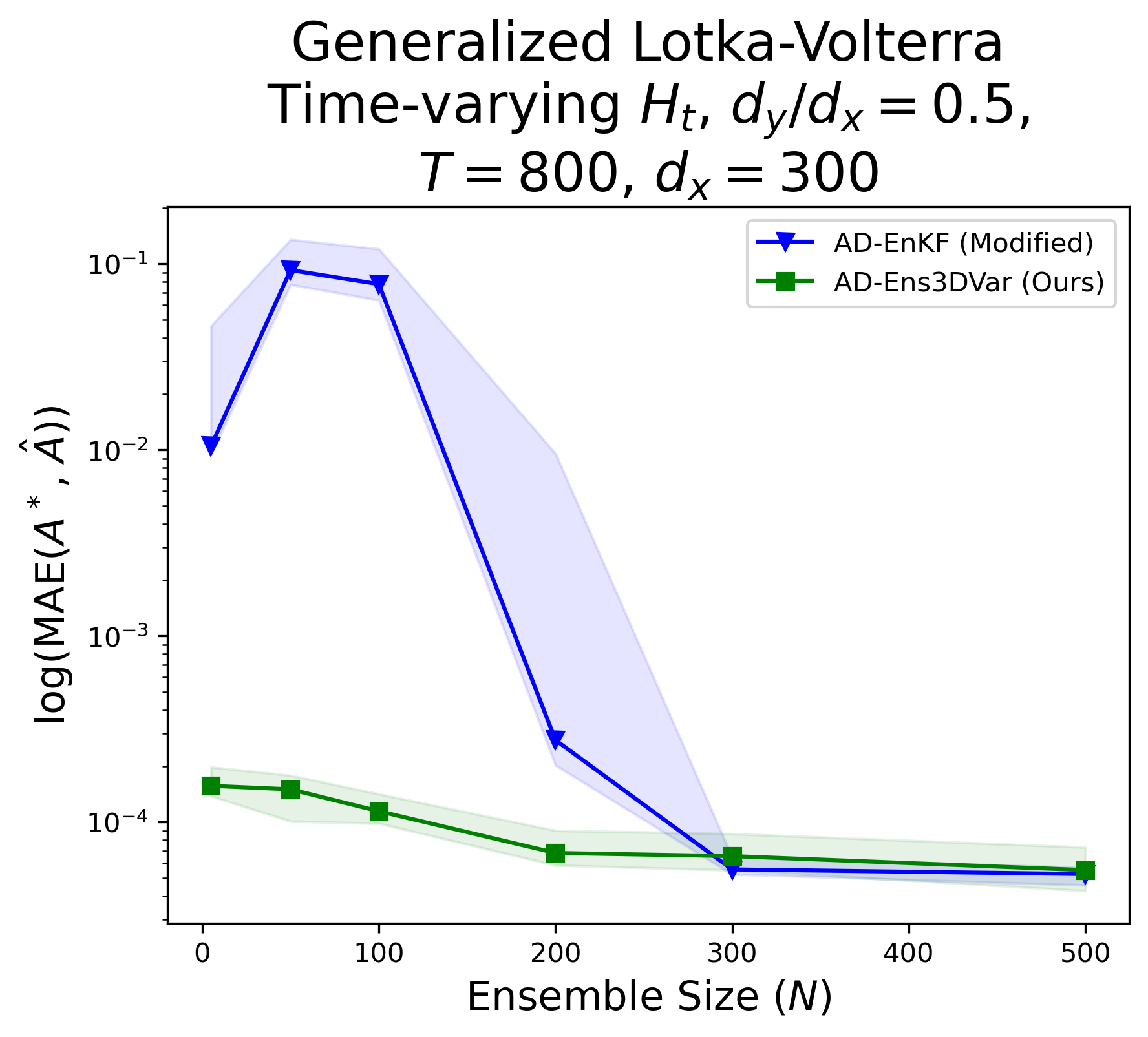}
        \caption{Generalized Lotka-Volterra (Section \ref{sec:GLV}): learning the forecast parameters $A$ where $H_t$ varies with time, $d_y/d_x=0.5,$ $T=800$, and $d_x=300$.}\label{subfig:GLV_varyN_A}
    \end{subfigure}
    ~
    \begin{subfigure}{0.33\textwidth}
        \centering
        \includegraphics[width=\textwidth]{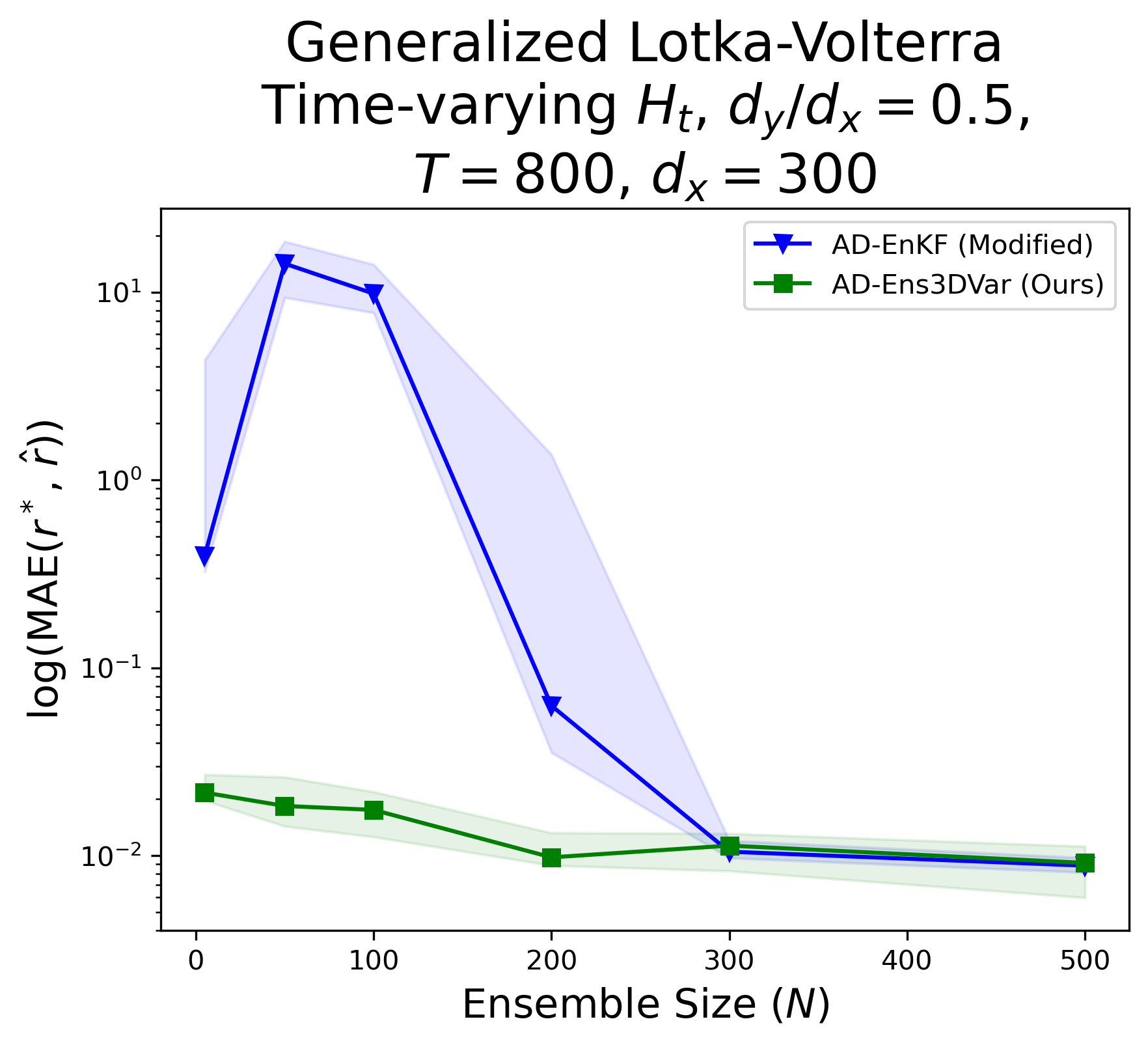}
        \caption{Generalized Lotka-Volterra (Section \ref{sec:GLV}): learning the forecast parameters $r$ where $H_t$ varies with time, $d_y/d_x=0.5,$ $T=800$, and $d_x=300$.}\label{subfig:GLV_varyN_r}
    \end{subfigure}
    
    \caption{Varying the ensemble size for AD-EnKF (Modified) and AD-Ens3DVar for the Clohessy-Wiltshire (Section \ref{sec:CW}), Lorenz-96 (Section \ref{sec:L96}), and generalized Lotka-Volterra (Section \ref{sec:GLV}) examples. The shaded regions correspond to the 0.2 and 0.8 quantiles of forecast performance or forecast parameter estimation error across 10 independent simulations.}
    \label{fig:vary_N}
\end{figure}

Figure \ref{fig:vary_N} shows the evaluation of the learned forecast models across independent simulations for varying ensemble sizes used in AD-EnKF (Modified) and AD-Ens3DVar. Figures \ref{subfig:CW_varyN}, \ref{subfig:GLV_varyN_A}, and \ref{subfig:GLV_varyN_r} show
a sharp degradation in the ability of AD-EnKF (Modified) to learn the true forecast model parameters $\theta_1^*$ when $N<d_x$. This performance degradation can be attributed to the fact that covariance tapering was not used in the Clohessy-Wiltshire or generalized Lotka-Volterra examples, which leads to poor state estimation from EnKF, and therefore a limited ability to learn from this estimation. This phenomenon is less apparent in the Lorenz-96 example in Figure \ref{subfig:L96_varyN} since a covariance tapering radius of 5 is used, and therefore better state estimation for EnKF is possible for $N<d_x$. 

AD-Ens3DVar does not suffer from this degradation in performance due to the additional learned time-independent 3DVar-type covariance matrix that is mixed with an EnKF-type covariance via a learnable covariance mixing weight. Since the covariance mixing parameter is learned, AD-Ens3DVar has the ability to automatically shift the mixing parameter away from the poorly estimated EnKF covariance matrix in favor of the more stable 3DVar-type covariance. Therefore, AD-Ens3DVar provides a more flexible means to learn from observations that better combats ensemble-based failure modes.

For $N > d_x$, AD-Ens3DVar and AD-EnKF (Modified) perform comparably, with no clear advantage in performance for one method versus the other at fixed ensemble sizes.

\section{Evaluation metrics}\label{app:eval_metrics}

\paragraph{Forecast RMSE.} We compute the test forecast root mean square error (RMSE) in the Lorenz-96 experiments to evaluate the performance of the learned model $\mathcal{F}_{\hat \theta_1}$ against the true model $\mathcal{F}_{\theta_1^*}$. To compute these forecast errors, we randomly select $P=500$ states on the Lorenz-96 attractor to serve as initial conditions for forecasts, and compute one-step-ahead forecasts using the true and learned dynamics. The forecast-RMSE is computed as follows
\begin{equation}
    \text{forecast-RMSE} = \sqrt{\frac{1}{d_xP} \sum_{p=1}^P \|\mathcal{F}_{\hat\theta_1}(x_p) - \mathcal{F}_{\theta_1^*}(x_p)\|^2},
\end{equation}
where $x_p$ for $p=1,\dots,P$ is an initial condition sampled on the Lorenz-96 attractor for a fixed simulation set of size $P$, and a one-step-ahead forecast is computed for each of these $P$ initial conditions with the imperfect $\mathcal{F}_{\theta_1}$ and the true dynamics $\mathcal{F}$.

\paragraph{Filter RMSE.} Filter RMSE is computed as 
\begin{equation}
    \text{filter-RMSE} = \sqrt{\frac{1}{d_xT}\sum_{t=1}^T\|x_t - x_t^*\|^2},
\end{equation}
where $x_t = \frac{1}{N}\sum_{n=1}^N x_{t}^{n}$ is the analysis mean at time $t$ computed with $N$ ensemble members, and $x_t^*$ is the ground truth state at time $t$.

\paragraph{Parameter Estimation MAE.} We compute the parameter estimation mean absolute error (MAE) in our generalized Lotka-Volterra and Clohessy-Wiltshire examples as
\begin{equation}
    \text{MAE} = \frac{1}{|\theta_1^*|}\|\hat \theta_{1} - \theta_1^*\|_1,
\end{equation}
where $|\theta_1^*|$ is the number of parameters in vector of true forecast model parameters $\theta_1^*$, and $\hat\theta_1$ is the vector of estimated forecast model parameters.

\section{Supplementary experimental details}\label{app:training_details}

\paragraph{General training details.} Unless otherwise noted, in all experiments, forecast error metrics are computed based on the average forecast errors in the last 10 epochs, with 40 epochs of training total. All experiments demonstrate proper convergence well before training for the full 40 epochs. We additionally utilize the \texttt{torch.optim.Adam} optimizer with varied learning rates for $\theta$ and $\phi$ depending on the experiment, with the remaining settings set to default. We use the \texttt{torch.optim.lr\_scheduler.ReduceLROnPlateau} scheduler, where the patience is set to 5 epochs with the remaining settings set to default. For hyperparameter tuning the learning rates for $\theta$ and $\phi$, we tuned these hyperparameters based on the forecast log-likelihood evaluated at validation observations for 4 independent validation trajectories. All experiments are computed using a single GPU.

We note that the learning rate hyperparameters for training each method are tuned via a loss computed on observational \textit{validation} data (Eq. \eqref{eq:loss} with validation observations), which makes these hyperparameter choices subject to random noise in the data. Therefore, this approach may lead to suboptimal parameter choices. However, this approach is what a practitioner would have access to in a real setting. 

We additionally note that AD-EnKF and AD-Ens3DVar are trained with a fixed ensemble size in an effort to fairly compare the methods. This choice, however, makes AD-Ens3DVar more computationally expensive since an additional $C_\phi$ matrix needs to be learned, but generally in practice, one may be able to learn from smaller ensemble sizes while maintaining some performance threshold.

\paragraph{Learning $\phi$ in AD-EnKF (modified).} Across all AD-EnKF (modified) experiments, the covariance inflation factor is learned, which is a positive scalar; therefore, we learned a ``latent'' representation of this scalar that is allowed to take any value on the real line, and the covariance inflation is computed to be a transformation of this quantity that maps it to $[0,1].$ More formally, we indirectly optimize $\phi$ by optimizing a latent $\tilde \phi$ and applying the transformation $\phi : =  \text{Sigmoid}(\tilde\phi)$.

\paragraph{Learning $\phi_1$ in AD-Ens3DVar.} Similarly, across all AD-Ens3DVar experiments, the covariance mixing parameter is learned, which is another positive value in $[0,1].$ We similarly learn a ``latent'' representation of this scalar that can take any value on the real line, and we define a transformation that maps this value to $[0,1]$. More specifically, we optimize a latent $\tilde \phi_1$ such that $\phi_1 = \text{Sigmoid}(\tilde\phi_1)$.

\paragraph{Learning $\theta_2$ in AD-Ens3DVar and AD-EnKF.} In both AD-Ens3DVar and AD-EnKF, we learn the forecast error covariance $Q_{\theta_2}$ parameterized by $\theta_2$. Across all experiments, we specify that this matrix is diagonal with strictly positive elements along the diagonal. To ensure that the diagonals are strictly positive, we specify that $Q_{\theta_2} = \text{diag}(\text{Softplus}(\theta_2))$ for $\theta_2\in\mathbb{R}^{d_x}$.

\begin{figure}
\centering
\includegraphics[width=\textwidth]{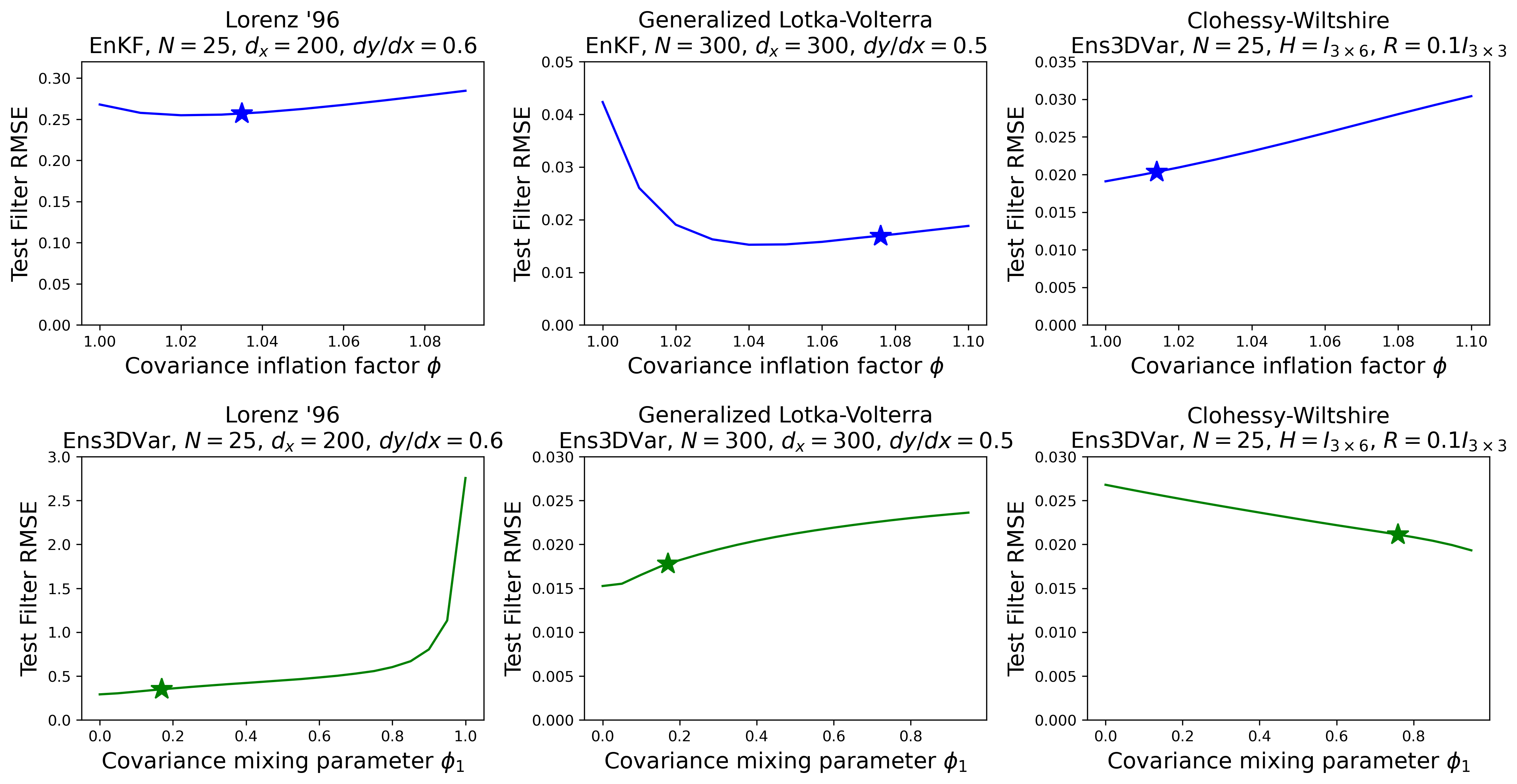}
\caption{Visualization of the test filter RMSEs across examples and methods. The lines in each plot correspond to the test filter RMSE across parameter choices for either the covariance mixing parameter $\phi_1$ in Ens3DVar or the covariance inflation factor $\phi$ in EnKF. The top row corresponds to filtering results using EnKF, the bottom row corresponds to filtering results using Ens3DVar, and each column corresponds to results from one of our experimental examples. The star in each plot corresponds to the value chosen by the learning algorithm. The EnKF evaluations use the learned forecast parameters $\hat\theta$ in the filter, and the Ens3DVar evaluations use both the learned forecast parameters $\hat\theta$ and the learned 3DVar background covariance parameters $\phi_2$ in the filter.}
\label{fig:learned_alpha_gamma}
\end{figure}

\paragraph{Evaluating the learned covariance inflation factor $\phi$ in AD-EnKF and the learned covariance mixing parameter $\phi_1$ in AD-Ens3DVar.} Figure \ref{fig:learned_alpha_gamma} visualizes how close to optimal the learned covariance inflation factors are for AD-EnKF across the three examples, as well as how close to optimal the learned covariance mixing parameters are for AD-Ens3DVar across the three examples. Since we learn from noisy, potentially sparse observations in a joint optimization with both forecast and filtering parameters, the converged values for these parameters are close to but not exactly optimal. 

\subsection{Clohessy-Wiltshire}\label{app:CW_training_LL}
\subsubsection{Data generation}

We specify that the true parameter that we want to recover is $\theta^*=0.0013$. The system is observed with a time step $\Delta t = 10.$ All experiments are trained with $T=800$ time points, where 8 independent trajectories of length $T$ were used for training, 4 independent trajectories of length $T$ were used for validation, and 4 independent trajectories of length $T$ were used for testing.

\subsubsection{Training details}

The subsequence length $L$ for the maximum number of assimilation time points before auto-differentiating is set to $L=20$ for all experiments. To construct $H$ when $d_y/d_x<1$, we systematically remove observability, where for example, if $d_y/d_x = 0.66$, we do not observe every third dimension of the state.

For AD-3DVar-$C$ experiments, we initialize $B_{\phi^{(0)}}$ as $0.1I_{d_x}$, for AD-3DVar-$K$ experiments, we initialize $K_{\phi^{(0)}}$ as $H = 0.1I_{d_y, d_x}$, for AD-EnKF experiments, we initialize $\hat C_0 := 5I_{d_x\times d_x}$ and the covariance inflation factor $\phi = 0.1$, and for AD-Ens3DVar experiments, we initialize $\tilde C^{\theta,\phi}_0$ with $B_{\phi_2}:= 0.1I_{d_x\times d_x}$, the covariance mixing parameter $\phi_1 := 0.5$, and $\hat C_0:=5I_{d_x\times d_x}$. For both AD-EnKF and AD-Ens3DVar, we initialize $Q_{\theta_2^{(0)}} = 0.1I_{d_x\times d_x}$. The initial condition $x_0$ is drawn from the distribution $\mathcal{N}(x_0^*,\tilde C^{\theta,\phi}_0)$, where $x_0^*=[1.0,0.5,1.0,-0.001375, -0.000275, -0.001375]^\top$, which provides a perturbation of the true initial state $x_0^*$.

We also note that the learnable forecast parameter $\theta_1$ is a strictly positive number. In order to strictly enforce this property, we learned a latent representation of this parameter, denoted as $\tilde \theta_1$, such that $\theta_1 =\text{Softplus}(\tilde \theta_1)$. We directly optimize $\tilde \theta_1$, and feeding this parameter through the Softplus operation ensures that $\theta_1$ is always positive.

\subsubsection{Additional filtering visualization} \label{app:CW_filtering_yz}

\begin{figure}
\centering
\includegraphics[width=\textwidth]{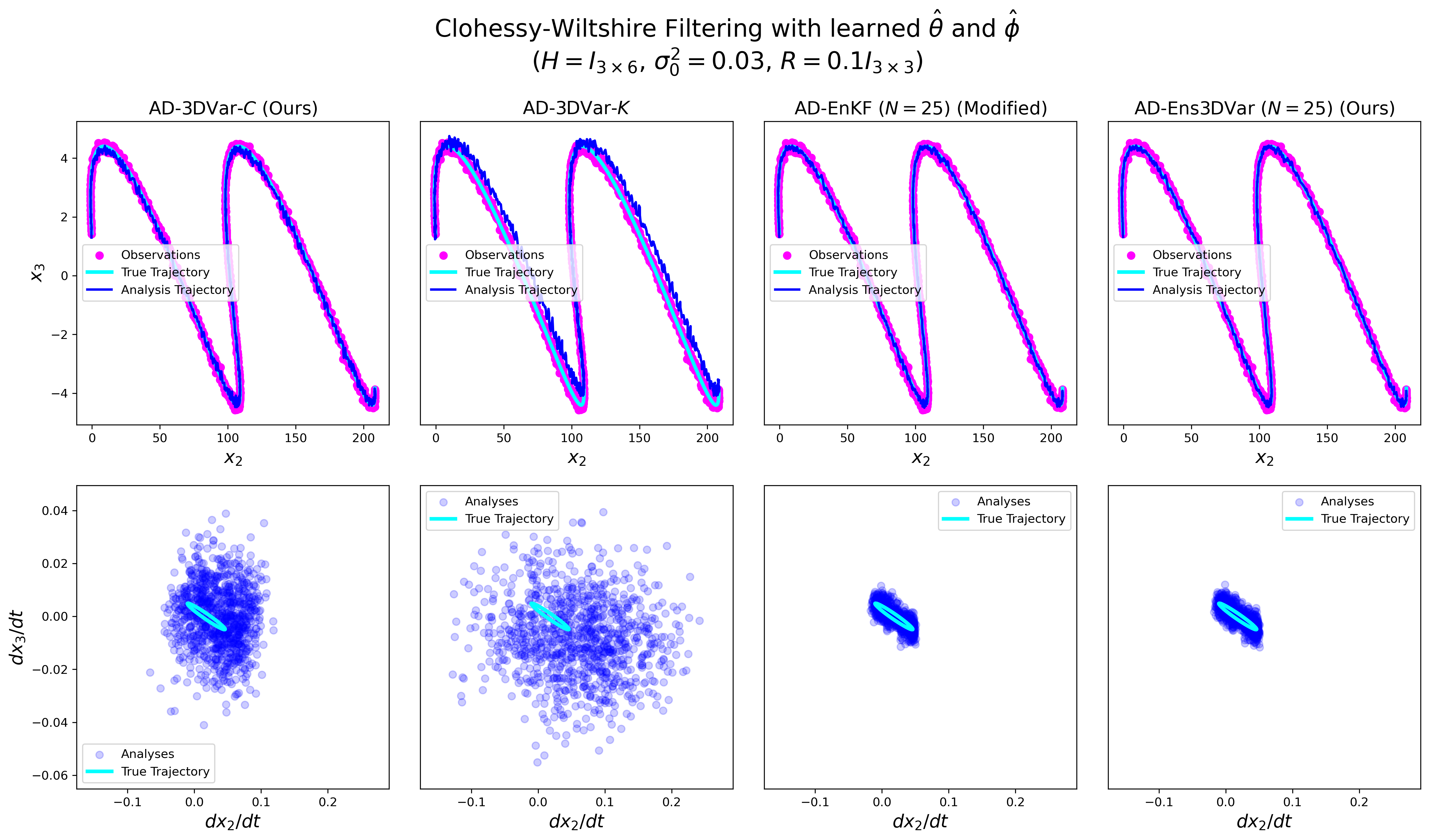}
\caption{Clohessy-Wiltshire filtering (Section \ref{sec:CW}). Visualization of observations, true trajectories, and estimated positions for $y$ and $z$, and velocities $\frac{dy}{dt}$ and $\frac{dz}{dt}$ across each of the four filtering methods using the learned forecast parameters $\hat \theta$ and learned filtering parameters $\hat \phi$, evaluated on test data. In this example, the observation noise is fixed at $R = 0.1I_{d_y\times d_y}$, and the variance of the initialization of $\theta_1^{(0)}$ is fixed at $\sigma_0^2 = 0.03$.}
\label{fig:CW_filtering_yz}
\end{figure}

To supplement Figure \ref{fig:CW_filtering}, we additionally plot the observed, filtered, and true $y$ and $z$ positions as well as the filtered and true $\frac{dy}{dt}$ and $\frac{dz}{dt}$ velocities in Figure \ref{fig:CW_filtering_yz}. Similar to the conclusions drawn from Figure \ref{fig:CW_filtering}, AD-EnKF and AD-Ens3DVar accurately recover the positions and velocities with relatively low error. AD-3DVar-$C$ struggles more to estimate the velocity components but can reasonably capture the positions, which suggests that $C$ is reasonably estimated along dimensions where observations are present and struggles otherwise. AD-3DVar-$K$ also struggles to capture the velocity components and additionally shows a bias in its estimates of the position components, which suggests that the estimate of $K$ is suboptimal.

\subsubsection{Kalman filter implementation}\label{app:CW_Kalman_filter}

The Clohessy-Wiltshire equations describe a deterministic \textit{linear} differential equation, where equations \eqref{eq:CW_d2xdt2}-\eqref{eq:CW_d2zdt2} can be rewritten in the notation
\begin{equation}
    \frac{d x}{dt} = F_\theta  x,
\end{equation}
\noindent where $ x = \left[x_1,x_2,x_3,\frac{dx_1}{dt}, \frac{dx_2}{dt}, \frac{dx_3}{dt}\right]$, and $F_\theta \in \mathbb{R}^{6 \times 6}$ is a matrix parameterized by $\theta$. Since these equations are linear, forecasts of this system are provided by the closed-form solution
\begin{equation}
    x_{t' + \Delta t'} = e^{F_\theta \Delta t'} x_{t'}
\end{equation}
\noindent for a time step $\Delta t'>0$, where $t'$ denotes the time of the system (as opposed to the time step $t$ of the filtering process). 

The Kalman filter algorithm used to compute the optimal log-likelihood is given in Algorithm \ref{alg:CW_KF}, where the log-likelihood is given by 
\begin{equation}
    \mathcal{L}_{\text{KF}}(\theta) := \sum_{t=1}^T\log \mathcal{N}(y_t; H_t\hat x_t, H_t P_{t|t-1}^\theta H_t^\top + R_t),
\end{equation}
\noindent and $\hat x_t$ and $P_{t|t-1}^\theta$ are defined in Algorithm \ref{alg:CW_KF}. We note that the log-likelihood drops the parameters $\phi$ since the optimal filtering parameters are specified by the Kalman filter, and therefore do not need to be learned or prespecified.

\begin{algorithm}[h]
\caption{Kalman filter \citep{Kalman1960}}\label{alg:CW_KF}
\begin{algorithmic}
\State \textbf{Input:} Initialization $x_0$, observations $y_{1:T}$, observation matrices $H_{1:T}$, observation error matrices $R_{1:T}$, forecast matrix $e^{F_\theta \Delta t'}$ with timestep $\Delta t'$, initialization error covariance matrix $P_0$

\For{$t=1,\dots,T$}
\State \# Compute the forecast mean and covariance
\State $\hat x_t = e^{F_\theta \Delta t'}x_{t-1}$ 
\State $P_{t|t-1}^\theta = (e^{F_\theta \Delta t})P_{t-1|t-1}^\theta (e^{F_\theta \Delta t})^\top$

\State 
\State \# Compute the Kalman gain
\State $K_t = P_{t|t-1}^\theta H_t^\top (H_tP_{t|t-1}^\theta H_t^\top + R_t )^{-1}$

\State 
\State \# Compute the analysis and analysis covariance
\State $x_t = \hat x_t + K_t(y_t - H_t \hat x_t)$
\State $P_{t|t}^{\theta} = (I-K_tH_t)P^\theta_{t|t-1}(I-K_tH_t)^\top + K_t R_tK_t^\top$ \\
\EndFor

\State \textbf{Output:} Analysis estimates $x_{1:T}$
\end{algorithmic}
\end{algorithm}

\begin{figure}
\centering
\includegraphics[width=\textwidth]{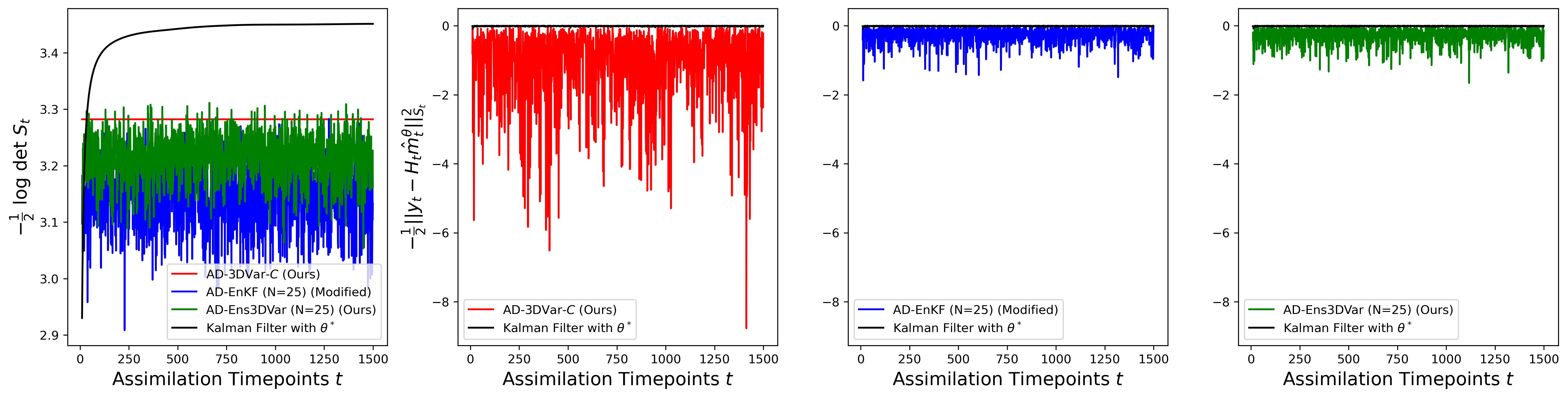}
\caption{Visualization of each component of the forecast log-likelihood evaluated at observations, given in \eqref{eq:loss} for filtering with the learned $\hat \theta$ and $\hat \phi$ from AD-3DVar-$C$, AD-EnKF, and AD-Ens3DVar compared to the optimal Kalman filter. \textbf{(Left)} Component of the log-likelihood that computes $-\frac{1}{2}\log \det S_t$ for each time $t$. \textbf{(Right three)} Residual component of the loss that computes $-\frac{1}{2}\|y_t - H_t \hat m_t^\theta\|_{S_t}^2$ for each time $t$.}
\label{fig:CW_LL_components}
\end{figure}

Figure \ref{fig:CW_LL_components} supplements Figure \ref{fig:CW_LL} by visualizing how each component contributes to the overall computed log-likelihood values. Interestingly, the covariance learned from AD-3DVar-$C$ is the closest match to the true $-\frac{1}{2}\log \det S_t$ given by the Kalman filter using the true forecast parameters $\theta^*$, and both AD-EnKF and AD-Ens3DVar provide noisier and more biased estimates of this quantity. However, if we visualize the contribution of the $-\frac{1}{2}\|y_t - H_t\hat m_t^\theta\|_{S_t}^2$ term to the overall log-likelihood, we can see that AD-3DVar-$C$ provides much noisier estimates, which ultimately dominates the computed error in the log-likelihood computation shown in Figure \ref{fig:CW_LL}. Both AD-EnKF and AD-Ens3DVar better estimate this term given by the optimal Kalman filter and show similar noise levels.

\subsection{Lorenz-96}\label{app:details_L96}

\subsubsection{Data generation} 

We implement the Lorenz-96 system with a forcing term $F =8$, which elicits chaotic behavior of the system. The system is solved with a fourth-order Runge-Kutta (RK4) solver with time step $\Delta t = 0.05$, which is the same time step assumed for observations to arrive. All experiments are trained with $T=1,200$, where 8 independent trajectories of length $T$ were used for training, 4 independent trajectories of length $T$ were used for validation, and $4$ independent trajectories of length $T$ were used for testing.

\begin{figure}
\centering
\includegraphics[width=0.3\textwidth]{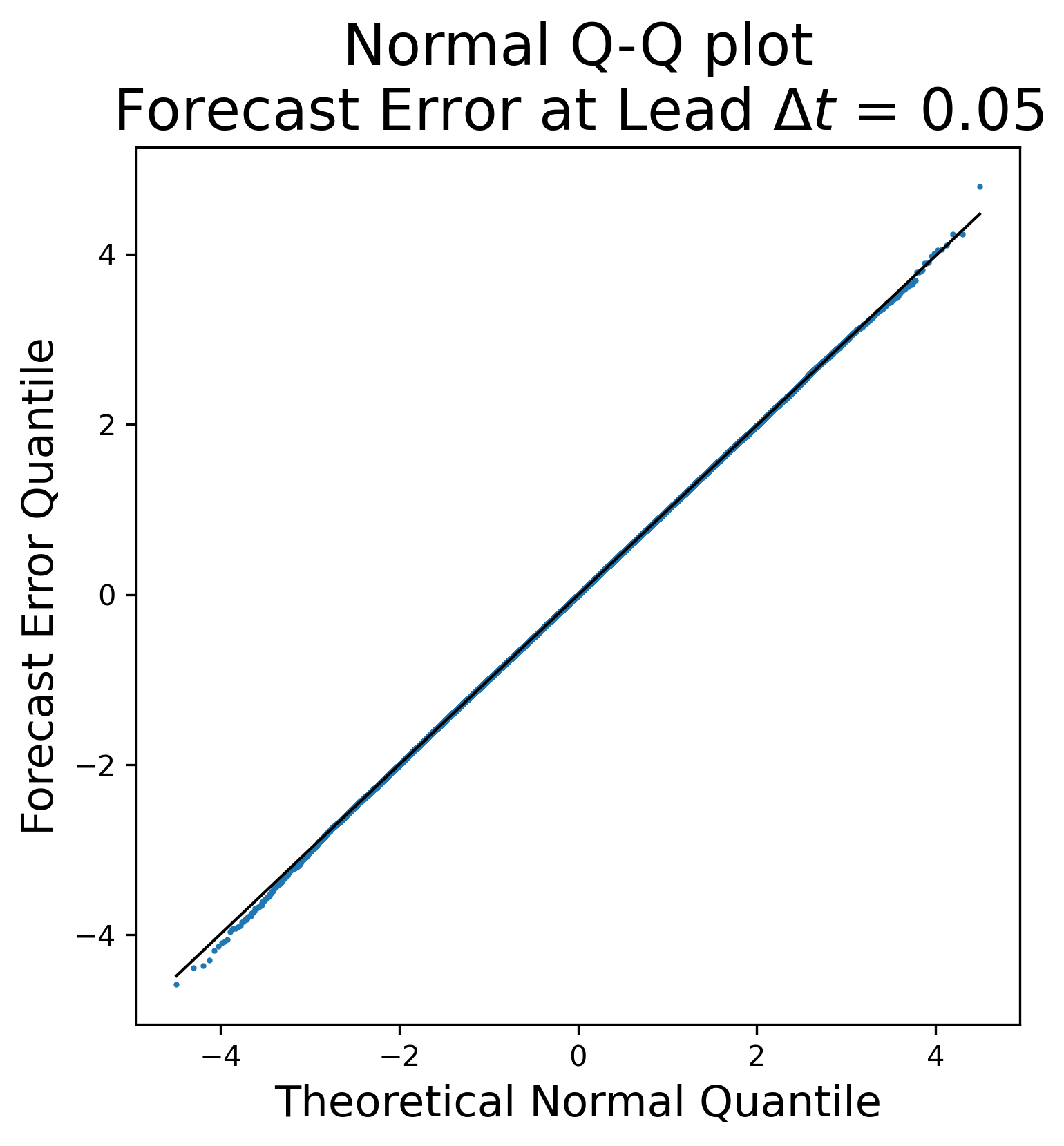}
\caption{Visualization of the normal quantile-quantile (QQ-plot) of the forecast error statistics for the $d_x=40$ Lorenz-96 system for forecast time steps $\Delta t$. Each point represents the difference $x^*_{i,t} - \hat x_{i,t}$ across time points $t$ for dimensions $i$}. 
\label{fig:l96_qq}
\end{figure}

For this chaotic system, we assess the Gaussianity of the forecast error statistics computed as $x_{i,t}^* - \hat x_{i,t}$ for our chosen $\Delta t$. Figure \ref{fig:l96_qq} shows a Normal Q-Q plot for the forecast error statistics of the $d_x=40$ Lorenz-96 system with a forecast time step
$\Delta t = 0.05$. The plot clearly shows that the marginal empirical quantiles of the forecast residuals are closely aligned with the theoretical quantiles of a Normal distribution, which suggests 
that the Gaussian forecast assumptions are reasonable in this setting.

\subsubsection{Training details}

The subsequence length $L$ for the maximum number of assimilation time points before auto-differentiating is set to $L=20$ for all experiments. For AD-3DVar-$C$ experiments, we initialize $C_{\phi^{(0)}}$ as $I_{d_x}$, for AD-3DVar-$K$ experiments, we initialize $K_{\phi^{(0)}}$ as $H = I_{d_y, d_x}$, for AD-EnKF experiments, we initialize $\hat C_0 := 25I_{d_x\times d_x}$ and $\phi = 0.1$, and for AD-Ens3DVar experiments, we initialize $\tilde C^{\theta,\phi}_0$ with $C_{\phi_2}:= I_{d_x\times d_x}$, $\phi_1 := 0.5$, and $\hat C_0:=25I_{d_x\times d_x}$. For AD-EnKF and AD-Ens3DVar experiments, we utilize the Gaspari-Cohn tapering matrix \citep{Gaspari1999} with a tapering radius of 5, and initialize $Q_{\theta_2^{(0)}} = 2I_{d_x\times d_x}$. Lastly, the initialization $x_0$ is randomly drawn from $\mathcal{N}(0,\tilde C_0)$, which indicates a lack of knowledge in the initial condition.

\subsection{Generalized Lotka-Volterra}

\subsubsection{Data generation}

The generalized Lotka-Volterra equations are solved with a fourth-order Runge-Kutta (RK4) solver with time step $\Delta t = 0.05$, which is the same time step assumed for observations to arrive. For all experiments, for a particular choice of trajectory length $T$, 8 independent trajectories of length $T$ were used for training, 4 independent trajectories of length $T$ were used for validation, and $4$ independent trajectories of length $T$ were used for testing. 

To specify the true values $A^*$, we specified this matrix to have the same block structure shown in Figure \ref{fig:A_GLV} where $a_1^*=a_5^*=a^*_8=a_{10}^*=0.1$, $a_2^*=a_6^*=a^*_9=0.05$, $a_3^*=a_7^*=0.01$, and $a_4^*=-0.04$. 
To generate a true $r^*$ vector for a fixed value of $d_x$, a $d_x$ length vector is simulated from a standard Gaussian distribution $\mathcal{N}(0,I_{d_x})$, which we denote as $\tilde x^*_s$. Since $x^*_s$ represents the steady-state species abundances, this vector is strictly non-negative. We form $x^*_s$ as $x_s^* = \text{Softplus}(\tilde x_s^*)$, where $r^*$ is then constructed as $r^* = -A^*x_s^*$.

\subsubsection{Training details}

The subsequence length $L$ for the maximum number of assimilation time points before auto-differentiating is set to $L=20$ for all experiments. For AD-3DVar-$C$ experiments, we initialize $B_{\phi^{(0)}}$ as $0.1I_{d_x}$, for AD-EnKF experiments, we initialize $\hat C_0 := I_{d_x\times d_x}$ and $\phi = 0.1$, and for AD-Ens3DVar experiments, we initialize $\tilde C^{\theta,\phi}_0$ with $B_{\phi_2}:= 0.1I_{d_x\times d_x}$, $\phi_1 := 0.5$, and $\hat C_0:=I_{d_x\times d_x}$. For both AD-EnKF and AD-3DVar experiments, we initialize $Q_{\theta_2^{(0)}} = 0.1I_{d_x\times d_x}$. The initialization $x_0$ is drawn from $\mathcal{N}(\textbf{1}_{d_x},\tilde C_0^{\theta,\phi})$.

We also note that the solutions of the generalized Lotka-Volterra equations are strictly positive since the equations describe abundances, so we impose some additional transformations at particular steps of the problem to encourage positive analyses. At steps of the assimilation procedure that require drawing random Gaussian samples, such as in the forecast step $\mathcal{M}_\theta(\cdot)$ and the observation perturbation $\mathcal{Y}(\cdot;R)$ in \eqref{eq:filter}, we feed the value plus the simulated noise through the absolute value function. 

\subsection{Two-layer Quasi-Geostrophic}

\subsubsection{Data generation}\label{app:subsubsec:qg_datagen}

To construct a set of true model parameters, we specified the true values as $\beta^* = 0.196$, $U^* = 1$, $\tau_f^* = 0.2$, $k_d^* = 1$ and $\nu^* = 1.024 \times 10^{-3}$, which corresponds to a system that exhibits turbulent eddies throughout the domain.

\paragraph{Parameter estimation.}The two-layer QG model equations are solved with a fourth-order Runge-Kutta (RK4) solver with time step $\Delta t = 0.005$, and observations arrive every $\Delta t_\text{obs} = 0.5$. For all experiments, for a particular choice of trajectory length $T$, 1 trajectory of length $T$ is used for training, and two other trajectories temporally disjoint from each other and the training data are used for validation and testing. We train with $T =100$ time points where observations are collected, with 20 time points for validation, and 20 for testing. This data sequence is generated from a random initialization and spun up for 300 time units to reach equilibrium.

\paragraph{Emulation.} We simulate a slightly different dataset for the emulator task to learn shorter-range dynamics by simulating a ground truth trajectory with $\Delta t=0.0025$, saved every $0.05$ time units, where observations arrive every $\Delta t_\text{obs} =0.1$. We train with $T = 1000$ observation time points (100 time units), with 20 observation times for validation and 20 for testing. This data sequence is generated from a random initialization and spun up for 800 time units to reach equilibrium.

\begin{figure}[htbp]
    \centering
    \begin{minipage}[c]{0.46\linewidth}
      \centering
      \begin{tikzpicture}[
        font=\footnotesize, >=stealth, line width=0.4pt,
        box/.style={draw, rectangle, minimum height=0.55cm, inner sep=3pt},
        circ/.style={draw, circle, minimum size=0.7cm, inner sep=1pt},
        op/.style={draw, circle, minimum size=0.45cm, inner sep=0pt},
        lab/.style={font=\scriptsize, midway}]
      \node[circ] (x)   at (0, 0)     {$x$};
      \node[box]  (nrm) at (0,-1.1)   {$\mathrm{Norm}$};
      \node[box]  (e1)  at (0,-2.2)   {$\mathrm{Enc}_1$};
      \node[box]  (e2)  at (0,-3.3)   {$\mathrm{Enc}_2$};
      \node[box]  (e3)  at (0,-4.4)   {$\mathrm{Enc}_3$};
      \node[box]  (bt)  at (1.6,-5.5) {$\mathrm{Bot}$};
      \node[box]  (d3)  at (3.2,-4.4) {$\mathrm{Dec}_3$};
      \node[box]  (d2)  at (3.2,-3.3) {$\mathrm{Dec}_2$};
      \node[box]  (d1)  at (3.2,-2.2) {$\mathrm{Dec}_1$};
      \node[box]  (hd)  at (3.2,-1.1) {$\mathrm{Head}$};
      \node[op]   (mul) at (3.2, 0)   {$\odot$};
      \node[op]   (add) at (1.6, 0)   {$+$};
      \node[circ] (out) at (1.6, 1.1) {$\hat x$};

      \draw[->] (x) -- (nrm);
      \draw[->] (nrm) -- (e1);
      \draw[->] (e1) -- node[lab, left]{$\downarrow 2$} (e2);
      \draw[->] (e2) -- node[lab, left]{$\downarrow 2$} (e3);
      \draw[->] (e3) |- node[lab, pos=0.3, left]{$\downarrow 2$} (bt);
      \draw[->] (bt) -| node[lab, pos=0.7, right]{$\uparrow 2$} (d3);
      \draw[->] (d3) -- node[lab, right]{$\uparrow 2$} (d2);
      \draw[->] (d2) -- node[lab, right]{$\uparrow 2$} (d1);
      \draw[->] (d1) -- (hd);
      \draw[->] (hd) -- (mul);
      \draw[->] (mul) -- (add);
      \draw[->] (x) -- (add);
      \draw[->] (add) -- (out);
      \draw[->, dashed] (e1) -- (d1);
      \draw[->, dashed] (e2) -- (d2);
      \draw[->, dashed] (e3) -- (d3);
      \node[font=\scriptsize, right=0.25cm] (s) at (mul.east) {$s$};
      \draw[->] (s) -- (mul);
      \end{tikzpicture}

      \vspace{2pt}
      {\footnotesize \textbf{(a)} Network structure}
    \end{minipage}%
    \hfill
    \begin{minipage}[c]{0.52\linewidth}
      \centering
      \footnotesize
      \setlength{\tabcolsep}{3pt}
      \begin{tabular}{|l|l|l|l|l|}
        \hline
                         & \#in            & \#out & grid            & kernel size \\ \hline
        $\mathrm{Enc}_1$ & 3 ($2{+}1$)     & 32    & $96\times48$    & $3$, circ. \\ \hline
        $\mathrm{Enc}_2$ & 32              & 64    & $48\times24$    & $3$, circ. \\ \hline
        $\mathrm{Enc}_3$ & 64              & 128   & $24\times12$    & $3$, circ. \\ \hline
        $\mathrm{Bot}$   & 128             & 256   & $12\times6$     & $3$, circ. \\ \hline
        $\mathrm{Dec}_3$ & 384 ($256{+}128$) & 128 & $24\times12$    & $3$, circ. \\ \hline
        $\mathrm{Dec}_2$ & 192 ($128{+}64$)  & 64  & $48\times24$    & $3$, circ. \\ \hline
        $\mathrm{Dec}_1$ & 96 ($64{+}32$)    & 32  & $96\times48$    & $3$, circ. \\ \hline
        $\mathrm{Head}$  & 32              & 2     & $96\times48$    & $1$ \\ \hline
      \end{tabular}

      \vspace{4pt}
      {\footnotesize \textbf{(b)} Network details}
    \end{minipage}

    \caption{Structure of the QG emulator with learnable parameters $\theta_1$, a residual U-Net with 1,949,730 learnable parameters.
    Each $\mathrm{Enc}$, $\mathrm{Bot}$ and $\mathrm{Dec}$ block consists of two repetitions of the sequence $3\times3$ convolution $\to$ GroupNorm $\to$ SiLU with circular padding in
    both directions. $\downarrow 2$ is $2\times2$ average pooling, and
    $\uparrow 2$ is nearest-neighbor upsampling. Dashed arrows are skip
    connections, concatenated along the channel dimension. $\mathrm{Head}$ is a zero-initialized $1\times1$
    convolution, so the untrained network is the identity map. The output is
    $\hat x = x + s\odot\mathrm{Head}(\cdot)$, where $s$ is the standard deviation of the one-step increment $x_{t+1} - x_t$. The input to the network is $x.$}
    \label{fig:qg_emulator_arch}
  \end{figure}

  \begin{figure}
\centering
\includegraphics[width=0.8\textwidth]{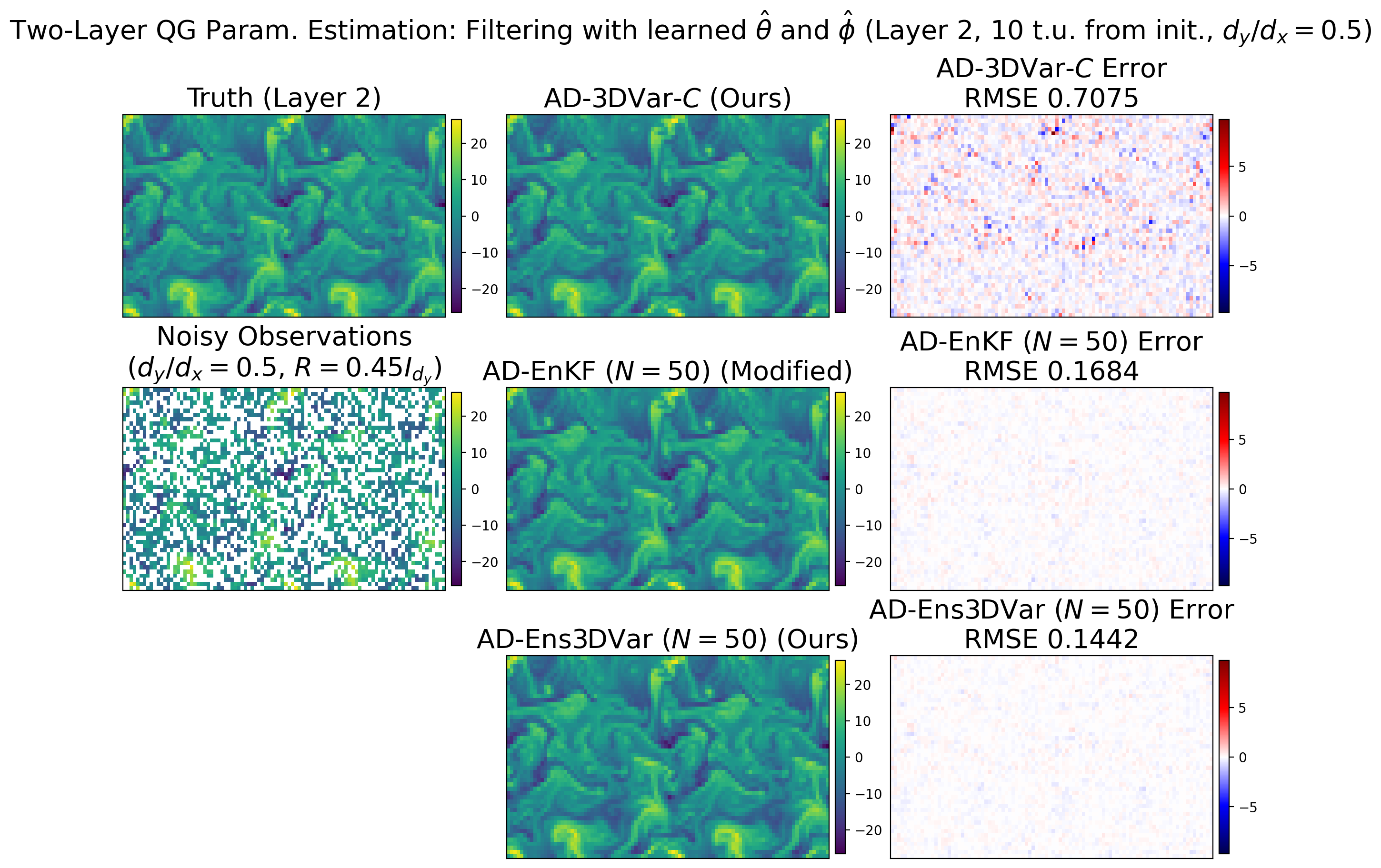}
\caption{Two-layer QG parameter estimation (Section \ref{sec:QG}). Visualization of filtering on test observational data for layer 2 using the learned $\hat\theta$ and $\hat\phi$ values across the methods AD-3DVar-$C$, AD-EnKF with $N=50$ ensemble members, and AD-Ens3DVar with $N=50$ ensemble members where $d_{y_t}/d_x=0.5$. The image shows a snapshot of the analyses across methods at the end of the test filter window, corresponding to 10 time units (t.u.) from initialization.}
\label{fig:QG_filtering_layer2}
\end{figure}

\begin{figure}
\centering
\includegraphics[width=0.8\textwidth]{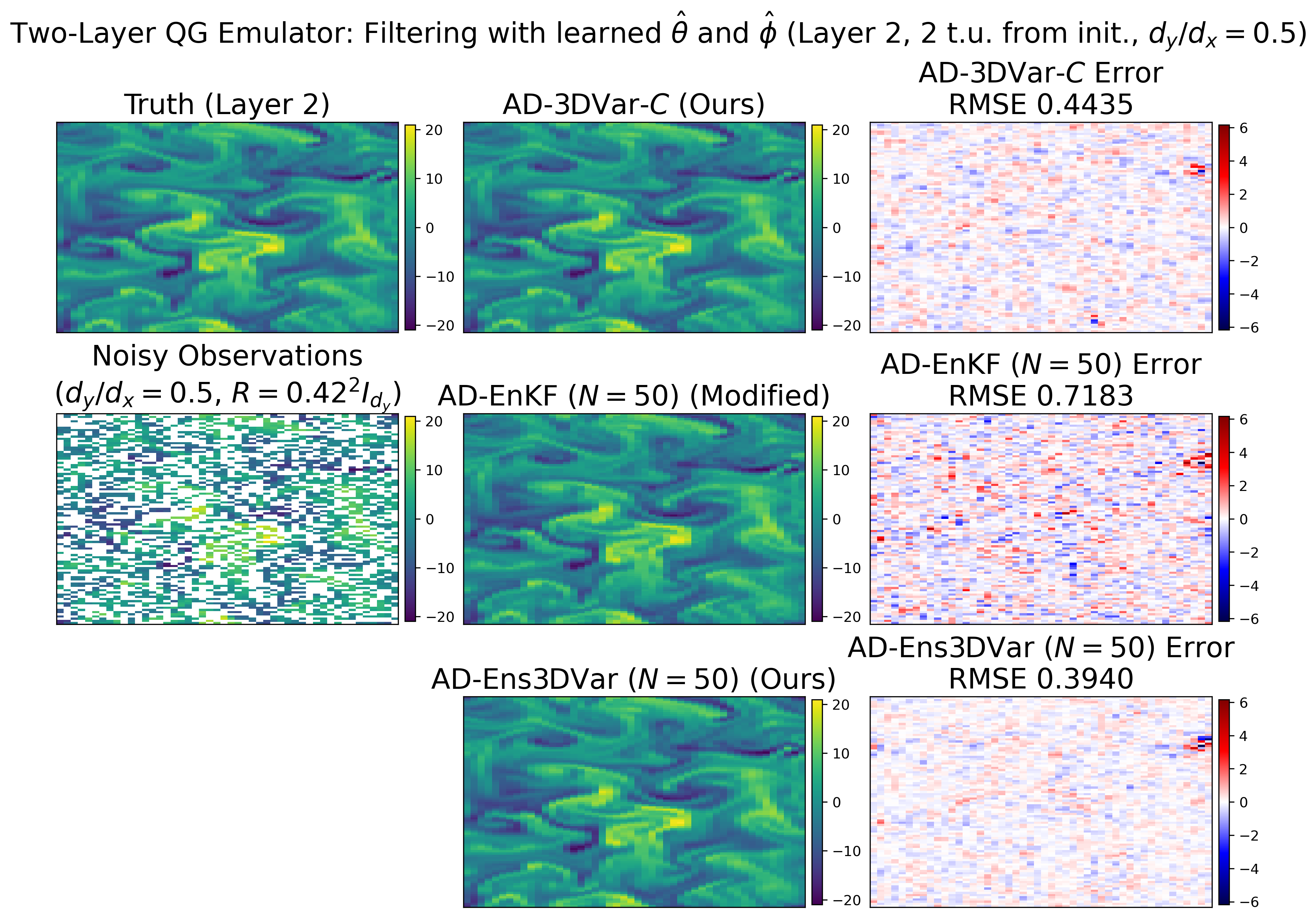}
\caption{Two-layer QG emulator (Section \ref{sec:QG}). Visualization of filtering on test observational data for layer 2 using the learned $\hat\theta$ and $\hat\phi$ values across the methods AD-3DVar-$C$, AD-EnKF with $N=50$ ensemble members, and AD-Ens3DVar with $N=50$ ensemble members where $d_{y_t}/d_x=0.5$. The image shows a snapshot of the analyses across methods at the end of the test filter window, corresponding to 2 time units (t.u.) from initialization.}
\label{fig:QG_filtering_emulator_layer2}
\end{figure}

\subsubsection{Training details}

\paragraph{Parameter estimation.} We train for 20 epochs for each model and choose hyperparameters from the average performance on validation data on the final 3 epochs of training. The subsequence length $L$ for the maximum number of assimilation time points before auto-differentiating is set to $L=10$ for all two-layer QG parameter estimation experiments. In the initialization of $L_0$ for AD-3DVar-$C$, we set $\sigma_\text{init} = 1.8$ and $\gamma = 0.2$ using the method outlined in Appendix \ref{app:3dvarC_qg}. For AD-Ens3DVar and AD-EnKF, $Q_{\theta_2^{(0)}}= 0.033I_{d_x\times d_x}$. For all three methods, we perturb a true state as an initial condition with $\hat C_0:=1.8I_{d_x\times d_x}$, and all three approaches use a Gaspari–Cohn tapering matrix $\rho$ with radius 5. For both AD-3DVar-$C$ and AD-Ens3DVar, for each of the 10 initialized models, $C_{\phi^{(0)}}$ is initialized at a climatological covariance of its own perturbed model forecasts using 801 snapshots.

\paragraph{Emulation.} We train for 20 epochs for each model and similarly choose hyperparameters from the average performance on validation data over the final three epochs
of training. The subsequence length $L$ for the maximum number of assimilation time points before auto-differentiating is set to $L=10$ for all two-layer QG emulator experiments. In the initialization of $L_0$ for AD-3DVar-$C$, we set $\sigma_\text{init} = 1.66$ and $\gamma = 0.415$. For AD-Ens3DVar and AD-EnKF, $Q_{\theta_2^{(0)}}= I_{d_x\times d_x}$. For all three methods, we perturb a true state as an initial condition with $\hat C_0:=2.75I_{d_x\times d_x}$, and all three approaches use a Gaspari–Cohn tapering matrix $\rho$ with radius 5. For AD-3DVar and AD-Ens3DVar, $C_{\phi^{(0)}}$ is initialized to a climatological covariance built from 801 snapshots of a two-layer QG model with perturbed coefficients. We provide a sketch of the neural network details in Figure \ref{fig:qg_emulator_arch}.

\subsubsection{Supplementary figures}

To supplement Figure \ref{fig:QG_filtering_layer1} in the main text that shows the filtering estimate for layer 1 across our auto-differentiable filtering methods for the parameter estimation problem, we include the layer 2 estimates in Figure \ref{fig:QG_filtering_layer2}. Similarly, we visualize layer 2 estimates across methods for the emulator setting in Figure \ref{fig:QG_filtering_emulator_layer2}.

\end{appendix}
\end{document}